\documentclass[twoside]{article}

\usepackage[accepted]{aistats2025}

\usepackage[misc]{ifsym}
\usepackage{natbib}

\usepackage{mathtools}
\usepackage{amsmath}
\usepackage{amssymb}
\usepackage{amsthm}
\usepackage{bm}
\DeclareFontFamily{U}{mathx}{\hyphenchar\font45}
\DeclareFontShape{U}{mathx}{m}{n}{
      <5> <6> <7> <8> <9> <10>
      <10.95> <12> <14.4> <17.28> <20.74> <24.88>
      mathx10
      }{}
\DeclareSymbolFont{mathx}{U}{mathx}{m}{n}
\DeclareFontSubstitution{U}{mathx}{m}{n}
\DeclareMathAccent{\widebar}{0}{mathx}{"73}

\usepackage{microtype}
\usepackage{graphicx}
\usepackage{subfigure}
\usepackage{booktabs} 
\usepackage{tabularx} 
\usepackage{rotating} 
\usepackage{fontawesome}

\usepackage{multirow}

\usepackage{url}

\usepackage{breakurl}
\usepackage[breaklinks]{hyperref}

\usepackage{algorithm}
\usepackage{algorithmic}

\usepackage{titletoc}
\usepackage[subfigure]{tocloft}
\usepackage[dvipsnames, table, xcdraw]{xcolor}
\definecolor{LightGrey}{RGB}{190,190,190}

\usepackage{pgfplots}
\usepgfplotslibrary{groupplots}
\pgfplotsset{compat=newest}

\usepackage{tikz}
\usepackage{subcaption}
\usepgfplotslibrary{fillbetween}

\providecolor{dtured}{rgb}{0.60, 0.00, 0.00} %
\definecolor{darkblue}{rgb}{0.0, 0.0, 0.55}

\usepackage{hyperref}
\hypersetup{
    colorlinks=true,
    linkcolor=darkblue,
    citecolor=darkblue,
    urlcolor=darkblue,
}

\DeclareMathOperator*{\argmin}{arg\,min}

\theoremstyle{plain}
\newtheorem{theorem}{Theorem}[section]
\newtheorem{proposition}[theorem]{Proposition}

\theoremstyle{definition}

\theoremstyle{remark}

\usepackage[textsize=tiny]{todonotes}

\newcommand\norm[1]{\left\lVert#1\right\rVert}
\newcommand\abs[1]{\left|#1\right|}

\renewcommand{\vec}[1]{\mathbf{#1}}
\renewcommand{\d}[1]{\ensuremath{\operatorname{d}\!{#1}}}

\def\K{\mathcal{K}}

\def\P{\mathcal{P}}
\def\RR{\mathbb{R}}
\def\SS{\mathbb{S}}
\def\EE{\mathbb{E}}

\def\PP{\mathbb{P}}

\def\M{\mathcal{M}}

\def\ot{\mathcal{W}^2}
\def\sot{\mathcal{SW}^2}
\def\tnabla{{\widetilde{\nabla}}}
\def\dd{\mathrm{d}}

\def\retr{\text{Retr}}
\def\renabla{\widetilde{\nabla}}
\def\bL{\widetilde{L}}
\def\discount{\textit{Discount}}
\def\objvar{P}

\usepackage{glossaries}
\glsdisablehyper

\newacronym{xai}{XAI}{explainable artificial intelligence}
\newacronym{ot}{OT}{optimal transport}
\newacronym{mmd}{MMD}{maximum mean discrepancy}
\newacronym{eot}{EOT}{entropic optimal transport}
\newacronym{ce}{CE}{counterfactual explanations}
\newacronym{dce}{DCE}{distributional counterfactual explanations}
\newacronym{gce}{GCE}{group-based counterfactual explanations}
\newacronym{kl}{KL}{Kullback-Liebler}
\newacronym{clt}{CLT}{central limit theorem}
\newacronym{1d}{1D}{one-dimensional}
\newacronym{sgd}{SGD}{stochastic gradient descent}
\newacronym{bcd}{BCD}{block coordinated descent}
\newacronym{dkw}{DKW}{Dvoretzky-Kiefer-Wolfowitz}
\newacronym{ucl}{UCL}{Upper Confidence Limit}
\newacronym{dnn}{DNN}{deep neural network}
\newacronym{svm}{SVM}{support vector machine}
\newacronym{rbfnet}{RBFNet}{radial basis function network}
\newacronym{bmi}{BMI}{body mass index}
\newacronym{heloc}{HELOC}{home equity line of credit}
\newacronym{compas}{COMPAS}{correctional offender management profiling for alternative sanctions)}
\newacronym{ares}{AReS}{actionable resource summaries}
\newacronym{globe}{GLOBE}{global \& efficient counterfactual explanations}
\newacronym{dice}{DiCE}{diverse counterfactual explanations}

\usepackage{environ} 
\providecolor{dtured}{rgb}{0.60, 0.00, 0.00} %

\newcounter{ineq}
\newcommand{\ineq}[1]{\stepcounter{ineq}\stackrel{(\roman{ineq})}{#1}}

\begin{document}

%

%
\runningauthor{Lei You, Lele Cao, Mattias Nilsson, Bo Zhao, Lei Lei}

\twocolumn[

\aistatstitle{Distributional Counterfactual Explanations With Optimal Transport}

\aistatsauthor{ Lei You \textnormal{\textsuperscript\Letter}  \And Lele Cao \And  Mattias Nilsson} 
\aistatsaddress{Technical University of Denmark \And  King (Part of Microsoft) \And NekoHealth}

\aistatsauthor{ Bo Zhao \And Lei Lei }
\aistatsaddress{  Aalto University \And Xi'an Jiaotong University }

]

\begin{abstract}
Counterfactual explanations (CE) are the de facto method for providing insights into black-box decision-making models by identifying alternative inputs that lead to different outcomes. However, existing CE approaches, including group and global methods, focus predominantly on specific input modifications, lacking the ability to capture nuanced distributional characteristics that influence model outcomes across the entire input-output spectrum.
This paper proposes distributional counterfactual explanation (DCE), shifting focus to the distributional properties of observed and counterfactual data, thus providing broader insights. DCE is particularly beneficial for stakeholders making strategic decisions based on statistical data analysis, as it makes the statistical distribution of the counterfactual resembles the one of the factual when aligning model outputs with a target distribution\textemdash something that the existing CE methods cannot fully achieve.
We leverage optimal transport (OT) to formulate a chance-constrained optimization problem, deriving a counterfactual distribution aligned with its factual counterpart, supported by statistical confidence. The efficacy of this approach is demonstrated through experiments, highlighting its potential to provide deeper insights into decision-making models.
\end{abstract}

\section{Background}

In the field of \gls{xai}, \gls{ce} has become the quintessential method for providing insight and explainability in complex decision-making models. This prominence is particularly marked in domains where understanding the causal impact of variables is pivotal for informed decision-making \citep{verma2020counterfactual}. The utility of \glsplural{ce} is grounded in their ability to answer \emph{``what-if'' scenarios}, offering tangible insights into how slight alterations in input can lead to different model outputs. This not only aids in demystifying the decision process of black-box models but also helps in identifying potential biases and areas for model improvement.

\begin{figure}
    \centering
    \scalebox{1.0}{\input{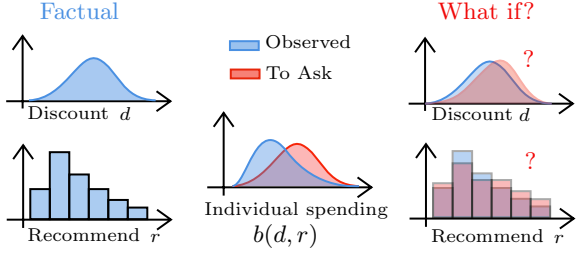}}
    \caption{\small  Consider a retail business aiming to understand how a machine learning model $b$ predicts individual customer spending for revenue forecasting, based on discount rates $d$ and product  recommendations $r$. Based on any \textcolor[RGB]{86,151,228}{observed model outputs}, the business asks to understand how adjusting the entire distributions of discounts and recommendations would impact the prediction \textcolor[RGB]{246,37,9}{towards another pattern}. Such insights are crucial for 1) making strategic operational decisions and 2) verifying whether the model behavior aligns with the real-world causal relationships. This calls for \gls{dce}, which allows the business to find explanations to the model at the distribution level\textemdash not just individual instances. To answer the ``what-if'' question, the \textcolor[RGB]{246,37,9}{counterfactual $d$ and $r$} that the business seeks should resemble the \textcolor[RGB]{86,151,228}{observed factual $d$ and $r$}. This is because drastic deviations from current practices may be impractical, and similarity ensures actionable recourse. Traditional \gls{ce}, including group or global method, fail in this aspect.}
    \label{fig:dce_example}%
\end{figure}

\paragraph{Motivation and Challenges} Traditional \gls{ce} focuses on individual data points, analyzing how changes in specific features influence the model's output. While this approach is useful, it overlooks scenarios where the statistical distribution of counterfactuals should resemble with that of the original instances \citep{carrizosa2024mathematical}. See \figurename~\ref{fig:dce_example} for an example. It is important to note that group \gls{ce} does not bridge this gap because they neglect the statistical properties' alignment between the factual and the counterfactual. Whereas \gls{dce} operates at the distributional level, where the entire input and output of a model are treated as distributions. Since strategical business decisions are made based on trends and patterns rather than single instance, this approach could practically benefit more for the stakeholders making strategic decisions based on statistical data analysis. Related work of this paper is detailed in Appendix~\ref{sec:related_work}.

Despite the need for \gls{dce}, several challenges arise in it. Metrics exist to quantify differences between distributions, but practical interpretation of these differences is often unclear. Unlike traditional \gls{ce} or group \gls{ce}, which adjust individual feature values, \gls{dce} must extend its mechanism from modifying ``quantities'' to adapting ``quantiles,'' fitting naturally within the data science paradigm of using quantile plots and histograms. Furthermore, determining differences between distributions involves sampling, introducing challenges like sampling error, increased sample complexity, and the curse of dimensionality\textemdash all of which must be addressed to ensure accurate and computationally feasible counterfactual generation.

\paragraph{Main Contribution} We propose a novel approach to \gls{dce} using \gls{ot}, commonly known as Wasserstein distance, along with its sliced version to quantify divergence between distributions. By formulating \gls{dce} as a chance-constrained optimization problem, we derive distributional counterfactuals that not only remain close to the original data but also ensure alignment with a predefined target distribution. Our method incorporates statistically rigorous confidence intervals within the optimization process, providing guarantees on the validity and proximity of  counterfactuals. The proposed optimization algorithm effectively balances the trade-off between maintaining input similarity and fulfilling counterfactual requirements, with proven convergence rates. This approach establishes a statistically sound link between \gls{ot} theory and \gls{ce}, advancing the capability to understand model behavior at a distributional level, and is demonstrated to be highly effective in practical applications.

\section{Prelimenaries and Problem Setup}
\label{sec:problem}

We first give a brief introduction of \gls{ot} and its sliced version. Then we formulate the \gls{dce} problem as a chance constrained optimization problem based on \gls{ot}.

\paragraph{\Gls{ot} and the Wasserstein Distance} Denote by $\P(\mathbb{R})$ the set of probability measures with finite second moments. Let $\gamma_1, \gamma_2 \in \mathbb{R}$, and let $\Pi(\gamma_1, \gamma_2)$ denote the set of all joint probability measures on $\mathbb{R} \times \mathbb{R}$ with marginals $\gamma_1$ and $\gamma_2$. We refer to $\Pi(\gamma_1, \gamma_2)$ as the set of transport plans between $\gamma_1$ and $\gamma_2$. The \gls{1d} squared $2$-Wasserstein distance is defined as
\begin{equation}
\ot(\gamma_1, \gamma_2) \triangleq \inf_{\pi\in\Pi}  \int_{\RR\times \RR} \norm{a_1-a_2}^2 \d{\pi(a_1,a_2)},
\label{eq:ot}
\end{equation}
which represents the \gls{ot} cost between $\gamma_1$ and $\gamma_2$ with respect to the squared Euclidean distance $\norm{a_1 - a_2}^2$ under the optimized \gls{ot} plan $\pi$.

\paragraph{Sliced Wasserstein Distance} Different from its original version, the sliced Wasserstein  distance  performs \gls{1d} linear projections of two high-dimensional measures $\bm{\gamma}_1$ and $\bm{\gamma}_2$, then it computes the Wasserstein distance between the projected measures. Formally, let $\SS^{d-1}\subset\RR^d$ stand for the $d$-dimensional unit sphere: 
\begin{equation}
\sot(\bm{\gamma}_1, \bm{\gamma}_2) \triangleq \int_{\SS^{d-1}} \ot(\bm{\theta} \sharp \bm{\gamma}_1, \bm{\theta} \sharp \bm{\gamma}_2)\d{\sigma(\bm{\theta})}, 
\label{eq:eot}
\end{equation}
where $\sigma$ is the uniform distribution on $\SS^{d-1}$ and $\sharp$ is the push-forward operator, which projects (high-dimensional) measrues $\bm{\gamma}_1$ and $\bm{\gamma}_2$ onto a \gls{1d} space.

We choose $\sot$ over other approximations of $\ot$ because it retains the same sample complexity as $\ot$, enables the derivation of confidence intervals crucial for the DCE framework, and provides an intuitive, quantile-based interpretation for comparing distributions, as shown by the next section.

\paragraph{\Gls{dce} Problem Formulation} Denote $b:\RR^d\rightarrow \RR$ a black-box model. Given any data distribution $\vec{x}'$ ($\vec{x}'\in\RR^d$) as a factual (i.e. an observed distribution) input, denote by $y'=b(\vec{x}')$ the factual model output distribution\footnote{We use the bold $\vec{y}$ to represent its empirical distribution version $[y_1,\ldots,y_n]$ ($\forall y_i\in\RR$).}. Denote by $y^*$ ($y^*\in\RR$) a target distribution, and $y^*\neq y'$. 
The goal is to find a counterfactual \( \vec{x} \) that closely matches \( \vec{x}' \), so that the resulting model output distribution \( b(\vec{x}) \) closely approximates the target distribution \( y^* \). In practical applications, closed-form expressions for these distributions are rarely available, so empirical distributions are used instead. However empirical distributions may contain outliers, we hereby align \( \vec{x} \) with \( \vec{x}' \) and \( b(\vec{x}) \) with \( y^* \) probabilistically and hence \gls{dce} is formulated as a chance-constrained optimization problem, shown in \eqref{eq:main_problem}.
\begin{subequations}
\hspace{-1cm}
\begin{align}
[\textcolor{darkblue}{\text{DCE Problem}}]\quad  & \max_{\vec{x}, \objvar} \objvar \\
 \text{s.t.} \quad & \objvar \leq  \PP\left[\sot(\vec{x},\vec{x}')< U_{x}\right] \label{eq:main_problem_C1} \\
                   & \objvar \leq  \PP\left[\ot(b(\vec{x}), y^*)< U_{y}\right] \label{eq:main_problem_C2} \\
                   & \objvar \geq 1-\frac{\alpha}{2}
\end{align}
\label{eq:main_problem}%
\end{subequations}
The inequality \eqref{eq:main_problem_C1} imposes that the sliced Wasserstein distance between the multi-dimensional $\vec{x}'$ and its counterfactual distribution $\vec{x}$ is subject to a chance constraint such that the distance is smaller than a threshold $U_{x}$ with a probability no lower than $\objvar$. The inequality \eqref{eq:main_problem_C2} is for the same goal but for the model output\footnote{This formulation generalizes to multi-dimensional output by using sliced wasserstein distance in \eqref{eq:main_problem_C2}. The method and conclusions proposed in this paper still hold.}. We set a confidence level $\alpha/2$, such that a feasible solution needs to make both chance constraints hold beyond the level probabilistically. 

Revisiting the example in \figurename~\ref{fig:dce_example}, \eqref{eq:main_problem_C1} guarantees that the counterfactual joint distribution of discount rates and product recommendations ensemble the factual one. And \eqref{eq:main_problem_C2} makes sure the counterfactual achieves the ``To Ask'' model output $y^*$. Maximizing $P$ ensures that the probability of both the input and output distributions meeting their respective constraints is as high as possible, with a confidence level for at least $1-\frac{\alpha}{2}$.

\section{Theoretical Foundations of \gls{dce}}
\label{sec:foundations}

This sections shows useful properties of the formulation~\eqref{eq:main_problem}. First, \gls{dce} extends \gls{ce} from quantity comparison to quantile comparison.  Second, the slicing approximation in dealing with high-dimensional data does not sacrifice sample complexity. Third, statistical confidence intervals are derived for the chance constraints \eqref{eq:main_problem_C1} and \eqref{eq:main_problem_C2}. Otto \citep{otto2001geometry} showed that the space metrized by the $2$-Wasserstein distance admits the structure of a formal Riemannian metric. The properties enable solving the \gls{dce} problem via Riemannian \gls{bcd}.

\paragraph{Quantile-Based Expression}  The Wasserstein distance in~\eqref{eq:main_problem_C2} can alternatively be written as follows (with $y = b(\vec{x})$).
\begin{align}
\ot(y, y^*) & = \int_{0}^{1} \norm{F^{-1}_{y}(q) - F^{-1}_{y^*}(q)}^2 \d{q} \label{eq:y-quantile}
\end{align}
In the equation above, $F^{-1}_y(q)$ and $F^{-1}_{y^*}(q)$ are the inverse cumulative distribution functions (i.e. quantile functions) of $y$ and $y^{*}$.
The equation holds due to both $y$ and $y^*$ being \gls{1d} distributions, providing an alternative explanation of the Wasserstein distance: It is the integral of the square of the difference between the corresponding quantiles in the interval $[0,1]$. Essentially, this aligns the model output distribution $y$ ($y=b(\vec{x})$) and the target distribution $y^*$ in a way that we are comparing the location of equivalent ``percentiles'' of mass in each distribution. In fact, it accounts for the full range of variability in the outputs, not just their central tendency (e.g. in contrast to comparing $\EE(y)$ and $\EE(y^*)$ only), which is crucial for many practical applications where distributional characteristics are important.

\paragraph{Sample Complexity of Slicing Approximation}
In \eqref{eq:main_problem}, the dissimilarity between \(\vec{x}\) and \(\vec{x}'\) is evaluated using the sliced Wasserstein distance, which serves as an approximation of their true Wasserstein distance. In real-world applications, we typically have finite samples drawn from the underlying distributions of \(\vec{x}\) and \(\vec{x}'\). The effectiveness of this approximation hinges on its ability to provide accurate distance estimates using a limited number of samples, a property known as ``good sample complexity.'' A natural question arises: Does the slicing approximation possess the same sample complexity as its original counterpart? We state \citet[Theorem 4]{nadjahi2020statistical} below as Proposition~\ref{prop:sample_complexity}. 

\begin{proposition}
\label{prop:sample_complexity}
Consider any measure \( z \in \mathcal{P}(\mathbb{R}) \) with an empirical measure \( \hat{\vec{z}} = \{z_i\}_{i=1}^n \), and a measure \( z' \in \mathcal{P}(\mathbb{R}) \) with an empirical measure \( \hat{\vec{z}}' = \{z'_j\}_{j=1}^n \). Let \( \xi(n) \) be any function of the sample size \( n \). Suppose that the squared Wasserstein distance \( \ot \) satisfies the following sample complexity:
\[
\mathbb{E}\left| \ot(z, z') - \ot(\hat{\vec{z}}, \hat{\vec{z}}') \right| \leq \xi(n).
\]
Then, for any measures \( \vec{x}, \vec{x}' \in \mathcal{P}(\mathbb{R}^d) \) with corresponding empirical measures \(\hat{\vec{x}} = \{\vec{x}_i\}_{i=1}^n \) and \( \hat{\vec{x}}' = \{\vec{x}'_i\}_{i=1}^n \), the sliced Wasserstein distance \( \sot \) satisfies
\[
\mathbb{E}\left| \sot(\vec{x}, \vec{x}') - \sot(\hat{\vec{x}}, \hat{\vec{x}}') \right| \leq \xi(n).
\]
\end{proposition}
Proposition~\ref{prop:sample_complexity} asserts that the sample complexity of the slicing approximation is directly proportional to that of the original Wasserstein distance. Notably, the sample complexity is independent of the dimensionality \( d \). Consequently, formulation~\eqref{eq:main_problem} scales effectively to high-dimensional inputs.

\paragraph{\gls{ucl}}
Let \( F^{-1} \) denote the quantile function of a distribution, and \( F_{n}^{-1} \) its empirical counterpart based on \( n \) samples. When specifying the distribution for a specific variable, the letter of the variable is put as subscript of them. We define some sequences functions  that can be used for any quantile function, which are used  construct confidence intervals (and hence \gls{ucl}). That is, a confidence interval is represented below
\begin{align}
&  \mathbb{P}\left( F_n^{-1}\left( \underline{q}_{\alpha, n}(u) \right) \leq F^{-1}(u)  \leq F_n^{-1}\left( \overline{q}_{\alpha, n}(u) \right) \right) \nonumber  \\ 
& \geq 1 - \frac{\alpha}{2},
    \label{eq:alpha_over_2N}
\end{align}
for some sequences of functions $\underline{q}_{\alpha, n}:(0,1)\rightarrow \mathbb{R}$ and $\overline{q}_{\alpha, n}:(0,1)\rightarrow \mathbb{R}$. We use the variable $u$ to refer to the quantile level, and the sequences $\underline{q}_{\alpha, n}$ and $\overline{q}_{\alpha, n}$ serve to transform this quantile level $u$ into a real number. Theorem~\ref{thm:interval} below follows \citep[Propositions 5 and 6]{manole2022minimax} (proof sketched in Appendix~\ref{sec:proof_thm_interval}) providing uniform confidence levels for \( \sot(\vec{x},\vec{x}') \) and \( \ot(b(\vec{x}),y^*) \).

\begin{theorem}
Let \( \delta \in (0, 1/2) \) be a trimming constant, the following inequalities hold:

1. For $\ot(b(\vec{x}),y^*)$ in \eqref{eq:main_problem_C2}, we have
\begin{align}
    &\!\!\!\!\mathbb{P}\left[ \ot(b(\vec{x}),y^*) \leq \frac{1}{1 - 2\delta} \!\int_{\delta}^{1 - \delta} \! D(u) \, \dd u \right] \geq 1 - \frac{\alpha}{2},
    \label{eq:int_D1}
\end{align}
\begin{align}
   &\text{where } D(u) \triangleq  \max\bigg\{  F_{y,n}^{-1}\left( \overline{q}_{\alpha, n}(u) \right) - F_{y^{*}, n}^{-1}\left( \underline{q}_{\alpha, n}(u) \right), \nonumber \\
    & F_{y^{*}, n}^{-1}\left( \overline{q}_{\alpha, n}(u) \right) - F_{y, n}^{-1}\left( \underline{q}_{\alpha, n}(u) \right) \bigg\}.
    \label{eq:D1}
\end{align}

2. For  \( \sot(\vec{x},\vec{x}') \) in \eqref{eq:main_problem_C1}, let the projection vectors \( \bm{\theta}_1, \ldots, \bm{\theta}_N \) be independent and identically distributed samples from a distribution \( \sigma \) on the unit sphere \( \mathbb{S}^{d-1} \). Let \( \sigma_N \) denote the empirical measure associated with these samples. Then,
\begin{align}
    &\mathbb{P}\left[ \sot(\vec{x},\vec{x}') \leq  \right. \nonumber\\
    &\left.\frac{1}{1 - 2\delta} \int_{\mathbb{S}^{d-1}}\!\!\int_{\delta}^{1 - \delta} \!\!D_{\bm{\theta},N}(u) \, \dd u \, \dd\sigma_N(\bm{\theta}) \right] \geq 1 - \frac{\alpha}{2}
    \label{eq:int_DN}
\end{align}
where
\begin{align}
    &D_{\bm{\theta},N}(u)\triangleq \max \bigg\{ F_{\bm{\theta}^\top \vec{x}, n}^{-1}\left( \overline{q}_{\alpha, n}(u) \right) 
        - F_{\bm{\theta}^\top \vec{x}', n}^{-1}\left( \underline{q}_{\alpha, n}(u) \right), \notag \\
    &  F_{\bm{\theta}^\top \vec{x}', n}^{-1}\left( \overline{q}_{\alpha, n}(u) \right) 
    - F_{\bm{\theta}^\top \vec{x}, n}^{-1}\left( \underline{q}_{\alpha, n}(u) \right)
    \bigg\}.
    \label{eq:DthetaN}
\end{align}
Here, \( F_{\bm{\theta}^\top \vec{x}, n}^{-1} \) denotes the empirical quantile function of  \( \bm{\theta}^\top \vec{x}_i \) for \( i = 1, \dots, n \), and similarly for \( F_{\bm{\theta}^\top \vec{x}', n}^{-1} \).

\label{thm:interval}
\end{theorem}
 The inequalities \eqref{eq:int_D1} and \eqref{eq:int_DN} provide methods to validate and substantiate the chance constraints \eqref{eq:main_problem_C2} and \eqref{eq:main_problem_C1}, respectively. The right-hand-side of \eqref{eq:main_problem_C2} is no less than $1-\alpha/2$ if and only if we have its corresponding \gls{ucl}, denoted by $\overline{\ot}$, no larger than $U_{y}$, i.e.
\begin{equation}
\overline{\ot}\triangleq  \frac{1}{1-2\delta} \int_{\delta}^{1-\delta} D(u)\dd u  \leq U_{y}.
\label{eq:overline_Q_nu}
\end{equation}
The computation of \eqref{eq:overline_Q_nu} lies on $D(u)$, defined in \eqref{eq:D1}, which quantifies the disparity between the quantile functions of $y=b(\vec{x})$ and $y^*$ by utilizing pre-specified function sequences $\underline{q}_{\alpha, n}(u)$ and $\overline{q}_{\alpha, n}(u)$. Practically, the quantile level $u$ can be uniformly sampled within $(0,1)$ and the remaining critical task is to formulate suitable $\underline{q}_{\alpha, n}$ and $\overline{q}_{\alpha, n}$ functions that are grounded in statistical theory\footnote{One classic example is $\underline{q}_{\alpha, n}=u-\beta_n$,  $\overline{q}_{\alpha, n}=u+\beta_n$, and $\beta_n=\sqrt{\log(4/\alpha)/2n}$. Then \eqref{eq:overline_Q_nu} holds by \gls{dkw} inequalities \citep{dvoretzky1956asymptotic}. One could refer to \citep[Section 4]{manole2022minimax} for more examples. This transformation adheres to the principles of uniform quantile bounds, a topic extensively discussed in the work of \citep{shorack2009empirical}.}.

Similarly, the right-hand-side of \eqref{eq:main_problem_C1} is no less than $1-\alpha/2$ if and only if we have its corresponding \gls{ucl}, denoted by $\overline{\sot}$, follows
\begin{equation}
\!\!\!\!\overline{\sot}\!\triangleq \!\frac{1}{1-2\delta}\!\int_{\SS^{d-1}}\!\!\int_{\delta}^{1-\delta}\!\!\!D_{\bm{\theta}, N}(u)\dd u\dd\sigma_N(\theta)  \leq U_{x},
\label{eq:overline_Q_mu}
\end{equation}
with $D_{\bm{\theta},N}$ defined in \eqref{eq:DthetaN}. Compared to $\overline{\ot}$, $\overline{\sot}$ aggregates the integral result on $D_{\bm{\theta}, N}(u)$ further on $\bm{\theta}$ that is distributed on the $d$-dimensional unit sphere\footnote{Remark $\overline{\ot}$ is finite sample inference whereas $\overline{\sot}$ is not. One needs to enlarge $\overline{\sot}$ with a factor that is inversely proportional to $N$ to obtain the finite version \citep{manole2022minimax}.}.

Alternatively, the \gls{clt} is applicable to both \gls{ot} and its sliced counterpart, allowing $\overline{\sot}$ and $\overline{\ot}$ to be estimated via the bootstrap method. For further information, refer to the theorems presented in \citep[Theorem 4.1]{del2019central} and \citep[Theorem 4]{manole2022minimax}. Regardless of the approaches, both chance constraints in the \gls{dce} formulation \eqref{eq:main_problem} are rigorously respected.

\section{Optimization for Problem Solving}
\label{sec:problem_solving}

\paragraph{Revisit the \gls{dce} Problem}

We re-write the \gls{dce} problem in \eqref{eq:main_problem} in a form for empirical distribution. Let $\vec{x}=\{\vec{x}_i\}_{i=1}^n$ and $\vec{y}^*=\{y^*_j\}_{j=1}^n$ be empirical distributions.
Denote the set of projection vectors generated from a uniform distribution on a unit sphere by $\Theta = \{\bm{\theta}_1,\bm{\theta}_2\ldots \bm{\theta}_N\}$.
Define
\begin{align}
Q_{x}(\vec{x}, \bm{\mu}) & \triangleq \sum_{k}^N\sum_{i=1}^{n}\sum_{j=1}^{n}\abs{\bm{\theta}^\top\vec{x}_i - \bm{\theta}^{\top}\vec{x}'_j}^2\mu^{(k)}_{ij}, \label{eq:Q_mu} \\ 
Q_{y}(\vec{x}, \bm{\nu}) & \triangleq   \sum_{i=1}^{n}\sum_{j=1}^{n}\abs{b(\vec{x}_i) - y^{*}_j}^2 \nu_{ij}. \label{eq:Q_nu}%
\end{align}
Both $Q_{x}$ and $Q_{y}$ are smooth in $\vec{x}$, $\bm{\mu}$ and $\bm{\nu}$. The empirical version of \eqref{eq:main_problem} reads:
\begin{subequations}
\begin{align}
 & \max_{\vec{x}\in\M, \objvar\geq 0} \objvar \\
 \text{s.t.} \quad & \objvar \leq   \PP_n \left[ \inf_{\bm{\mu}\in \Pi} Q_{x}(\vec{x},\bm{\mu})\leq U_{x} \right] \label{eq:main_problem_empirical-U_1}\\
                   & \objvar \leq   \PP_n \left[\inf_{\bm{\nu}\in \Pi} Q_{y}(\vec{x}, \bm{\nu})\leq U_{y}\right] \label{eq:main_problem_empirical-U_2}\\
                   & \objvar \geq 1-\alpha / 2 \label{eq:main_problem_empirical-alpha_2}
\end{align}
\label{eq:main_problem_empirical}%
\end{subequations}
Note that $Q_{x}$ and $Q_{y}$ are the empirical versions of \eqref{eq:ot} and \eqref{eq:eot}  (without $\inf$), respectively. That is, for any given $\vec{x}$, optimizing the \gls{ot} plan $\mu^{(k)}_{ij}$ towards minimizing $Q_{x}$, i.e. $ \inf_{\bm{\mu}\in \Pi} Q_{x}(\vec{x},\bm{\mu})$, yields $\sot(\vec{x}, \vec{x}')$. Similarly, $\inf_{\bm{\nu}\in \Pi} Q_{y}(\vec{x}, \bm{\nu})$ computes $\ot(b(\vec{x}),\vec{y}^*)$.


\paragraph{Riemannian \gls{bcd} Optimization} Initially, we established Theorem~\ref{thm:interval} to manage the chance constraints \eqref{eq:main_problem_empirical-U_1} and \eqref{eq:main_problem_empirical-U_2}. 
Secondly, we demonstrate below a partial optimality condition for resolving \eqref{eq:main_problem_empirical}, uncovering that achieving the optimum (or local optima) is contingent upon identifying an appropriate balance between $Q_{x}$ and $Q_{y}$. Thirdly, we introduce our algorithm \discount, which follows essentially a \gls{bcd} framework \citep{peng2023block, huang2021riemannian, gutman2023coordinate} and supports both discrete and continuous strategies for this balancing.  The discrete approach, termed \textit{Set Shrinking}, optimizes the weights by selecting from a finite set of candidate values. In contrast, the continuous method \textit{Interval Narrowing}, utilizes a bisection strategy applied to a progressively narrowing domain of variables. Finally, we remark that the proposed \discount~comes with a guarantee of convergence regardless of which method is chosen, elaborated in Section~\ref{sec:convergence}.

Let $\M$ be a compact smooth Riemannian manifold. Let $\vec{x}\in\M_1, \bm{\mu}\in\M_2, \bm{\nu}\in\M_3$ ($\M=\M_1\times\M_2\times\M_3$), where $\M_i$ ($i=1,2,3$) are bounded submanifolds.
Our optimization is around a function $Q(\vec{x},\bm{\mu}, \bm{\nu}, \eta)$:
\begin{equation}
Q(\vec{x},\bm{\mu}, \bm{\nu}, \eta) \triangleq  (1-\eta)\cdot Q_{x}(\vec{x},\bm{\mu}) +  \eta \cdot Q_{y}(\vec{x}, \bm{\nu}),
\label{eq:Q-eta}
\end{equation}
and the theorem below establishes the partial optimality condition for optimization.
\begin{theorem}
Assume that the optimization problem in \eqref{eq:main_problem_empirical} is feasible, i.e.  \eqref{eq:main_problem_empirical-U_1} and \eqref{eq:main_problem_empirical-alpha_2} can be achieved simultaneously with at least $1-\alpha/2$. Then, there exists a value \( \eta^* \in [0, 1] \) such that the solution \( \vec{x}^* \) obtained by optimizing \( Q(\vec{x}, \bm{\mu}, \bm{\nu} \mid \eta^*) \), defined as
\begin{equation}
  \vec{x}^{*} \triangleq \argmin_{\vec{x}, \bm{\mu}, \bm{\nu}} Q(\vec{x}, \bm{\mu}, \bm{\nu}, \eta^*),
  \label{eq:main_problem_empirical_alternative}
\end{equation}
is also the optimal solution to the original problem \eqref{eq:main_problem_empirical}, that is,
\[
\vec{x}^* = \argmin_{\vec{x}} \varphi \quad \text{s.t. } \eqref{eq:main_problem_empirical-U_1} - \eqref{eq:main_problem_empirical-alpha_2}.
\]
\label{thm:optimality}
\end{theorem}
The proof of Theorem~\ref{thm:optimality} is in Appendix~\ref{sec:proof_thm_optimality}. The theorem demonstrates addressing \eqref{eq:main_problem_empirical} (and hence the \gls{dce} problem in \eqref{eq:main_problem}) necessitates finding a proper $\eta$.

\begin{algorithm}[t]
\caption{\textbf{Dis}tributional \textbf{count}erfactual}
\begin{algorithmic}[1]
\REQUIRE $\vec{x}$, $\vec{y}^*$, model $b$, projections $\Theta$, bounds $U_{x},U_{y}$ and significance level $\alpha$.
\ENSURE Counterfactual $\vec{x}$ or $\varnothing$.
\STATE $\vec{x}^{0}\leftarrow \vec{x}' + \sigma$; $t\leftarrow 0$
\REPEAT
    \STATE $\bm{\mu}^{t}\leftarrow \argmin_{\bm{\mu}}Q_{x}(\vec{x}^t,\bm{\mu})$ \label{alg:riemannian-mu} 
    \STATE $\bm{\nu}^{t}\leftarrow \argmin_{\bm{\nu}}Q_{y}(\vec{x}^t,\bm{\nu})$ \label{alg:riemannian-nu}
    \STATE $\overline{\ot}\leftarrow$ Eq.~\eqref{eq:overline_Q_nu} \label{alg:riemannian-Q_nu} 
    \STATE $\overline{\sot}\leftarrow$ Eq.~\eqref{eq:overline_Q_mu} \label{alg:riemannian-Q_mu} 
    \STATE $\eta^{t}\leftarrow$ Algorithm~\ref{alg:eta_bisection} (or \ref{alg:eta_discrete} in Appendix~\ref{sec:eta_optimization})  \label{alg:riemannian-eta}
    \STATE $\renabla Q\leftarrow\renabla_{\vec{x}}Q\left(\vec{x},\bm{\mu}^{t}, \bm{\nu}^{t},\eta^{t}\right)$ \label{alg:riemannian-nabla_Q}
    \STATE $\vec{x}^{t+1}\leftarrow \retr({-\tau \renabla Q})$ \label{alg:riemannian-x^{t+1}}
    \STATE $t\leftarrow t + 1$
\UNTIL{$\norm{\vec{x}^{t+1}-\vec{x}^{t}}\leq \epsilon$}
\IF{$\overline{\sot}\leq U_{x}$ \textbf{and} $\overline{\ot}\leq U_{y}$}
    \STATE \textbf{return} $\vec{x}^{t+1}$
\ENDIF
\STATE \textbf{return} $\varnothing$
\end{algorithmic}
\label{alg:riemannian}%
\end{algorithm}

Algorithm~\ref{alg:riemannian} is designed following the framework of \gls{bcd}, which performs alternating optimization, where one side focuses on optimizing the counterfactual $\vec{x}$, and the \gls{ot} plans $\bm{\mu}$ and $\bm{\nu}$, and the other side searches for a good performed $\eta$.
First, Lines~\ref{alg:riemannian-mu} and \ref{alg:riemannian-nu} compute the \gls{ot} distances defined in \eqref{eq:Q_mu} and \eqref{eq:Q_nu} and obtain the \gls{ot} plans $\bm{\mu}$ and $\bm{\nu}$, respectively.  Second, in lines \ref{alg:riemannian-Q_nu} and \ref{alg:riemannian-Q_mu}, we compute the two corresponding \glsplural{ucl} $\overline{\ot}$ and $\overline{\sot}$ that are defined in \eqref{eq:overline_Q_nu} and \eqref{eq:overline_Q_mu}. 
Because $\vec{x}^t$ is updated in every iteration, both \glsplural{ucl} are re-computed per iteration. Third, we use the two \glsplural{ucl} to compute an $\eta$ as shown in line \ref{alg:riemannian-eta}. The computed $\eta$ influences the optimization direction of the counterfactual distribution $\{\vec{x}_i\}_{i=1}^n$ (shown by line \ref{alg:riemannian-nabla_Q}) to balance the satisfaction of the two chance constraints \eqref{eq:main_problem_empirical-U_1} and \eqref{eq:main_problem_empirical-U_2}, ensuring that both are adequately addressed in the solution. Fourth, line \ref{alg:riemannian-nabla_Q} computes the Riemannian gradient with respect to $\vec{x}$, denoted by $\renabla_{\vec{x}}Q$. Line~\ref{alg:riemannian-x^{t+1}} makes a descent step with step size $\tau$ using the retraction  denoted by $\retr$ (which is a mapping from the tangent space at any point on the manifold back onto the manifold itself \citep{absil2008optimization}) and the Riemannian gradient $\renabla_{\vec{x}}Q$.

\begin{algorithm}[t]
\caption{Interval Narrowing}
\begin{algorithmic}[1]
\REQUIRE  $\overline{\sot}$, $\overline{\ot}$, $U_{x}$,$U_{y}$, $[l,r]$, and $\kappa$ ($0<\kappa<1$)
\ENSURE $\eta$
\STATE $\eta \leftarrow$  Balance the gaps $U_{x}\!-\!\overline{\sot}$ and $U_{y}\!-\!\overline{\ot}$\label{alg:eta_bisection-eta}
\IF{ $\eta > (l+r)/2$ } \label{alg:eta_bisection-if}
    \STATE $l \leftarrow l + \kappa (r-l) $
\ELSE
    \STATE $r \leftarrow r - \kappa (r-l) $ 
\ENDIF
\STATE Save $[l,r]$ and $\kappa$ as the input for the next run \label{alg:eta_bisection-save}
\STATE \textbf{return} $\eta$
\end{algorithmic}
\label{alg:eta_bisection}
\end{algorithm}
Algorithm \ref{alg:eta_bisection} optimizes $\eta$ within a pre-defined interval $[l,r]$, shown by line \ref{alg:eta_bisection-eta}. See Appendix \ref{sec:eta_optimization} for how the balance is achieved. 
In each run, the interval gets narrowed by a small proportion of $\kappa$, following a bi-section strategy, shown by lines \ref{alg:eta_bisection-if}--\ref{alg:eta_bisection-save}. The parameter $\eta$ can also be optimized in a discrete manner, see Algorithm \ref{alg:eta_discrete} in Appendix \ref{sec:eta_optimization}.

\section{Convergence Rate Analysis}
\label{sec:convergence}

 To ease the presentation, we let $\vec{v} = [\vec{x}, \bm{\mu}, \bm{\nu}]$, and perform analysis for the optimization of $\vec{v}$ and $\eta$. Recall that $\vec{v}\in\M$ where $\M$ is a Riemannian manifold. Assume $b$ is continuously differentiable and Lipschitz smooth with respect to $\vec{x}$ amd constant $L$. Following \cite[Lemma 1]{peng2023block}, we have that 1) there exists $\rho>0$ such that $\|\retr_{\vec{x}}(\vec{s})-\vec{x}\|\leq \rho\|\vec{s}\|$ where $\vec{s}$ belongs to the tangent space associated with $\vec{x}$ ($\vec{x}\in\M_1$), and 2) there exists a constant $\bL$ (dependent on $L$, $\M_1$, and \retr) such that the difference between $Q(\vec{\retr_{\vec{x}}(\vec{s}), \bm{\mu}, \bm{\nu}}, \eta)$ and $Q(\vec{x}, \bm{\mu}, \bm{\nu}, \eta)$ is bounded by the Riemannian gradient of $\renabla_{\vec{x}}Q$ and $\bL$. Using the two constants $\rho$ and $\bL$, the convergence rate of \discount~is derived in 
Theorem \ref{thm:convergence_combined} below. 
\begin{theorem}
\label{thm:convergence_combined}
Let $\{\vec{v}^t\}_{t=0}^T$ denote the iterates of Algorithm~\ref{alg:riemannian} combined with either Algorithm~\ref{alg:eta_discrete} or Algorithm~\ref{alg:eta_bisection}, using a stepsize $\tau = \frac{1}{\bL}$. Define 
\begin{align*}
C  \triangleq  \sqrt{2\bL} + \rho L\cdot\sqrt{\frac{2}{\bL}} \text{ and } B \triangleq \sup_{\vec{v}}\{Q_{x}(\vec{v}),Q_{y}(\vec{v})\},
\end{align*}
where $B$ is ascertained finite with finite $b$. Then, the following convergence guarantees hold:
\begin{align*}
&\min_{t=0,1,\ldots,T} \norm{\tnabla Q(\vec{v}^t,\eta^t)} \\
&\leq C \left[ \frac{1}{T+1} \left( Q_{x}(\vec{v}^{0}) + Q_{y}(\vec{v}^{0}) + \Delta \right) \right]^{\frac{1}{2}},
\end{align*}
where
\[
\Delta = 
\begin{cases}
0, & \text{if Algorithm~\ref{alg:eta_discrete} is used}, \\
\frac{r - l}{\kappa} B, & \text{if Algorithm~\ref{alg:eta_bisection} is used}.
\end{cases}
\]
\end{theorem}
Briefly, the theorem states that the magnitude of the Riemannian gradient goes towards zero with the number of iterations. For Set Shrinking, the gradient is bounded by the values of $Q_{x}$ and $Q_{y}$ at the starting point. For Interval Narrowing, the upper bound of $Q_{x}$ and $Q_{y}$ matters on top of the size and narrowing rate of the interval. The proof is derived in Appendix~\ref{sec:convergence_proof}. 

\section{Numerical Results}
\label{sec:numerical}

\begin{table*}[t]
\centering
\caption{\small [\textit{HELOC (top table) and COMPAS (bottom table)}] The rows with considerably low coverage scores ($\leq 0.8$) are marked gray. In each column, we use bold text to highlight the row with the best performance, ignoring all gray ones,  as one can always gain good performance in cost and proximity by sacrificing coverage, losing comparability. The model test accuracy is stated. The experiments are averaged over 10 runs.}
\label{tab:baseline_heloc}
\begin{tabular}{ccc|cccccc|cc|c}
\toprule
                                                                            &                              &                              & \multicolumn{6}{c|}{Cost}                                                                                                                                                                                                                                                      & \multicolumn{2}{c|}{Proximity}                            &                                                                      \\ \cline{4-11}
                                                                            &                              &                              & \multicolumn{1}{c|}{}                                                                       & \multicolumn{5}{c|}{\% Difference at Percentiles}                                                                                                                                &                                    &                                     &                                                                      \\
\multirow{-3}{*}{Model}                                                     & \multirow{-3}{*}{Algo.}      & \multirow{-3}{*}{Cover.}       & \multicolumn{1}{c|}{\multirow{-2}{*}{\begin{tabular}[c]{@{}c@{}}AReS \\ Cost\end{tabular}}} & 0-15                               & 15-30                              & 30-70                              & 70-85                              & 85-100                       & \multirow{-2}{*}{OT}               & \multirow{-2}{*}{MMD}               & \multirow{-3}{*}{\begin{tabular}[c]{@{}c@{}}Time\\ (s)\end{tabular}} \\ \hline
                                                                            & {\color[HTML]{656565} AReS}  & {\color[HTML]{656565} 0.019} & \multicolumn{1}{c|}{{\color[HTML]{656565} 0.038}}                                           & {\color[HTML]{656565} 0.000}       & {\color[HTML]{656565} 0.187}       & {\color[HTML]{656565} 43.83}       & {\color[HTML]{656565} 0.321}       & {\color[HTML]{656565} 0.000} & {\color[HTML]{656565} 0.000}       & {\color[HTML]{656565} 0.000}        & {\color[HTML]{656565} 12}                                            \\
                                                                            & Globe                        & 0.962                        & \multicolumn{1}{c|}{\textbf{3.346}}                                                         & 101.6                              & 45.08                              & 515.8                              & 95.56                              & 230.5                        & 8.521                              & 0.039                               & \textbf{5.3}                                                         \\
                                                                            & {\color[HTML]{656565} DiCE}  & {\color[HTML]{656565} 0.796} & \multicolumn{1}{c|}{{\color[HTML]{656565} 18.41}}                                           & {\color[HTML]{656565} 100.0}       & {\color[HTML]{656565} 100.0}       & {\color[HTML]{656565} 326.0}       & {\color[HTML]{656565} 12.04}       & {\color[HTML]{656565} 30.68} & {\color[HTML]{656565} 0.324}       & {\color[HTML]{656565} 0.107}        & {\color[HTML]{656565} 7235}                                          \\
\multirow{-4}{*}{\begin{tabular}[c]{@{}c@{}}DNN\\ (74.9\%)\end{tabular}}    & Discount                          & \textbf{0.981}               & \multicolumn{1}{c|}{22.87}                                                                  & \textbf{7.801}                     & \textbf{8.183}                     & \textbf{265.0}                     & \textbf{9.422}                     & \textbf{6.325}               & \textbf{0.202}                     & \textbf{0.036}                      & 632                                                                  \\ \hline
                                                                            & {\color[HTML]{656565} AReS}  & {\color[HTML]{656565} 0.058} & \multicolumn{1}{c|}{{\color[HTML]{656565} 0.049}}                                           & {\color[HTML]{656565} 0.000}       & {\color[HTML]{656565} 1.002}       & {\color[HTML]{656565} 6.000}       & {\color[HTML]{656565} 0.418}       & {\color[HTML]{656565} 0.000} & {\color[HTML]{656565} 0.002}       & {\color[HTML]{656565} 0.001}        & {\color[HTML]{656565} 11}                                            \\
                                                                            & {\color[HTML]{656565} Globe} & {\color[HTML]{656565} 0.038} & \multicolumn{1}{c|}{{\color[HTML]{656565} 3.302}}                                           & {\color[HTML]{656565} 41.93}       & {\color[HTML]{656565} 28.28}       & {\color[HTML]{656565} 87.70}       & {\color[HTML]{656565} 57.29}       & {\color[HTML]{656565} 117.8} & {\color[HTML]{656565} 2.899}       & {\color[HTML]{656565} 0.039}        & {\color[HTML]{656565} 4.9}                                           \\
                                                                            & {\color[HTML]{656565} DiCE}  & {\color[HTML]{656565} 0.174} & \multicolumn{1}{c|}{{\color[HTML]{656565} 10.54}}                                           & {\color[HTML]{656565} 100.0}       & {\color[HTML]{656565} 100.0}       & {\color[HTML]{656565} 95.28}       & {\color[HTML]{656565} 4.257}       & {\color[HTML]{656565} 32.39} & {\color[HTML]{656565} 0.330}       & {\color[HTML]{656565} 0.148}        & {\color[HTML]{656565} 2735}                                          \\
\multirow{-4}{*}{\begin{tabular}[c]{@{}c@{}}RBFNet\\ (73.6\%)\end{tabular}} & Discount                          & \textbf{0.962}               & \multicolumn{1}{c|}{\textbf{16.22}}                                                         & \multicolumn{1}{l}{\textbf{8.461}} & \multicolumn{1}{l}{\textbf{21.65}} & \multicolumn{1}{l}{\textbf{185.5}} & \multicolumn{1}{l}{\textbf{36.89}} & \textbf{19.55}               & \multicolumn{1}{l}{\textbf{0.563}} & \multicolumn{1}{l|}{\textbf{0.039}} & \textbf{621}                                                         \\ \hline
                                                                            & {\color[HTML]{656565} AReS}  & {\color[HTML]{656565} 0.038} & \multicolumn{1}{c|}{{\color[HTML]{656565} 0.577}}                                           & {\color[HTML]{656565} 0.000}       & {\color[HTML]{656565} 0.000}       & {\color[HTML]{656565} 0.794}       & {\color[HTML]{656565} 0.935}       & {\color[HTML]{656565} 0.000} & {\color[HTML]{656565} 0.001}       & {\color[HTML]{656565} 0.000}        & {\color[HTML]{656565} 14}                                            \\
                                                                            & Globe                        & \textbf{1.000}               & \multicolumn{1}{c|}{\textbf{3.600}}                                                         & 155.1                              & 95.49                              & \textbf{152.2}                     & 34.98                              & 178.6                        & 13.72                              & 0.039                               & \textbf{4.8}                                                         \\
                                                                            & DiCE                         & \textbf{1.000}               & \multicolumn{1}{c|}{13.01}                                                                  & 100.0                              & 100.0                              & 636.7                              & 44.13                              & 30.36                        & 0.514                              & 0.130                               & 5767                                                                 \\
\multirow{-4}{*}{\begin{tabular}[c]{@{}c@{}}SVM \\ (75.0\%)\end{tabular}}    & Discount                          & \textbf{1.000}               & \multicolumn{1}{c|}{4.357}                                                                  & \textbf{5.591}                     & \textbf{14.07}                     & 256.6                              & \textbf{16.63}                     & \textbf{9.811}               & \textbf{0.342}                     & \textbf{0.036}                      & 244                                                                  \\ \bottomrule
\end{tabular}
\renewcommand{\arraystretch}{0.2} 
\begin{tabular}{@{}c@{}} \\ \end{tabular}
\renewcommand{\arraystretch}{1}
\begin{tabular}{ccc|cccc|cc|cc|c}
\toprule
                                                                            &                              &                                       & \multicolumn{6}{c|}{Cost}                                                                                                                                                                                                                                                                                                                                & \multicolumn{2}{c|}{Proximity}                                 &                                                                      \\ \cline{4-11}
                                                                            &                              &                                       & \multicolumn{1}{c|}{}                                                                       & \multicolumn{1}{c|}{}                                                                      & \multicolumn{4}{c|}{Num. Distribution Shift \%}                                                                                                               &                                       &                                       &                                                                      \\ \cline{6-9}
                                                                            &                              &                                       & \multicolumn{1}{c|}{}                                                                       & \multicolumn{1}{c|}{}                                                                      & \multicolumn{2}{c|}{Priors Count}                                             & \multicolumn{2}{c|}{Time Served}                                              &                                       &                                       &                                                                      \\
\multirow{-4}{*}{Model}                                                     & \multirow{-4}{*}{Algo.}      & \multirow{-4}{*}{Cover.}                & \multicolumn{1}{c|}{\multirow{-3}{*}{\begin{tabular}[c]{@{}c@{}}AReS \\ Cost\end{tabular}}} & \multicolumn{1}{c|}{\multirow{-3}{*}{\begin{tabular}[c]{@{}c@{}}Cat.\\ Diff.\end{tabular}}} & Mean                                  & \multicolumn{1}{c|}{Std}              & Mean                                  & Std                                   & \multirow{-3}{*}{OT}                  & \multirow{-3}{*}{MMD}                 & \multirow{-4}{*}{\begin{tabular}[c]{@{}c@{}}Time\\ (s)\end{tabular}} \\ \hline
                                                                            & {\color[HTML]{656565} AReS}  & {\color[HTML]{656565} 0.103}          & \multicolumn{1}{c|}{{\color[HTML]{656565} 6.162}}                                           & \multicolumn{1}{c|}{{\color[HTML]{656565} 0.241}}                                          & {\color[HTML]{656565} 79.20}          & {\color[HTML]{656565} 87.61}          & {\color[HTML]{656565} 82.49}          & {\color[HTML]{656565} 89.06}          & {\color[HTML]{656565} 0.130}          & {\color[HTML]{656565} 0.169}          & {\color[HTML]{656565} 10}                                            \\
                                                                            & {\color[HTML]{000000} Globe} & {\color[HTML]{000000} \textbf{1.000}} & \multicolumn{1}{c|}{{\color[HTML]{000000} \textbf{0.785}}}                                  & \multicolumn{1}{c|}{{\color[HTML]{000000} -}}                                              & {\color[HTML]{000000} -}              & {\color[HTML]{000000} -}              & {\color[HTML]{CB0000} 1729}           & {\color[HTML]{000000} -}              & {\color[HTML]{000000} 8.007}          & {\color[HTML]{000000} 0.220}          & {\color[HTML]{000000} \textbf{4.4}}                                  \\
                                                                            & {\color[HTML]{000000} DiCE}  & {\color[HTML]{000000} 0.939}          & \multicolumn{1}{c|}{{\color[HTML]{000000} 7.678}}                                           & \multicolumn{1}{c|}{{\color[HTML]{000000} 0.485}}                                          & {\color[HTML]{000000} 78.78}          & {\color[HTML]{000000} 70.11}          & {\color[HTML]{000000} \textbf{64.07}} & {\color[HTML]{000000} 80.59}          & {\color[HTML]{000000} 0.406}          & {\color[HTML]{000000} 0.369}          & {\color[HTML]{000000} 503}                                           \\
\multirow{-4}{*}{\begin{tabular}[c]{@{}c@{}}DNN\\ (67.1\%)\end{tabular}}    & {\color[HTML]{000000} Discount}   & {\color[HTML]{000000} 0.976}          & \multicolumn{1}{c|}{{\color[HTML]{000000} 4.716}}                                           & \multicolumn{1}{c|}{{\color[HTML]{000000} \textbf{0.156}}}                                 & {\color[HTML]{000000} \textbf{59.69}} & {\color[HTML]{000000} \textbf{8.999}} & {\color[HTML]{000000} 85.29}          & {\color[HTML]{000000} \textbf{20.88}} & {\color[HTML]{000000} \textbf{0.130}} & {\color[HTML]{000000} \textbf{0.116}} & {\color[HTML]{000000} 125}                                           \\ \hline
                                                                            & {\color[HTML]{656565} AReS}  & {\color[HTML]{656565} 0.143}          & \multicolumn{1}{c|}{{\color[HTML]{656565} 2.345}}                                           & \multicolumn{1}{c|}{{\color[HTML]{656565} 0.040}}                                          & {\color[HTML]{656565} 0.000}          & {\color[HTML]{656565} 0.000}          & {\color[HTML]{656565} 0.000}          & {\color[HTML]{656565} 0.000}          & {\color[HTML]{656565} 0.020}          & {\color[HTML]{656565} 0.025}          & {\color[HTML]{656565} 9.3}                                           \\
                                                                            & {\color[HTML]{656565} Globe} & {\color[HTML]{656565} 0.690}          & \multicolumn{1}{c|}{{\color[HTML]{656565} 2.827}}                                           & \multicolumn{1}{c|}{{\color[HTML]{656565} 0.000}}                                          & {\color[HTML]{656565} 162.2}          & {\color[HTML]{656565} 0.000}          & {\color[HTML]{656565} 226.3}          & {\color[HTML]{656565} 0.000}          & {\color[HTML]{656565} 0.333}          & {\color[HTML]{656565} 0.133}          & {\color[HTML]{656565} 4.1}                                           \\
                                                                            & {\color[HTML]{656565} DiCE}  & {\color[HTML]{656565} 0.001}          & \multicolumn{1}{c|}{{\color[HTML]{656565} 6.653}}                                           & \multicolumn{1}{c|}{{\color[HTML]{656565} 0.297}}                                          & {\color[HTML]{656565} 46.99}          & {\color[HTML]{656565} 67.96}          & {\color[HTML]{656565} 49.47}          & {\color[HTML]{656565} 78.10}          & {\color[HTML]{656565} 0.269}          & {\color[HTML]{656565} 0.270}          & {\color[HTML]{656565} 544}                                           \\
\multirow{-4}{*}{\begin{tabular}[c]{@{}c@{}}RBFNet\\ (66.9\%)\end{tabular}} & Discount                          & \textbf{0.912}                        & \multicolumn{1}{c|}{\textbf{6.158}}                                                         & \multicolumn{1}{c|}{\textbf{0.200}}                                                        & \textbf{163.88}                       & \textbf{11.71}                        & \textbf{54.26}                        & \textbf{0.118}                        & \textbf{0.441}                        & \textbf{0.207}                        & \textbf{433}                                                         \\ \hline
                                                                            & {\color[HTML]{656565} AReS}  & {\color[HTML]{656565} 0.184}          & \multicolumn{1}{c|}{{\color[HTML]{656565} 2.000}}                                           & \multicolumn{1}{c|}{{\color[HTML]{656565} 0.324}}                                          & {\color[HTML]{656565} 0.000}          & {\color[HTML]{656565} 0.000}          & {\color[HTML]{656565} 0.000}          & {\color[HTML]{656565} 0.000}          & {\color[HTML]{656565} 0.015}          & {\color[HTML]{656565} 0.019}          & {\color[HTML]{656565} 9.2}                                           \\
                                                                            & Globe                        & \textbf{1.000}                        & \multicolumn{1}{c|}{\textbf{1.464}}                                                         & \multicolumn{1}{c|}{0.000}                                                                 & -                                     & -                                     & {\color[HTML]{CB0000} 3704}           & -                                     & 30.75                                 & 0.226                                 & \textbf{4.2}                                                         \\
                                                                            & DiCE                         & 0.974                                 & \multicolumn{1}{c|}{5.004}                                                                  & \multicolumn{1}{c|}{\textbf{0.172}}                                                        & \textbf{49.48}                        & 68.79                                 & \textbf{51.85}                        & 79.28                                 & 0.230                                 & 0.228                                 & 693                                                                  \\
\multirow{-4}{*}{\begin{tabular}[c]{@{}c@{}}SVM\\ (64.7\%)\end{tabular}}    & Discount                          & 0.983                                 & \multicolumn{1}{c|}{4.758}                                                                  & \multicolumn{1}{c|}{\textbf{0.172}}                                                        & 53.80                                 & \textbf{1.649}                        & 53.36                                 & \textbf{1.344}                        & \textbf{0.141}                        & \textbf{0.106}                        & 138                                                                  \\ \bottomrule
\end{tabular}
\end{table*}

We validate the concept of \gls{dce} with four datasets, HELOC \citep{holter2018fico}, COMPAS \citep{jeff2016we}, German Credit \citep{misc_statlog_(german_credit_data)_144}, and Cardiovascular Disease \citep{7qm5-dz13-20}. Three models are considered:  \gls{dnn}, \gls{rbfnet}, and \gls{svm}. Throughout our experiments\footnote{The code is available on \faGithub \\  \url{https://github.com/youlei202/distributional-counterfactual-explanation}}, $\alpha$ is set to $0.1$. This setting implies that the significance level for testing the distinctness of two distributions is $5\%$ as $\alpha/2$ corresponds to the one-tailed significance threshold. Quantitative experiments are performed in HELOC and COMPAS, justifying the advantage of \gls{dce} (solved by \discount) over existing \gls{ce} methods: \gls{dice} \citep{mothilal2020explaining}, \gls{ares} \citep{rawal2020beyond}, and \gls{globe} \citep{ley2023globe}. \gls{dice} finds counterfactuals for each factual instance and is hence an instance-based \gls{ce} method. Both \gls{ares} and \gls{globe} are group-based \gls{ce}. The main experiments focus on the validity and proximity of \gls{dce} in the distributional setup, as well as algorithm runtime and convergence. Extended numerical results are shown in Appendix~\ref{sec:numerical_extended}, focus on data diversity and counterfactual reasoning.

The effectiveness of \discount~is demonstrated via five major metrics: \textit{coverage, cost, proximity, runtime, and diversity}. The comparison of the first four metrics is shown in \tablename~\ref{tab:baseline_heloc}. Diversity is shown in Appendix~\ref{sec:numerical_extended}. 
Specifically, coverage evaluates how well the counterfactual output distribution approximates the target output distribution. Cost evaluates the payoff on changing the factual input to the counterfactual  for higher coverage. Proximity measures how well the counterfactual distribution resembles the factual one. Essentially, any \gls{ce} method plays trade-offs among coverage, cost, and proximity.

\paragraph{\discount~achieves the best coverage-cost balance with moderate runtime} The experiments in \tablename~\ref{tab:baseline_heloc} are performed on the \textit{HELOC} dataset \citep{holter2018fico} (top table) and \textit{COMPAS} dataset \citep{jeff2016we} (bottom table). The setup of the experiments are as follows. We train models for binary classifications on the two datasets. Then for each model, we sample an observed distribution $\vec{x}'$ from the test set and keep the data points with $\vec{y}'=b(\vec{x}')$ being $\vec{0}$. Every explainer is expected to find a counterfactual distribution $\vec{x}$ such that $\vec{y}=b(\vec{x})$ is as close as possible as a distribution to $\vec{y}^*=\vec{1}$. The metric coverage is defined to be the proportion of the $1$ values in $\vec{y}$. The \gls{ares} cost metric \citep{rawal2020beyond} bins continuous features and it is used to evaluate the cost of moving between two adjacent bins, while its counterpart columns evaluate feature changes in specific percentiles. Namely, the smaller changes, the fewer cost caused by the counterfactual. The distributional proximity between factual and counterfactual is measured by two different metrics that are commonly used for divergence between distributions: \gls{ot} and \gls{mmd}. 
\gls{ares} (rule-based) and \gls{globe} \citep{ley2023globe} (translation-based) are used to benchmark the \gls{ares} cost performance, as both algorithms are specifically designed for optimizing this cost metric. \Gls{dice} \citep{mothilal2020explaining} is used as a gradient-based baseline approach for our proposed \discount. Neither \gls{dice} nor \discount~has any knowledge to the \gls{ares} cost information. Gray rows in \tablename~\ref{tab:baseline_heloc} are not considered for comparison due to low ($\leq 0.8$) coverage, because it is trivial for any algorithm to sacrifice coverage for better performance in any other metric.

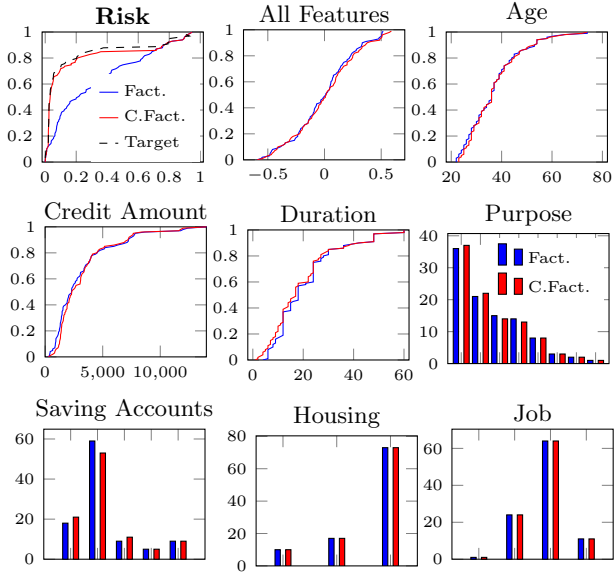
\begin{figure}[t]
    \centering
    \hspace{-3mm}
        \begin{tikzpicture}
    \begin{axis}[
    title={\textbf{Risk}},
    title style={yshift=-1.5ex,font=\small},
    xticklabel style={font=\tiny},
    yticklabel style={font=\tiny},
    legend style={
        draw=none,
        font=\tiny,
        at={(0.98,0.68)}, 
        legend image code/.code={
            \draw[mark repeat=2,mark phase=2]
                plot coordinates {
                    (0.0cm,0.0cm)
                    (0.25cm,0cm) 
                };
        },
    },
    xmin=-0.02, xmax=1.02,
    ymin=0, ymax=1,
    grid style=dashed,
    legend cell align={left},
    width=0.45\linewidth,
    height=0.4\linewidth,
    scaled x ticks=false,
    xticklabel style={/pgf/number format/fixed}
    ]
    
    \addplot[
        color=blue,
        ]
        coordinates {
 (0.000206,0.0) (0.000363,0.010101) (0.001001,0.020202) (0.001928,0.030303) (0.003202,0.040404) (0.003832,0.050505) (0.007581,0.060606) (0.007664,0.070707) (0.009124,0.080808) (0.009632,0.090909) (0.010335,0.10101) (0.014805,0.111111) (0.019517,0.121212) (0.026822,0.131313) (0.036625,0.141414) (0.038807,0.151515) (0.038896,0.161616) (0.045884,0.171717) (0.053714,0.181818) (0.054917,0.191919) (0.062517,0.20202) (0.062695,0.212121) (0.063506,0.222222) (0.068304,0.232323) (0.070695,0.242424) (0.072231,0.252525) (0.073494,0.262626) (0.073965,0.272727) (0.076062,0.282828) (0.078589,0.292929) (0.078857,0.30303) (0.086446,0.313131) (0.087687,0.323232) (0.09414,0.333333) (0.094927,0.343434) (0.096309,0.353535) (0.101598,0.363636) (0.102985,0.373737) (0.106526,0.383838) (0.120993,0.393939) (0.124236,0.40404) (0.13681,0.414141) (0.139389,0.424242) (0.145988,0.434343) (0.153896,0.444444) (0.159166,0.454545) (0.159771,0.464646) (0.174521,0.474747) (0.198398,0.484848) (0.201492,0.494949) (0.217658,0.505051) (0.22336,0.515152) (0.231583,0.525253) (0.260168,0.535354) (0.263799,0.545455) (0.264119,0.555556) (0.282232,0.565657) (0.303706,0.575758) (0.312769,0.585859) (0.326657,0.59596) (0.331893,0.606061) (0.342792,0.616162) (0.36759,0.626263) (0.375173,0.636364) (0.378536,0.646465) (0.384604,0.656566) (0.410081,0.666667) (0.413037,0.676768) (0.425459,0.686869) (0.425764,0.69697) (0.42998,0.707071) (0.454896,0.717172) (0.469166,0.727273) (0.477993,0.737374) (0.500605,0.747475) (0.540555,0.757576) (0.583748,0.767677) (0.613443,0.777778) (0.615904,0.787879) (0.627164,0.79798) (0.633162,0.808081) (0.64137,0.818182) (0.669707,0.828283) (0.683338,0.838384) (0.688328,0.848485) (0.709594,0.858586) (0.714846,0.868687) (0.75548,0.878788) (0.759045,0.888889) (0.759775,0.89899) (0.782674,0.909091) (0.793156,0.919192) (0.818434,0.929293) (0.88526,0.939394) (0.894962,0.949495) (0.897628,0.959596) (0.907074,0.969697) (0.91969,0.979798) (0.93435,0.989899) (0.949217,1.0)

        };
        \addlegendentry{Fact.}
    
    \addplot[
        color=red,
        ]
        coordinates {
 (0.000194,0.0) (0.000359,0.010101) (0.001015,0.020202) (0.001874,0.030303) (0.003196,0.040404) (0.003813,0.050505) (0.00761,0.060606) (0.007648,0.070707) (0.009343,0.080808) (0.009788,0.090909) (0.009795,0.10101) (0.015242,0.111111) (0.01882,0.121212) (0.019678,0.131313) (0.019688,0.141414) (0.019691,0.151515) (0.019695,0.161616) (0.019695,0.171717) (0.019704,0.181818) (0.019857,0.191919) (0.019948,0.20202) (0.020416,0.212121) (0.020465,0.222222) (0.02052,0.232323) (0.020699,0.242424) (0.020852,0.252525) (0.021711,0.262626) (0.021801,0.272727) (0.02187,0.282828) (0.021971,0.292929) (0.022751,0.30303) (0.022872,0.313131) (0.022872,0.323232) (0.023535,0.333333) (0.023908,0.343434) (0.02406,0.353535) (0.024105,0.363636) (0.024169,0.373737) (0.024231,0.383838) (0.024532,0.393939) (0.024567,0.40404) (0.025099,0.414141) (0.025427,0.424242) (0.025477,0.434343) (0.025858,0.444444) (0.028255,0.454545) (0.029312,0.464646) (0.029968,0.474747) (0.031175,0.484848) (0.031666,0.494949) (0.032357,0.505051) (0.032558,0.515152) (0.032866,0.525253) (0.034854,0.535354) (0.03683,0.545455) (0.03729,0.555556) (0.038646,0.565657) (0.040573,0.575758) (0.042775,0.585859) (0.046354,0.59596) (0.047721,0.606061) (0.050445,0.616162) (0.052174,0.626263) (0.052803,0.636364) (0.053951,0.646465) (0.056025,0.656566) (0.076553,0.666667) (0.082412,0.676768) (0.089272,0.686869) (0.093142,0.69697) (0.096816,0.707071) (0.099113,0.717172) (0.119727,0.727273) (0.126732,0.737374) (0.167848,0.747475) (0.167848,0.757576) (0.193805,0.767677) (0.194517,0.777778) (0.202492,0.787879) (0.203068,0.79798) (0.230664,0.808081) (0.261081,0.818182) (0.290197,0.828283) (0.332856,0.838384) (0.358406,0.848485) (0.7162,0.858586) (0.716988,0.868687) (0.749404,0.878788) (0.749849,0.888889) (0.765637,0.89899) (0.781138,0.909091) (0.78903,0.919192) (0.81789,0.929293) (0.886183,0.939394) (0.896284,0.949495) (0.898269,0.959596) (0.90685,0.969697) (0.917797,0.979798) (0.935408,0.989899) (0.947655,1.0)
        };
        \addlegendentry{C.Fact.}

\addplot[
    color=black,
    dashed,
]
coordinates {
    (0.000204,0.03) (0.000377,0.040101) (0.001066,0.050202) (0.001967,0.060303) 
    (0.003356,0.070404) (0.003998,0.080505) (0.00799,0.090606) (0.00798,0.100707) 
    (0.00981,0.110808) (0.00994,0.120909) (0.00989,0.13101) (0.015504,0.141111) 
    (0.019761,0.151212) (0.020661,0.161313) (0.020573,0.171414) (0.020696,0.181515) 
    (0.020680,0.191616) (0.020680,0.201717) (0.020789,0.211818) (0.020850,0.221919) 
    (0.020890,0.23202) (0.020936,0.242121) (0.021016,0.252222) (0.021048,0.262323) 
    (0.021734,0.272424) (0.021904,0.282525) (0.022797,0.292626) (0.022991,0.302727) 
    (0.022964,0.312828) (0.022980,0.322929) (0.023351,0.33303) (0.023728,0.343131) 
    (0.023776,0.353232) (0.024211,0.363333) (0.024103,0.373434) (0.024133,0.383535) 
    (0.024115,0.393636) (0.024163,0.403737) (0.024242,0.413838) (0.024648,0.423939) 
    (0.024524,0.43404) (0.025248,0.444141) (0.025520,0.454242) (0.025502,0.464343) 
    (0.025900,0.474444) (0.028468,0.484545) (0.029778,0.494646) (0.030466,0.504747) 
    (0.031734,0.514848) (0.031824,0.524949) (0.032375,0.535051) (0.032621,0.545152) 
    (0.032934,0.555253) (0.035097,0.565354) (0.037172,0.575455) (0.03776,0.585556) 
    (0.038578,0.595657) (0.041601,0.605758) (0.044914,0.615859) (0.048167,0.62596) 
    (0.048689,0.636061) (0.050967,0.646162) (0.052782,0.656263) (0.052993,0.666364) 
    (0.054149,0.676465) (0.056826,0.686566) (0.078381,0.696667) (0.086532,0.706768) 
    (0.093735,0.716869) (0.098799,0.72697) (0.101656,0.737071) (0.104069,0.747172) 
    (0.125713,0.757273) (0.133069,0.767374) (0.17624,0.777475) (0.17624,0.787576) 
    (0.203495,0.797677) (0.203747,0.807778) (0.212616,0.817879) (0.212222,0.82798) 
    (0.242197,0.838081) (0.273135,0.848182) (0.303707,0.858283) (0.349498,0.868384) 
    (0.375326,0.878485) (0.75201,0.888586) (0.752837,0.898687) (0.792874,0.908788) 
    (0.792341,0.918889) (0.803419,0.92899) (0.820195,0.939091) (0.7985,0.949192) 
    (0.859785,0.959293) (0.930002,0.969394) (0.910598,0.979495) (0.900382,0.989596) 
    (0.913192,0.999697) (0.923687,1.009798) (0.982178,1.019899) (0.994038,1.03)
};
\addlegendentry{Target}
\addlegendentry{Target}
    
    \end{axis}
    \end{tikzpicture}\hspace{-4mm}
        \begin{tikzpicture}
    \begin{axis}[
    title={All Features},
    title style={yshift=-1.5ex,font=\small},
    xticklabel style={font=\tiny},
    yticklabel style={font=\tiny},
    legend style={
        draw=none,
       font=\scriptsize,
        legend image code/.code={
            \draw[mark repeat=2,mark phase=2]
                plot coordinates {
                    (0cm,0cm)
                    (0.3cm,0cm) 
                };
        },
    },
    ymin=0, ymax=1,
    legend pos=south east,
    legend cell align={left},
    grid style=dashed,
    width=0.45\linewidth,
    height=0.4\linewidth,
    scaled x ticks=false,
    xticklabel style={/pgf/number format/fixed}
    ]
    
    \addplot[
        color=blue,
        ]
        coordinates {
 (-0.595756,0.0) (-0.55905,0.01) (-0.548583,0.02) (-0.48635,0.03) (-0.48161,0.04) (-0.47934,0.05) (-0.471447,0.06) (-0.465658,0.07) (-0.457966,0.08) (-0.415545,0.09) (-0.387872,0.1) (-0.373835,0.11) (-0.350797,0.12) (-0.346036,0.13) (-0.3133,0.14) (-0.26506,0.15) (-0.259927,0.16) (-0.252193,0.17) (-0.23644,0.18) (-0.23642,0.19) (-0.231177,0.2) (-0.225758,0.21) (-0.220038,0.22) (-0.212794,0.23) (-0.209641,0.24) (-0.193021,0.25) (-0.184359,0.26) (-0.173397,0.27) (-0.147337,0.28) (-0.143707,0.29) (-0.141965,0.3) (-0.137977,0.31) (-0.12641,0.32) (-0.126296,0.33) (-0.112107,0.34) (-0.101129,0.35) (-0.098822,0.36) (-0.09189,0.37) (-0.090265,0.38) (-0.089068,0.39) (-0.083623,0.4) (-0.076271,0.41) (-0.071167,0.42) (-0.069393,0.43) (-0.037524,0.44) (-0.033449,0.45) (-0.03231,0.46) (-0.015558,0.47) (-0.010255,0.48) (0.001133,0.49) (0.010666,0.5) (0.012436,0.51) (0.026676,0.52) (0.031346,0.53) (0.042217,0.54) (0.045505,0.55) (0.047831,0.56) (0.050908,0.57) (0.052759,0.58) (0.056428,0.59) (0.056445,0.6) (0.066088,0.61) (0.06961,0.62) (0.07143,0.63) (0.078168,0.64) (0.078657,0.65) (0.101344,0.66) (0.108909,0.67) (0.123537,0.68) (0.125601,0.69) (0.134423,0.7) (0.139788,0.71) (0.143828,0.72) (0.156701,0.73) (0.162054,0.74) (0.170754,0.75) (0.198452,0.76) (0.199759,0.77) (0.218237,0.78) (0.22304,0.79) (0.230961,0.8) (0.235666,0.81) (0.243019,0.82) (0.257088,0.83) (0.279385,0.84) (0.28119,0.85) (0.299665,0.86) (0.324504,0.87) (0.327467,0.88) (0.331608,0.89) (0.333369,0.9) (0.36495,0.91) (0.427199,0.92) (0.478055,0.93) (0.482491,0.94) (0.501993,0.95) (0.504053,0.96) (0.506654,0.97) (0.508822,0.98) (0.517265,0.99)

        };

    \addplot[
        color=red,
        ]
        coordinates {
 (-0.593633,0.0) (-0.548797,0.01) (-0.523284,0.02) (-0.50437,0.03) (-0.469348,0.04) (-0.462087,0.05) (-0.448181,0.06) (-0.439951,0.07) (-0.420186,0.08) (-0.417235,0.09) (-0.410247,0.1) (-0.371734,0.11) (-0.352467,0.12) (-0.346597,0.13) (-0.312144,0.14) (-0.308999,0.15) (-0.301622,0.16) (-0.293041,0.17) (-0.278574,0.18) (-0.246671,0.19) (-0.220464,0.2) (-0.212989,0.21) (-0.211367,0.22) (-0.197818,0.23) (-0.193922,0.24) (-0.192574,0.25) (-0.180486,0.26) (-0.172609,0.27) (-0.146468,0.28) (-0.142668,0.29) (-0.139829,0.3) (-0.131223,0.31) (-0.113705,0.32) (-0.111858,0.33) (-0.105046,0.34) (-0.095571,0.35) (-0.084763,0.36) (-0.075634,0.37) (-0.066744,0.38) (-0.060097,0.39) (-0.052787,0.4) (-0.051849,0.41) (-0.049434,0.42) (-0.033929,0.43) (-0.028906,0.44) (-0.027214,0.45) (-0.010471,0.46) (-0.009408,0.47) (0.007322,0.48) (0.007582,0.49) (0.014393,0.5) (0.017897,0.51) (0.020683,0.52) (0.028697,0.53) (0.044754,0.54) (0.045543,0.55) (0.052866,0.56) (0.07646,0.57) (0.080254,0.58) (0.087699,0.59) (0.094254,0.6) (0.096082,0.61) (0.098461,0.62) (0.099659,0.63) (0.104665,0.64) (0.114206,0.65) (0.116893,0.66) (0.123569,0.67) (0.137561,0.68) (0.140905,0.69) (0.153934,0.7) (0.161606,0.71) (0.169362,0.72) (0.199006,0.73) (0.199309,0.74) (0.199777,0.75) (0.215864,0.76) (0.236419,0.77) (0.243292,0.78) (0.248665,0.79) (0.248805,0.8) (0.256502,0.81) (0.263799,0.82) (0.300465,0.83) (0.303303,0.84) (0.31901,0.85) (0.329774,0.86) (0.33178,0.87) (0.355485,0.88) (0.366253,0.89) (0.376385,0.9) (0.40858,0.91) (0.486361,0.92) (0.489331,0.93) (0.504246,0.94) (0.507442,0.95) (0.507475,0.96) (0.539753,0.97) (0.571625,0.98) (0.589502,0.99)
        };

    \end{axis}
    \end{tikzpicture}\hspace{-2.5mm}
        \begin{tikzpicture}
    \begin{axis}[
    title={Age},
    title style={yshift=-1.5ex,font=\small},
    xticklabel style={font=\tiny},
    yticklabel style={font=\tiny},
    legend style={
        draw=none,
        legend image code/.code={
            \draw[mark repeat=2,mark phase=2]
                plot coordinates {
                    (0cm,0cm)
                    (0.3cm,0cm) 
                };
        },
    },
    xmin=18, xmax=82,
    ymin=0, ymax=1,
    legend pos=south east,
    legend cell align={left},
    grid style=dashed,
    width=0.45\linewidth,
    height=0.4\linewidth,
    scaled x ticks=false,
    xticklabel style={/pgf/number format/fixed}
    ]
    
    \addplot[
        color=blue,
        ]
        coordinates {
         (22.0,0.0) (22.0,0.01) (22.0,0.02) (22.0,0.03) (23.0,0.04) (23.0,0.05) (23.0,0.06) (24.0,0.07) (24.0,0.08) (24.0,0.09) (25.0,0.1) (25.0,0.11) (25.0,0.12) (25.0,0.13) (26.0,0.14) (26.0,0.15) (26.0,0.16) (26.0,0.17) (27.0,0.18) (27.0,0.19) (27.0,0.2) (27.0,0.21) (28.0,0.22) (28.0,0.23) (28.0,0.24) (28.0,0.25) (28.0,0.26) (29.0,0.27) (29.0,0.28) (30.0,0.29) (30.0,0.3) (30.0,0.31) (31.0,0.32) (31.0,0.33) (31.0,0.34) (31.0,0.35) (31.0,0.36) (32.0,0.37) (32.0,0.38) (32.0,0.39) (33.0,0.4) (33.0,0.41) (34.0,0.42) (34.0,0.43) (34.0,0.44) (35.0,0.45) (35.0,0.46) (35.0,0.47) (36.0,0.48) (36.0,0.49) (36.0,0.51) (36.0,0.52) (36.0,0.53) (36.0,0.54) (37.0,0.55) (37.0,0.56) (37.0,0.57) (37.0,0.58) (37.0,0.59) (37.0,0.6) (37.0,0.61) (38.0,0.62) (38.0,0.63) (39.0,0.64) (39.0,0.65) (39.0,0.66) (39.0,0.67) (39.0,0.68) (40.0,0.69) (40.0,0.7) (40.0,0.71) (41.0,0.72) (41.0,0.73) (41.0,0.74) (42.0,0.75) (43.0,0.76) (43.0,0.77) (43.0,0.78) (44.0,0.79) (44.0,0.8) (45.0,0.81) (45.0,0.82) (45.0,0.83) (47.0,0.84) (47.0,0.85) (48.0,0.86) (50.0,0.87) (51.0,0.88) (53.0,0.89) (53.0,0.9) (54.0,0.91) (54.0,0.92) (54.0,0.93) (54.0,0.94) (58.0,0.95) (58.0,0.96) (61.0,0.97) (65.0,0.98) (74.0,0.99) (74.0,1.0)
        };

    \addplot[
        color=red,
        ]
        coordinates {
         (23.0,0.0) (23.0,0.01) (23.0,0.02) (23.0,0.03) (24.0,0.04) (24.0,0.05) (25.0,0.06) (25.0,0.07) (25.0,0.08) (25.0,0.09) (25.0,0.1) (26.0,0.11) (26.0,0.12) (27.0,0.13) (27.0,0.14) (27.0,0.15) (27.0,0.16) (28.0,0.17) (28.0,0.18) (28.0,0.19) (28.0,0.2) (28.0,0.21) (28.0,0.22) (28.0,0.23) (29.0,0.24) (29.0,0.25) (29.0,0.26) (29.0,0.27) (29.0,0.28) (30.0,0.29) (31.0,0.3) (31.0,0.31) (31.0,0.32) (31.0,0.33) (31.0,0.34) (32.0,0.35) (32.0,0.36) (32.0,0.37) (32.0,0.38) (32.0,0.39) (33.0,0.4) (33.0,0.41) (33.0,0.42) (34.0,0.43) (34.0,0.44) (35.0,0.45) (36.0,0.46) (36.0,0.47) (36.0,0.48) (36.0,0.49) (36.0,0.51) (36.0,0.52) (36.0,0.53) (36.0,0.54) (36.0,0.55) (36.0,0.56) (37.0,0.57) (37.0,0.58) (37.0,0.59) (37.0,0.6) (37.0,0.61) (38.0,0.62) (38.0,0.63) (39.0,0.64) (39.0,0.65) (39.0,0.66) (39.0,0.67) (40.0,0.68) (40.0,0.69) (41.0,0.7) (41.0,0.71) (41.0,0.72) (41.0,0.73) (41.0,0.74) (42.0,0.75) (43.0,0.76) (44.0,0.77) (44.0,0.78) (44.0,0.79) (45.0,0.8) (45.0,0.81) (46.0,0.82) (47.0,0.83) (47.0,0.84) (47.0,0.85) (48.0,0.86) (50.0,0.87) (50.0,0.88) (51.0,0.89) (53.0,0.9) (54.0,0.91) (54.0,0.92) (54.0,0.93) (54.0,0.94) (58.0,0.95) (59.0,0.96) (62.0,0.97) (65.0,0.98) (72.0,0.99) (74.0,1.0)
        };

    \end{axis}
    \end{tikzpicture}
    ~\\
    \hspace{-4mm}
        \begin{tikzpicture}
    \begin{axis}[
    title={Credit Amount},
    title style={yshift=-1.5ex,font=\small},
    xticklabel style={font=\tiny},
    yticklabel style={font=\tiny},
    legend style={
        draw=none,
        legend image code/.code={
            \draw[mark repeat=2,mark phase=2]
                plot coordinates {
                    (0cm,0cm)
                    (0.1cm,0cm) 
                };
        },
    },
    xmin=-2, xmax=14005,
    ymin=0, ymax=1,
    legend pos=south east,
    grid style=dashed,
    legend cell align={left},
    width=0.45\linewidth,
    height=0.4\linewidth,
    scaled x ticks=false,
    xticklabel style={/pgf/number format/fixed}
    ]
    
    \addplot[
        color=blue,
        ]
        coordinates {
         (250.0,0.0) (385.0,0.01) (392.0,0.02) (426.0,0.03) (433.0,0.04) (652.0,0.05) (654.0,0.06) (846.0,0.07) (860.0,0.08) (884.0,0.09) (888.0,0.1) (907.0,0.11) (930.0,0.12) (932.0,0.13) (937.0,0.14) (1007.0,0.15) (1076.0,0.16) (1123.0,0.17) (1155.0,0.18) (1163.0,0.19) (1236.0,0.2) (1237.0,0.21) (1262.0,0.22) (1295.0,0.23) (1308.0,0.24) (1316.0,0.25) (1318.0,0.26) (1355.0,0.27) (1374.0,0.28) (1402.0,0.29) (1413.0,0.3) (1414.0,0.31) (1453.0,0.32) (1473.0,0.33) (1474.0,0.34) (1478.0,0.35) (1546.0,0.36) (1549.0,0.37) (1553.0,0.38) (1597.0,0.39) (1755.0,0.4) (1858.0,0.41) (1919.0,0.42) (1922.0,0.43) (1957.0,0.44) (1984.0,0.45) (2028.0,0.46) (2039.0,0.47) (2100.0,0.48) (2278.0,0.49) (2302.0,0.51) (2303.0,0.52) (2326.0,0.53) (2406.0,0.54) (2442.0,0.55) (2503.0,0.56) (2622.0,0.57) (2625.0,0.58) (2762.0,0.59) (2767.0,0.6) (2835.0,0.61) (2991.0,0.62) (3051.0,0.63) (3060.0,0.64) (3079.0,0.65) (3343.0,0.66) (3378.0,0.67) (3380.0,0.68) (3447.0,0.69) (3488.0,0.7) (3565.0,0.71) (3763.0,0.72) (3857.0,0.73) (3863.0,0.74) (3914.0,0.75) (3965.0,0.76) (4006.0,0.77) (4042.0,0.78) (4380.0,0.79) (4526.0,0.8) (4591.0,0.81) (4788.0,0.82) (5129.0,0.83) (5190.0,0.84) (5965.0,0.85) (6288.0,0.86) (6761.0,0.87) (7119.0,0.88) (7228.0,0.89) (7238.0,0.9) (7297.0,0.91) (7511.0,0.92) (7685.0,0.93) (7758.0,0.94) (7824.0,0.95) (8648.0,0.96) (11760.0,0.97) (11998.0,0.98) (12389.0,0.99) (14027.0,1.0)
        };
    
    \addplot[
        color=red,
        ]
        coordinates {
         (284.0,0.0) (431.0,0.01) (662.0,0.02) (820.0,0.03) (844.0,0.04) (886.0,0.05) (959.0,0.06) (1123.0,0.07) (1169.0,0.08) (1181.0,0.09) (1240.0,0.1) (1252.0,0.11) (1258.0,0.12) (1287.0,0.13) (1320.0,0.14) (1332.0,0.15) (1337.0,0.16) (1338.0,0.17) (1366.0,0.18) (1385.0,0.19) (1399.0,0.2) (1403.0,0.21) (1408.0,0.22) (1430.0,0.23) (1441.0,0.24) (1442.0,0.25) (1485.0,0.26) (1540.0,0.27) (1578.0,0.28) (1585.0,0.29) (1599.0,0.3) (1623.0,0.31) (1629.0,0.32) (1715.0,0.33) (1716.0,0.34) (1799.0,0.35) (1809.0,0.36) (1898.0,0.37) (1901.0,0.38) (1907.0,0.39) (1927.0,0.4) (1946.0,0.41) (2069.0,0.42) (2071.0,0.43) (2099.0,0.44) (2123.0,0.45) (2174.0,0.46) (2181.0,0.47) (2291.0,0.48) (2296.0,0.49) (2336.0,0.51) (2489.0,0.52) (2586.0,0.53) (2639.0,0.54) (2684.0,0.55) (3029.0,0.56) (3049.0,0.57) (3065.0,0.58) (3083.0,0.59) (3086.0,0.6) (3092.0,0.61) (3127.0,0.62) (3169.0,0.63) (3211.0,0.64) (3351.0,0.65) (3390.0,0.66) (3407.0,0.67) (3452.0,0.68) (3478.0,0.69) (3613.0,0.7) (3656.0,0.71) (3764.0,0.72) (3764.0,0.73) (3901.0,0.74) (3912.0,0.75) (3975.0,0.76) (3979.0,0.77) (4095.0,0.78) (4104.0,0.79) (4472.0,0.8) (4563.0,0.81) (4621.0,0.82) (4773.0,0.83) (5126.0,0.84) (5152.0,0.85) (5993.0,0.86) (6435.0,0.87) (6867.0,0.88) (6976.0,0.89) (7194.0,0.9) (7244.0,0.91) (7288.0,0.92) (7472.0,0.93) (7636.0,0.94) (7654.0,0.95) (8366.0,0.96) (11609.0,0.97) (11766.0,0.98) (11987.0,0.99) (14007.0,1.0)
        };
    
    \end{axis}
    \end{tikzpicture}\hspace{-4mm}
        \begin{tikzpicture}
    \begin{axis}[
    title={Duration},
    title style={yshift=-1.5ex,font=\small},
    xticklabel style={font=\tiny},
    yticklabel style={font=\tiny},
    xmin=-2, xmax=62,
    ymin=0, ymax=1,
    legend pos=south east,
    grid style=dashed,
    legend cell align={left},
    width=0.45\linewidth,
    height=0.4\linewidth,
    scaled x ticks=false,
    xticklabel style={/pgf/number format/fixed}
    ]
    
    \addplot[
        color=blue,
        ]
        coordinates {
 (4.0,0.0) (6.0,0.01) (6.0,0.02) (6.0,0.03) (6.0,0.04) (6.0,0.05) (6.0,0.06) (6.0,0.07) (6.0,0.08) (7.0,0.09) (8.0,0.1) (8.0,0.11) (8.0,0.12) (9.0,0.13) (9.0,0.14) (9.0,0.15) (9.0,0.16) (10.0,0.17) (11.0,0.18) (12.0,0.19) (12.0,0.2) (12.0,0.21) (12.0,0.22) (12.0,0.23) (12.0,0.24) (12.0,0.25) (12.0,0.26) (12.0,0.27) (12.0,0.28) (12.0,0.29) (12.0,0.3) (12.0,0.31) (12.0,0.32) (12.0,0.33) (12.0,0.34) (12.0,0.35) (12.0,0.36) (12.0,0.37) (15.0,0.38) (15.0,0.39) (15.0,0.4) (15.0,0.41) (15.0,0.42) (15.0,0.43) (15.0,0.44) (16.0,0.45) (18.0,0.46) (18.0,0.47) (18.0,0.48) (18.0,0.49) (18.0,0.51) (18.0,0.52) (18.0,0.53) (18.0,0.54) (18.0,0.55) (18.0,0.56) (18.0,0.57) (21.0,0.58) (21.0,0.59) (24.0,0.6) (24.0,0.61) (24.0,0.62) (24.0,0.63) (24.0,0.64) (24.0,0.65) (24.0,0.66) (24.0,0.67) (24.0,0.68) (24.0,0.69) (24.0,0.7) (24.0,0.71) (24.0,0.72) (24.0,0.73) (24.0,0.74) (24.0,0.75) (27.0,0.76) (27.0,0.77) (27.0,0.78) (28.0,0.79) (28.0,0.8) (30.0,0.81) (30.0,0.82) (30.0,0.83) (30.0,0.84) (30.0,0.85) (36.0,0.86) (36.0,0.87) (36.0,0.88) (39.0,0.89) (42.0,0.9) (48.0,0.91) (48.0,0.92) (48.0,0.93) (48.0,0.94) (48.0,0.95) (48.0,0.96) (48.0,0.97) (60.0,0.98) (60.0,0.99) (60.0,1.0)

        };
    
    \addplot[
        color=red,
        ]
        coordinates {
 (1.0,0.0) (2.0,0.01) (2.0,0.02) (3.0,0.03) (4.0,0.04) (4.0,0.05) (4.0,0.06) (5.0,0.07) (5.0,0.08) (6.0,0.09) (6.0,0.1) (6.0,0.11) (7.0,0.12) (7.0,0.13) (7.0,0.14) (7.0,0.15) (8.0,0.16) (8.0,0.17) (9.0,0.18) (9.0,0.19) (9.0,0.2) (9.0,0.21) (10.0,0.22) (10.0,0.23) (10.0,0.24) (10.0,0.25) (10.0,0.26) (11.0,0.27) (11.0,0.28) (12.0,0.29) (12.0,0.3) (12.0,0.31) (12.0,0.32) (12.0,0.33) (12.0,0.34) (12.0,0.35) (12.0,0.36) (12.0,0.37) (12.0,0.38) (12.0,0.39) (13.0,0.4) (14.0,0.41) (14.0,0.42) (14.0,0.43) (14.0,0.44) (15.0,0.45) (15.0,0.46) (15.0,0.47) (16.0,0.48) (17.0,0.49) (17.0,0.51) (17.0,0.52) (17.0,0.53) (17.0,0.54) (17.0,0.55) (17.0,0.56) (18.0,0.57) (18.0,0.58) (18.0,0.59) (22.0,0.6) (22.0,0.61) (22.0,0.62) (22.0,0.63) (22.0,0.64) (23.0,0.65) (23.0,0.66) (24.0,0.67) (24.0,0.68) (24.0,0.69) (24.0,0.7) (24.0,0.71) (24.0,0.72) (24.0,0.73) (24.0,0.74) (24.0,0.75) (24.0,0.76) (26.0,0.77) (26.0,0.78) (27.0,0.79) (27.0,0.8) (27.0,0.81) (28.0,0.82) (29.0,0.83) (30.0,0.84) (30.0,0.85) (36.0,0.86) (36.0,0.87) (38.0,0.88) (39.0,0.89) (42.0,0.9) (48.0,0.91) (48.0,0.92) (48.0,0.93) (48.0,0.94) (48.0,0.95) (48.0,0.96) (48.0,0.97) (60.0,0.98) (60.0,0.99) (60.0,1.0)
        };
    
    \end{axis}
    \end{tikzpicture}\hspace{-4mm}
    \begin{tikzpicture}
\begin{axis}[
    xshift=0cm, 
    yshift=-5cm, 
    title={Purpose},
    title style={yshift=-1.5ex,font=\small},
    xlabel style={font=\tiny},
    ylabel style={font=\tiny},
    ymin = 0,
    xticklabel style={rotate=30, anchor=north east}, 
    yticklabel style={font=\tiny},
    legend style={
        font=\tiny,
        draw = none,
        legend image code/.code={
            \draw[mark repeat=2,mark phase=2]
                plot coordinates {
                    (0cm,0cm)
                    (0.3cm,0cm) 
                };
        },
    },
    symbolic x coords={Car, Radio/TV, Furniture/Equipment, Business, Education, Repairs, Domestic Appliances, Vacation/Others},
    xtick=data,
    xticklabels = {},
    ybar, 
    bar width=2pt, 
    legend pos=north east,
    legend cell align={left},
    xtick align=inside,
    width=0.45\linewidth,
    height=0.4\linewidth,
]

\addplot+[ybar, fill=blue, draw=black] coordinates {
    (Car, 36)
    (Radio/TV, 21)
    (Furniture/Equipment, 15)
    (Business, 14)
    (Education, 8)
    (Repairs, 3)
    (Domestic Appliances, 2)
    (Vacation/Others, 1)
};
\addlegendentry{Fact.}

\addplot+[ybar, fill=red, draw=black] coordinates {
    (Car, 37)
    (Radio/TV, 22)
    (Furniture/Equipment, 14)
    (Business, 13)
    (Education, 8)
    (Repairs, 3)
    (Domestic Appliances, 2)
    (Vacation/Others, 1)
};
\addlegendentry{C.Fact.}

\end{axis}
\end{tikzpicture}
    ~\\
    \hspace{-4mm}
    \begin{tikzpicture}
\begin{axis}[
    xshift=0cm, 
    yshift=-5cm, 
    title={Saving Accounts},
    title style={yshift=-1.5ex,font=\small},
    xlabel style={font=\tiny},
    ylabel style={font=\tiny},
    ymin = 0,
    xticklabel style={rotate=30, anchor=north east}, 
    yticklabel style={font=\tiny},
    legend style={
        legend image code/.code={
            \draw[mark repeat=2,mark phase=2]
                plot coordinates {
                    (0cm,0cm)
                    (0.3cm,0cm) 
                };
        },
    },
    symbolic x coords={Unknown, Little, Moderate, Rich, Quite Rich},
    xtick=data,
    xticklabels = {},
    ybar, 
    bar width=2pt, 
    enlarge x limits=0.25, 
    legend pos=north east,
    legend cell align={left},
    xtick align=inside,
    width=0.45\linewidth,
    height=0.4\linewidth,
]

\addplot+[ybar, fill=blue, draw=black] coordinates {
    (Unknown, 18)
    (Little, 59)
    (Moderate, 9)
    (Rich, 5)
    (Quite Rich, 9)
};

\addplot+[ybar, fill=red, draw=black] coordinates {
    (Unknown, 21)
    (Little, 53)
    (Moderate, 11)
    (Rich, 5)
    (Quite Rich, 9)
};

\end{axis}
\end{tikzpicture} \hspace{-4mm}
    \begin{tikzpicture}
\begin{axis}[
    title={Housing},
    title style={yshift=-1.5ex,font=\small},
    xlabel style={font=\tiny},
    ylabel style={font=\tiny},
    ymin = 0,
    xticklabel style={font=\tiny},
    yticklabel style={font=\tiny},
    legend style={
        legend image code/.code={
            \draw[mark repeat=2,mark phase=2]
                plot coordinates {
                    (0cm,0cm)
                    (0.3cm,0cm) 
                };
        },
    },
    symbolic x coords={Free, Rent, Own},
    xtick=data,
    xticklabels = {},
    ybar, 
    bar width=2pt, 
    enlarge x limits=0.25, 
    legend pos=north east,
    legend cell align={left},
    xtick align=inside,
    width=0.45\linewidth,
    height=0.4\linewidth,
]

\addplot+[ybar, fill=blue, draw=black] coordinates {
    (Free, 10)
    (Rent, 17)
    (Own, 73)

};

\addplot+[ybar, fill=red, draw=black] coordinates {
    (Free, 10)
    (Rent, 17)
    (Own, 73)

};

\end{axis}
\end{tikzpicture} \hspace{-4mm}
    \begin{tikzpicture}
\begin{axis}[
    xshift=0cm, 
    yshift=-5cm, 
    title={Job},
    title style={ yshift=-1.5ex,font=\small},
    xlabel style={font=\tiny},
    ylabel style={font=\tiny},
    xticklabel style={ rotate=30, anchor=north east}, 
    yticklabel style={font=\tiny},
    ymin = 0,
    legend style={
        legend image code/.code={
            \draw[mark repeat=2,mark phase=2]
                plot coordinates {
                    (0cm,0cm)
                    (0.3cm,0cm) 
                };
        },
    },
    symbolic x coords={Job0, Job1, Job2, Job3},
    xtick=data,
    xticklabels = {},
    ybar, 
    bar width=2pt, 
    enlarge x limits=0.25, 
    legend pos=north east,
    legend cell align={left},
    xtick align=inside,
    width=0.45\linewidth,
    height=0.4\linewidth,
]

\addplot+[ybar, fill=blue, draw=black] coordinates {
    (Job0, 1)
    (Job1, 24)
    (Job2, 64)
    (Job3, 11)
};

\addplot+[ybar, fill=red, draw=black] coordinates {
    (Job0, 1)
    (Job1, 24)
    (Job2, 64)
    (Job3, 11)
};

\end{axis}
\end{tikzpicture}
    \caption{\small [\textit{German-Credit}, DNN] The x-axis is feature/target value and the y-axis is quantile/quantities. The first plot ``Risk'' shows the model's output distribution. The second plot ``All Features'' shows the quantiles of the 1D projected (by $\Theta$) factual and counterfactual distributions.  The other plots show marginal distributions for each feature, where numerical ones are shown by quantile and categorical by histogram. Factual risk (average) is 31.3\% and counterfactual 17.5\%. }
    \label{fig:german_credit_validity}
\end{figure}

In \tablename~\ref{tab:baseline_heloc} (top), \gls{globe} outperforms the others in terms of \gls{ares} cost with reasonable coverage, as expected. The columns under ``\% Difference at Percentiles'' evaluates the percentage change in the value of the feature in the counterfactual compared to the factual in a specific percentile, being averaged across all features. With similar coverage, an algorithm achieving smaller values in these columns is more desirable. Compared to the \gls{ares} cost column that simply assumes uniform cost for the whole distribution spectrum of a feature, these columns may serve as an arguably better metric in practice\textemdash moving a value around the median should be easier than doing so at the distribution tails (i.e. extreme values), hence being with lower cost. \discount~remains stable under different models in terms of counterfactual coverage.

\tablename~\ref{tab:baseline_heloc} (bottom) shows similar results, but for \textit{COMPAS}, where most of the features are categorical (one-hot encoded as binary columns). We separately evaluate the action costs of categorical features (abbreviated as Cat.) and the two numerical features, "Priors Count" and "Time Served". The categorical difference (Cat. Diff) is defined as the average absolute difference (ranging from 0 to 1) between factual and counterfactual data points for all categorical features. 
\discount~significantly outperforms the others in both categorical and numerical feature action costs and exceeds \gls{dice} in \gls{ares} cost. 
It is worth noting that \discount~results in significantly better coverage stability and generally outperforms \gls{dice} in other cost metrics, as well as proximity.

\paragraph{\discount~achieves the best distributional proximity} In \tablename~\ref{tab:baseline_heloc}, the two columns \gls{ot} and \gls{mmd} suggest that \discount~results in better proximity than the others, with moderate runtime. \figurename~\ref{fig:german_credit_validity} does visualization for distributional counterfactual proximity and the corresponding counterfactual effect by \discount.  The figure is obtained by training a \gls{dnn} on the \textit{German-Credit} dataset with all features. We sample 100 data points from the test data set as the factual distribution, then apply the proposed \discount~on all training features to obtain the corresponding counterfactual distribution for explanation. Our target $\vec{y}^*$ is generated by a Beta distribution that represents a group of customers with very low risk, shown as the dashed black curve. \discount~achieves this target while having the feature distribution of the counterfactual resembles the one of the factual.

\begin{figure}
    \centering
    \hspace{-5mm}
    \pgfplotsset{scaled y ticks=false}
\begin{tikzpicture}
\begin{axis}[
    xticklabel style={font=\tiny},
    yticklabel style={font=\tiny},
    legend style={
        draw=none,
        legend image code/.code={
            \draw[mark repeat=2,mark phase=2]
                plot coordinates {
                    (0cm,0cm)
                    (0.3cm,0cm) 
                };
        },
    },
    xmin=0, xmax=199,
    legend cell align={left},
    legend pos = south east,
    width=0.45\linewidth,
    height=0.5\linewidth,
    scaled y ticks=false,
    y tick label style={
        /pgf/number format/fixed,
        /pgf/number format/precision=2,
        /pgf/number format/fixed zerofill
    },
]

\addplot[
    color=blue,
    ]
    coordinates {
(0,0.0)(1,0.002507180183645169)(2,0.004582773547884998)(3,0.00666269996337544)(4,0.008544549492828143)(5,0.009608986205392826)(6,0.012010725839140087)(7,0.013233650218991812)(8,0.014269230386125289)(9,0.015108004542078328)(10,0.01634171865948192)(11,0.01751973394252672)(12,0.018765800240157917)(13,0.01888185929897657)(14,0.01914923102998905)(15,0.020653497539405197)(16,0.020757187336173055)(17,0.023262685536609318)(18,0.02401184639720428)(19,0.024856478626105637)(20,0.02545506512026178)(21,0.0249836614213983)(22,0.028502346579374695)(23,0.02964526466666)(24,0.02891514919483407)(25,0.030300860205154875)(26,0.030389551619770517)(27,0.03130413192468787)(28,0.032245468795775716)(29,0.03329316078384396)(30,0.033169466506164494)(31,0.03500534213764512)(32,0.03500377699924387)(33,0.03621156799885914)(34,0.036930434398904026)(35,0.03631450100581708)(36,0.03756714404820389)(37,0.04015712514334322)(38,0.03928967178345561)(39,0.040412063363243024)(40,0.04051155948778555)(41,0.03952904573044558)(42,0.04036636489216638)(43,0.0437087989787319)(44,0.042623759607970865)(45,0.04356390027962679)(46,0.045281239215671516)(47,0.04565829882151908)(48,0.0465675091729724)(49,0.045967514886703795)(50,0.048690408894457256)(51,0.04826651360592284)(52,0.04838456049590138)(53,0.04765404865450894)(54,0.04857780885550282)(55,0.049559258602295306)(56,0.05020144164481857)(57,0.05069923799534201)(58,0.05121417201646982)(59,0.051622358902125576)(60,0.05252512237866564)(61,0.052966182201187546)(62,0.05338359840524351)(63,0.05279336678348808)(64,0.05413224831632501)(65,0.05475883255305702)(66,0.05569743156991482)(67,0.05518052532818997)(68,0.057726452343733324)(69,0.0572338295711951)(70,0.05702680897811811)(71,0.05765728036506367)(72,0.057292990120815304)(73,0.05773267862634603)(74,0.05823414829157209)(75,0.059448736431015274)(76,0.05805769303096268)(77,0.0609302746674655)(78,0.05855276077181732)(79,0.0605188385502429)(80,0.060312880874552464)(81,0.060742637834300714)(82,0.06205072445171458)(83,0.061028613318696336)(84,0.06274025650250192)(85,0.06255882656292734)(86,0.06310339124841641)(87,0.06210390100751714)(88,0.06319367338612945)(89,0.06309206193819535)(90,0.06338215265242324)(91,0.06388566438911666)(92,0.06474483646771252)(93,0.06543879111609537)(94,0.06537647137317806)(95,0.06587403512444512)(96,0.06587674953506852)(97,0.0662541758302349)(98,0.06620352570197319)(99,0.06730991086289534)(100,0.06742862702073164)(101,0.06656288831659127)(102,0.06949318942231383)(103,0.0673729049142805)(104,0.06837649339735263)(105,0.0680600441941537)(106,0.06890457227876506)(107,0.06919795932328673)(108,0.06877073291442314)(109,0.07033520927017485)(110,0.0696111099510773)(111,0.0709397791308036)(112,0.07064981175972715)(113,0.07042108824225404)(114,0.07122505989590297)(115,0.07124065096919156)(116,0.07250239453123707)(117,0.07105398287549995)(118,0.07124876263820866)(119,0.07224773721040723)(120,0.07115268338823144)(121,0.07263124904134992)(122,0.0709842489152132)(123,0.07319198237148763)(124,0.07308154351209127)(125,0.07441054838745066)(126,0.07211869371499038)(127,0.07393615098879203)(128,0.07413741042967924)(129,0.0742725439767826)(130,0.07526964661733858)(131,0.0743825130993828)(132,0.07524499340782072)(133,0.07465438821195376)(134,0.07627115248731105)(135,0.07551532771307619)(136,0.07614636012462961)(137,0.07645487501017861)(138,0.0756521801925305)(139,0.07658206568318968)(140,0.07485746453073533)(141,0.0756373904831617)(142,0.07709001950086015)(143,0.07607304649148412)(144,0.07622306887357694)(145,0.07651665801742016)(146,0.07641216264515907)(147,0.0764387879046654)(148,0.07775791719146596)(149,0.07667569868272538)(150,0.07736055514923133)(151,0.07753822147465182)(152,0.0774308214504632)(153,0.07663009614486974)(154,0.07807644848348855)(155,0.0786194926619428)(156,0.07816765143185223)(157,0.07853845403217723)(158,0.07652608719572405)(159,0.07725103563144894)(160,0.07821059980410706)(161,0.07842396805683806)(162,0.0778915571944573)(163,0.07840617467707132)(164,0.07901929092623712)(165,0.07900588854962856)(166,0.07905207807092553)(167,0.07988281120316176)(168,0.0793586222542663)(169,0.0785256360345106)(170,0.07975669674010769)(171,0.07850876048224223)(172,0.07847038610135075)(173,0.07941849190565399)(174,0.07940397824391764)(175,0.07931439980185209)(176,0.08069026163883411)(177,0.0807523814774118)(178,0.08104861788796026)(179,0.0823183452681193)(180,0.0794741388064586)(181,0.0798340954752464)(182,0.08077629475261941)(183,0.08179379288539497)(184,0.08054531428016357)(185,0.08003074351265732)(186,0.08083959468066715)(187,0.08138791010184838)(188,0.07962198267967699)(189,0.08069875850259936)(190,0.08021363167699869)(191,0.08007971375376918)(192,0.08144032764567079)(193,0.08178243930795841)(194,0.08105992706366354)(195,0.08097401606007046)(196,0.08136728469636056)(197,0.0812843086983518)(198,0.0832329071360801)(199,0.0825527993724089)

};
\addlegendentry{$\overline{\sot}$}

\addplot[
    color=red,
    ]
    coordinates {
(0,0.1)(1,0.1)(2,0.1)(3,0.1)(4,0.1)(5,0.1)(6,0.1)(7,0.1)(8,0.1)(9,0.1)(10,0.1)(11,0.1)(12,0.1)(13,0.1)(14,0.1)(15,0.1)(16,0.1)(17,0.1)(18,0.1)(19,0.1)(20,0.1)(21,0.1)(22,0.1)(23,0.1)(24,0.1)(25,0.1)(26,0.1)(27,0.1)(28,0.1)(29,0.1)(30,0.1)(31,0.1)(32,0.1)(33,0.1)(34,0.1)(35,0.1)(36,0.1)(37,0.1)(38,0.1)(39,0.1)(40,0.1)(41,0.1)(42,0.1)(43,0.1)(44,0.1)(45,0.1)(46,0.1)(47,0.1)(48,0.1)(49,0.1)(50,0.1)(51,0.1)(52,0.1)(53,0.1)(54,0.1)(55,0.1)(56,0.1)(57,0.1)(58,0.1)(59,0.1)(60,0.1)(61,0.1)(62,0.1)(63,0.1)(64,0.1)(65,0.1)(66,0.1)(67,0.1)(68,0.1)(69,0.1)(70,0.1)(71,0.1)(72,0.1)(73,0.1)(74,0.1)(75,0.1)(76,0.1)(77,0.1)(78,0.1)(79,0.1)(80,0.1)(81,0.1)(82,0.1)(83,0.1)(84,0.1)(85,0.1)(86,0.1)(87,0.1)(88,0.1)(89,0.1)(90,0.1)(91,0.1)(92,0.1)(93,0.1)(94,0.1)(95,0.1)(96,0.1)(97,0.1)(98,0.1)(99,0.1)(100,0.1)(101,0.1)(102,0.1)(103,0.1)(104,0.1)(105,0.1)(106,0.1)(107,0.1)(108,0.1)(109,0.1)(110,0.1)(111,0.1)(112,0.1)(113,0.1)(114,0.1)(115,0.1)(116,0.1)(117,0.1)(118,0.1)(119,0.1)(120,0.1)(121,0.1)(122,0.1)(123,0.1)(124,0.1)(125,0.1)(126,0.1)(127,0.1)(128,0.1)(129,0.1)(130,0.1)(131,0.1)(132,0.1)(133,0.1)(134,0.1)(135,0.1)(136,0.1)(137,0.1)(138,0.1)(139,0.1)(140,0.1)(141,0.1)(142,0.1)(143,0.1)(144,0.1)(145,0.1)(146,0.1)(147,0.1)(148,0.1)(149,0.1)(150,0.1)(151,0.1)(152,0.1)(153,0.1)(154,0.1)(155,0.1)(156,0.1)(157,0.1)(158,0.1)(159,0.1)(160,0.1)(161,0.1)(162,0.1)(163,0.1)(164,0.1)(165,0.1)(166,0.1)(167,0.1)(168,0.1)(169,0.1)(170,0.1)(171,0.1)(172,0.1)(173,0.1)(174,0.1)(175,0.1)(176,0.1)(177,0.1)(178,0.1)(179,0.1)(180,0.1)(181,0.1)(182,0.1)(183,0.1)(184,0.1)(185,0.1)(186,0.1)(187,0.1)(188,0.1)(189,0.1)(190,0.1)(191,0.1)(192,0.1)(193,0.1)(194,0.1)(195,0.1)(196,0.1)(197,0.1)(198,0.1)(199,0.1)

};
\addlegendentry{$U_{x}$}

\end{axis}
\end{tikzpicture} \hspace{-4mm}
    \begin{tikzpicture}
\begin{axis}[
    xticklabel style={font=\tiny},
    yticklabel style={font=\tiny},
    legend style={
        draw=none,
        legend image code/.code={
            \draw[mark repeat=2,mark phase=2]
                plot coordinates {
                    (0cm,0cm)
                    (0.3cm,0cm) 
                };
        },
    },
    xmin=0, xmax=199,
    legend pos=north east,
    legend cell align={left},
    width=0.45\linewidth,
    height=0.5\linewidth,
  scaled y ticks=false,
]

\addplot[
    color=blue,
    ]
    coordinates {
(0,0.44121451348978663)(1,0.4103527297848091)(2,0.39026298484897365)(3,0.37731410664365944)(4,0.3662184600057462)(5,0.35757627201037767)(6,0.3462432225326643)(7,0.34389356736226034)(8,0.33638434355601066)(9,0.3312798277453121)(10,0.3319726925992578)(11,0.3243264278022505)(12,0.3182336526828723)(13,0.3171263450828786)(14,0.3147111861506821)(15,0.3091124605503911)(16,0.3126430458262241)(17,0.29966185798921147)(18,0.3070901615830458)(19,0.30329427625794536)(20,0.2972152319488262)(21,0.3030154172693726)(22,0.29601218268889035)(23,0.2948895281932636)(24,0.29065160962797165)(25,0.2865321754158129)(26,0.29027304924353853)(27,0.2847282293540175)(28,0.29291843236514936)(29,0.2798764920236439)(30,0.28168570393173176)(31,0.2812598345944426)(32,0.28449669129770244)(33,0.2763889604086293)(34,0.2774604091964175)(35,0.27357486442086115)(36,0.2766425284342013)(37,0.27498198925719386)(38,0.2714459083255251)(39,0.2690447559747337)(40,0.2695057421514427)(41,0.26616844690364155)(42,0.2652893863310563)(43,0.2640831087537764)(44,0.2633567231765335)(45,0.2675680457952565)(46,0.2636170533373807)(47,0.2608641947994772)(48,0.26409746808248186)(49,0.2614187556716751)(50,0.2578242474466037)(51,0.2584472612286929)(52,0.2516107839396725)(53,0.2606974240072087)(54,0.2502723350458943)(55,0.25354157797686616)(56,0.25120615309980404)(57,0.2535380569549501)(58,0.2469573184734091)(59,0.24747378986586668)(60,0.2409362713727412)(61,0.2451172971155294)(62,0.24487852202633612)(63,0.24758046484347396)(64,0.24879866890547142)(65,0.24117816575010703)(66,0.24512569234159126)(67,0.24280561815424279)(68,0.24117230422565955)(69,0.23567623817527653)(70,0.24130375308064206)(71,0.23891474280845693)(72,0.24491176679428978)(73,0.23990252610258447)(74,0.23944944511902838)(75,0.23951579995913863)(76,0.2417736647810034)(77,0.2403271703294389)(78,0.23365141494127595)(79,0.23510992379605222)(80,0.2358625991379168)(81,0.22562881863088893)(82,0.23298411044927692)(83,0.23316903352351318)(84,0.23751682877522565)(85,0.2352122559342114)(86,0.23428688717605342)(87,0.22819492259722482)(88,0.23534986309371855)(89,0.23677333813766946)(90,0.22634086147237745)(91,0.224619529328732)(92,0.22353055940087)(93,0.22957995643727705)(94,0.22337888414796916)(95,0.22441427983637846)(96,0.22550801235852883)(97,0.22431460256519828)(98,0.22345327912648005)(99,0.22229630127852404)(100,0.22100491525100202)(101,0.22252365459886506)(102,0.2208082296961749)(103,0.2247130517513649)(104,0.22374453814662432)(105,0.2228516630913042)(106,0.22207813333044196)(107,0.2150881006280954)(108,0.21878164096607178)(109,0.21674676479262053)(110,0.2106853187414739)(111,0.21959472749045206)(112,0.21812933591642975)(113,0.21840759348043343)(114,0.21652127352422373)(115,0.2139696232981387)(116,0.2169327481954768)(117,0.21443889119902754)(118,0.21407988979715295)(119,0.21380200655634188)(120,0.2118316665998278)(121,0.2090940045309734)(122,0.21394367135027717)(123,0.21566458495086638)(124,0.2125519848007909)(125,0.21035825861276392)(126,0.21481092007098743)(127,0.21685402327451497)(128,0.21684682996526244)(129,0.2100940470923196)(130,0.20967007872715115)(131,0.20407297239880812)(132,0.20746860552468308)(133,0.20726627985832502)(134,0.2129852690433429)(135,0.2084036722750257)(136,0.21415415610637273)(137,0.21078879359852257)(138,0.20917986790553986)(139,0.2063660468117452)(140,0.20160487689660725)(141,0.20579150237349586)(142,0.20584583070523438)(143,0.20381763823857177)(144,0.20974150657521964)(145,0.2022922020779287)(146,0.21034837063321984)(147,0.20710159447228066)(148,0.21171102157788743)(149,0.202061528721598)(150,0.2050562356438481)(151,0.20723426779354728)(152,0.19866421412864546)(153,0.20215711160376817)(154,0.2055980875424834)(155,0.20719658462211848)(156,0.1993070070329803)(157,0.20083194456339176)(158,0.20437276107322905)(159,0.2036967703262673)(160,0.20696227841135287)(161,0.20242017113827218)(162,0.19915961177856625)(163,0.20496046801070367)(164,0.199909204869379)(165,0.19468405540849534)(166,0.19603208331186842)(167,0.2036796358789646)(168,0.19971140224831144)(169,0.19703712614003469)(170,0.19959983653590288)(171,0.20414944274215147)(172,0.19950374432356796)(173,0.1988866197244352)(174,0.2000529313266052)(175,0.20260416665645165)(176,0.19385839619900208)(177,0.1973351759626124)(178,0.20496975851225238)(179,0.2070749954125162)(180,0.2001177530543184)(181,0.19312708609728965)(182,0.19510228073370714)(183,0.19281186486074672)(184,0.20049089823973762)(185,0.2034147607529326)(186,0.1986915487001869)(187,0.20052859964234745)(188,0.19644821438112872)(189,0.1946401835886205)(190,0.19413156545827165)(191,0.19284860479850266)(192,0.19513123946619787)(193,0.19158451333053486)(194,0.19031464867756953)(195,0.19517546901183558)(196,0.1959389945536747)(197,0.1957077342764058)(198,0.18783833852806586)(199,0.19397538226688602)

};
\addlegendentry{$\overline{\ot}$}

\addplot[
    color=red,
    ]
    coordinates {
(0,0.25)(1,0.25)(2,0.25)(3,0.25)(4,0.25)(5,0.25)(6,0.25)(7,0.25)(8,0.25)(9,0.25)(10,0.25)(11,0.25)(12,0.25)(13,0.25)(14,0.25)(15,0.25)(16,0.25)(17,0.25)(18,0.25)(19,0.25)(20,0.25)(21,0.25)(22,0.25)(23,0.25)(24,0.25)(25,0.25)(26,0.25)(27,0.25)(28,0.25)(29,0.25)(30,0.25)(31,0.25)(32,0.25)(33,0.25)(34,0.25)(35,0.25)(36,0.25)(37,0.25)(38,0.25)(39,0.25)(40,0.25)(41,0.25)(42,0.25)(43,0.25)(44,0.25)(45,0.25)(46,0.25)(47,0.25)(48,0.25)(49,0.25)(50,0.25)(51,0.25)(52,0.25)(53,0.25)(54,0.25)(55,0.25)(56,0.25)(57,0.25)(58,0.25)(59,0.25)(60,0.25)(61,0.25)(62,0.25)(63,0.25)(64,0.25)(65,0.25)(66,0.25)(67,0.25)(68,0.25)(69,0.25)(70,0.25)(71,0.25)(72,0.25)(73,0.25)(74,0.25)(75,0.25)(76,0.25)(77,0.25)(78,0.25)(79,0.25)(80,0.25)(81,0.25)(82,0.25)(83,0.25)(84,0.25)(85,0.25)(86,0.25)(87,0.25)(88,0.25)(89,0.25)(90,0.25)(91,0.25)(92,0.25)(93,0.25)(94,0.25)(95,0.25)(96,0.25)(97,0.25)(98,0.25)(99,0.25)(100,0.25)(101,0.25)(102,0.25)(103,0.25)(104,0.25)(105,0.25)(106,0.25)(107,0.25)(108,0.25)(109,0.25)(110,0.25)(111,0.25)(112,0.25)(113,0.25)(114,0.25)(115,0.25)(116,0.25)(117,0.25)(118,0.25)(119,0.25)(120,0.25)(121,0.25)(122,0.25)(123,0.25)(124,0.25)(125,0.25)(126,0.25)(127,0.25)(128,0.25)(129,0.25)(130,0.25)(131,0.25)(132,0.25)(133,0.25)(134,0.25)(135,0.25)(136,0.25)(137,0.25)(138,0.25)(139,0.25)(140,0.25)(141,0.25)(142,0.25)(143,0.25)(144,0.25)(145,0.25)(146,0.25)(147,0.25)(148,0.25)(149,0.25)(150,0.25)(151,0.25)(152,0.25)(153,0.25)(154,0.25)(155,0.25)(156,0.25)(157,0.25)(158,0.25)(159,0.25)(160,0.25)(161,0.25)(162,0.25)(163,0.25)(164,0.25)(165,0.25)(166,0.25)(167,0.25)(168,0.25)(169,0.25)(170,0.25)(171,0.25)(172,0.25)(173,0.25)(174,0.25)(175,0.25)(176,0.25)(177,0.25)(178,0.25)(179,0.25)(180,0.25)(181,0.25)(182,0.25)(183,0.25)(184,0.25)(185,0.25)(186,0.25)(187,0.25)(188,0.25)(189,0.25)(190,0.25)(191,0.25)(192,0.25)(193,0.25)(194,0.25)(195,0.25)(196,0.25)(197,0.25)(198,0.25)(199,0.25)

};
\addlegendentry{$U_{y}$}

\end{axis}
\end{tikzpicture} \hspace{-4mm}
    \input{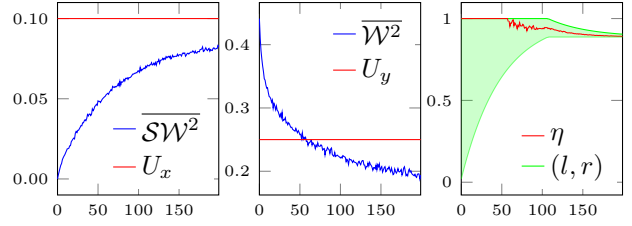}
    \caption{\small [\textit{Cardiovascular Disease}, DNN] Convergence of \discount~with Interval Narrowing. The optimization starts from a point that is near to the factual $\vec{x}'$. At the beginning $\eta$ stays at $r=1$ due to the violation of the chance constraint of $y$, such that the optimization leans entirely towards bringing $y$ to $y^{*}$ closer gradually. When $\overline{\ot}$ is below $U_2$ at iteration $57$, a feasible solution is found. Then $\eta$ is optimized within $[l,r]$ to balance the gaps $U_{x}-\overline{\sot}$ and $U_{y}-\overline{\ot}$, until the algorithm converges.}
    \label{fig:convergence}
\end{figure}

\paragraph{\discount~convergence visualization} \figurename~\ref{fig:convergence} demonstrates the convergence behavior of \discount, where $\eta$ is computed by Interval Narrowing in Algorithm \ref{alg:eta_bisection}. We argue that the convergence behavior of Set Shrinking in Algorithm \ref{alg:eta_discrete} is similar (guaranteed by Theorem \ref{thm:convergence_combined}). Since the interval $[l,r]$ is straightforward to visualize, the convergence behavior of Interval Narrowing is therefore selected to be shown here.

\section{Conclusions}
We have broadened the scope of \gls{ce} by introducing it in a distributional context, termed \gls{dce}, under a statistically rigorous framework. The numerical results demonstrate the effectiveness of this method. 
Some explorations on geometric interpretations as well as unresolved challenges are discussed in Appendix~\ref{sec:geometric}.




\bibliographystyle{apalike}
\bibliography{paper}

\section*{Checklist}



 \begin{enumerate}

 \item For all models and algorithms presented, check if you include:
 \begin{enumerate}
   \item A clear description of the mathematical setting, assumptions, algorithm, and/or model. [Yes]
   \item An analysis of the properties and complexity (time, space, sample size) of any algorithm. [Yes]
   \item (Optional) Anonymized source code, with specification of all dependencies, including external libraries. [Yes]
 \end{enumerate}

 \item For any theoretical claim, check if you include:
 \begin{enumerate}
   \item Statements of the full set of assumptions of all theoretical results. [Yes]
   \item Complete proofs of all theoretical results. [Yes]
   \item Clear explanations of any assumptions. [Yes]     
 \end{enumerate}

 \item For all figures and tables that present empirical results, check if you include:
 \begin{enumerate}
   \item The code, data, and instructions needed to reproduce the main experimental results (either in the supplemental material or as a URL). [Yes]
   \item All the training details (e.g., data splits, hyperparameters, how they were chosen). [Yes]
         \item A clear definition of the specific measure or statistics and error bars (e.g., with respect to the random seed after running experiments multiple times). [Not Applicable] \\
\textit{Explanations: This question relates to  \tablename~\ref{tab:baseline_heloc}  and \tablename~\ref{tab:diversity}. The rationale is that the  significance is evident, as the substantial differences between the values across methods are clear.}
         \item A description of the computing infrastructure used. (e.g., type of GPUs, internal cluster, or cloud provider). [Yes]
 \end{enumerate}

 \item If you are using existing assets (e.g., code, data, models) or curating/releasing new assets, check if you include:
 \begin{enumerate}
   \item Citations of the creator If your work uses existing assets. [Yes]
   \item The license information of the assets, if applicable. [Not Applicable]
   \item New assets either in the supplemental material or as a URL, if applicable. [Not Applicable]
   \item Information about consent from data providers/curators. [Not Applicable]
   \item Discussion of sensible content if applicable, e.g., personally identifiable information or offensive content. [Not Applicable]
 \end{enumerate}

 \item If you used crowdsourcing or conducted research with human subjects, check if you include:
 \begin{enumerate}
   \item The full text of instructions given to participants and screenshots. [Not Applicable]
   \item Descriptions of potential participant risks, with links to Institutional Review Board (IRB) approvals if applicable. [Not Applicable]
   \item The estimated hourly wage paid to participants and the total amount spent on participant compensation. [Not Applicable]
 \end{enumerate}

 \end{enumerate}

\appendix
 \onecolumn
\aistatstitlesupplementary{Distributional Counterfactual Explanations With Optimal Transport: \\
Supplementary Materials}

{\small
\setcounter{tocdepth}{1}
\startcontents[sections]
\printcontents[sections]{l}{1}{}
}

\section{Related Work}
\label{sec:related_work}
The pioneering study by \citep{wachter2017counterfactual} is a seminal work in the field of \gls{ce} and has gained notable recognition. This study approaches \gls{ce} through a minimization problem, laying the foundation for subsequent explorations in the field. 
A comprehensive overview of the CE literature is presented in the works of \citep{guidotti2022counterfactual} and \citep{bodria2023benchmarking}. Our focus aligns with the model-agnostic optimization approach for generating counterfactuals as proposed by \citep{wachter2017counterfactual}. Several studies have explored the generation of multiple counterfactuals in one shot \citep{ustun2019actionable, karimi2020model, 
mothilal2020explaining, pawelczyk2020learning, brughmans2023nice}, providing insight into the explanation of individual factual instances.

Recent developments in the field have seen a growing interest in group-based \gls{ce}.  \citep{rawal2020beyond} introduced a framework for generating \gls{ce} for sub-populations within datasets.  \citep{plumb2020explaining} proposed a counterfactual-summary method for explaining subgroup data points. Furthermore, \citep{warren2023explaining} developed a novel approach for creating \gls{gce}, focusing on common feature differences among similar individuals. \citep{carrizosa2024mathematical} takes a stakeholder's point of view to explain the necessity of considering GCE.

In \citep{kulinski2023towards,koebler2023towards}, the authors target interpreting the distribution shift phenomenon and monitoring that. We remark that leveraging methods like \cite{kulinski2023towards,koebler2023towards} post hoc to interpret the distributions and transportation maps produced by our proposed \gls{dce} could provide qualitative insights, which is further discussed in Appendix~\ref{sec:geometric}.

To our knowledge, no prior work has established a theoretical framework to ensure that the generated group of counterfactual data points is in provable proximity to the factual distribution. Our work addresses this critical gap by embedding statistical rigor into the CE framework, providing guarantees for distributional similarity.

\section{Proof of Theorem \ref{thm:interval}: \gls{ucl} for Problem Solving}
\label{sec:proof_thm_interval}
\begin{theorem}[\textbf{Theorem~\ref{thm:interval}} in the main text]
Let \( \delta \in (0, 1/2) \) be a trimming constant, the following inequalities hold:

1. For $\ot(b(\vec{x}),y^*)$ in \eqref{eq:main_problem_C2}, we have
\begin{align}
    &\mathbb{P}\left[ \ot(b(\vec{x}),y^*) \leq \frac{1}{1 - 2\delta} \int_{\delta}^{1 - \delta}  D(u) \, \dd u \right] \geq 1 - \frac{\alpha}{2},
    \label{eq:int_D1-appendix}
\end{align}
\begin{align}
   &\text{where } D(u) \triangleq  \max\bigg\{  F_{y,n}^{-1}\left( \overline{q}_{\alpha, n}(u) \right) - F_{y^{*}, n}^{-1}\left( \underline{q}_{\alpha, n}(u) \right),  F_{y^{*}, n}^{-1}\left( \overline{q}_{\alpha, n}(u) \right) - F_{y, n}^{-1}\left( \underline{q}_{\alpha, n}(u) \right) \bigg\}.
    \label{eq:D1-appendix}
\end{align}
2. For  \( \sot(\vec{x},\vec{x}') \) in \eqref{eq:main_problem_C1}, let the projection vectors \( \bm{\theta}_1, \ldots, \bm{\theta}_N \) be independent and identically distributed samples from a distribution \( \sigma \) on the unit sphere \( \mathbb{S}^{d-1} \). Let \( \sigma_N \) denote the empirical measure associated with these samples. Then,
\begin{align}
    &\mathbb{P}\left[ \sot(\vec{x},\vec{x}') \leq  \frac{1}{1 - 2\delta} \int_{\mathbb{S}^{d-1}}\int_{\delta}^{1 - \delta} D_{\bm{\theta},N}(u) \, \dd u \, \dd\sigma_N(\bm{\theta}) \right] \geq 1 - \frac{\alpha}{2}
    \label{eq:int_DN-appendix}
\end{align}
\begin{align}
    &\text{where } D_{\bm{\theta},N}(u)\triangleq \max \bigg\{ F_{\bm{\theta}^\top \vec{x}, n}^{-1}\left( \overline{q}_{\alpha, n}(u) \right) 
        - F_{\bm{\theta}^\top \vec{x}', n}^{-1}\left( \underline{q}_{\alpha, n}(u) \right), F_{\bm{\theta}^\top \vec{x}', n}^{-1}\left( \overline{q}_{\alpha, n}(u) \right) 
    - F_{\bm{\theta}^\top \vec{x}, n}^{-1}\left( \underline{q}_{\alpha, n}(u) \right)
    \bigg\}.
    \label{eq:DthetaN-appendix}
\end{align}
Here, \( F_{\bm{\theta}^\top \vec{x}, n}^{-1} \) denotes the empirical quantile function of  \( \bm{\theta}^\top \vec{x}_i \) for \( i = 1, \dots, n \), and similarly for \( F_{\bm{\theta}^\top \vec{x}', n}^{-1} \).

\end{theorem}
\begin{proof}
The proof is a sketch of the key results of \citep[Section 4]{manole2022minimax}, with necessary modifications in details to suit the need of this paper. The proof applies for $\delta$-trimmed ($\delta\in (0,1/2)$) empirical distributions, without assumptions on the knowledge of the distributions. The trimming is necessary, such that certain amount of data points are left out for capturing the tail behavior of the distributions.
Note that \eqref{eq:alpha_over_2N} reads
\begin{equation}
  \inf \PP_n\big(F_n^{-1}(\underline{q}_{\alpha, n}(u)) \leq  F^{-1}(u)\leq  F_n^{-1}(\overline{q}_{\alpha, n}(u)), \forall u\in(0,1) \big) \geq 1-\frac{\alpha}{2},
\label{eq:alpha_over_2N-appendix}
\end{equation}
which indeed establishes a confidence band for the two sequences of functions, $\underline{q}_{\alpha, n}$ and $\overline{q}_{\alpha, n}$. The proof then pivots on the assertion that the \gls{1d} $2$-Wasserstein distance can be characterized as the $L^2$-norm of the quantile functions of $y$ ($y=b(\vec{x})$) and $\vec{x}'$ ($y'=b(\vec{x}')$), namely\footnote{This equation is exactly discussed in \eqref{eq:y-quantile} in Section~\ref{sec:foundations}.},
\begin{equation}
\ot(\gamma_1, \gamma_2) = \int_{0}^1\left|F^{-1}_{\gamma_1,n}(u) - F^{-1}_{\gamma_2,n}(u)\right|^2 \dd u.
\label{eq:y-quantile-appendix}
\end{equation}
holds for any \gls{1d} empirical distribution $\gamma_1$ and $\gamma_2$.
Then, the uniform quantile bounds can be used to construct the two sequences functions $\underline{q}_{\alpha, n}$ and $\overline{q}_{\alpha, n}$ \citep{shorack2009empirical}, such that the above inequality \eqref{eq:alpha_over_2N-appendix} gets hold. Straightforwardly, this yields a \gls{ucl} on $F^{-1}_{\gamma_1,n}(u)$ and $F^{-1}_{\gamma_2,n}(u)$, and therefore a \gls{ucl} on the absolute difference between them, i.e. $|F^{-1}_{y,n}(u) - F^{-1}_{y^*,n}(u)|$. That is 
\[
|F^{-1}_{y,n}(u) - F^{-1}_{y^*,n}(u)|\leq \max\bigg\{  F_{y,n}^{-1}\left( \overline{q}_{\alpha, n}(u) \right) - F_{y^{*}, n}^{-1}\left( \underline{q}_{\alpha, n}(u) \right),  F_{y^{*}, n}^{-1}\left( \overline{q}_{\alpha, n}(u) \right) - F_{y, n}^{-1}\left( \underline{q}_{\alpha, n}(u) \right) \bigg\} = D(u)
\]
This inequality holds probabilistically with confidence level $1-\alpha/2$\footnote{
To explain this, notice that 
\[
\inf \PP_n\big( F_y^{-1}(u)\leq  F_{y,n}^{-1}(\overline{q}_{\alpha, n}(u)), \forall u\in(0,1) \big) \geq 1-\frac{\alpha}{4}, \text{ and}
\]
\[
\inf \PP_n\big( F_{y^*}^{-1}(u)\leq  F_{y^*,n}^{-1}(\overline{q}_{\alpha, n}(u)), \forall u\in(0,1) \big) \geq 1-\frac{\alpha}{4}.
\]
Hence the probability of $|F_y^{-1}(u)-F_{y^*}^{-1}(u)|$ being no larger than the maximum of the two possible differences between $F_{y,n}^{-1}(\overline{q}_{\alpha, n}(u))$ and $F_{y^*,n}^{-1}(\overline{q}_{\alpha, n}(u))$, is $1-\frac{\alpha}{2}$.
}.
Consequently, by \eqref{eq:y-quantile-appendix}, the \gls{ucl} on $\ot(b(\vec{x}), y^*)$ is hence derived by taking the square of the \gls{ucl} on $|F^{-1}_{y,n}(u) - F^{-1}_{y^*,n}(u)|$ then taking integral over $u$. Hence, given $y=b(\vec{x})$.
\begin{equation}
\ot(b(\vec{x}), y^{*}) =  \int_{0}^1|F^{-1}_{b(\vec{x}),n}(u) - F^{-1}_{y^{*},n}(u)|^2 \dd u\leq \frac{1}{1-2\delta} \int_{\delta}^{1-\delta} D(u)\d u,
\end{equation}
where the correction factor $1/(1-2\delta)$ on the right-hand side makes it a trimmed estimator, such that a certain percentage (determined by $\delta$) of the tail values of the distributions can be disregarded in the computation. Hence we obtain the \gls{ucl} in \eqref{eq:int_D1-appendix}, i.e.
\begin{equation}
\mathbb{P}\left[ \ot(b(\vec{x}),y^*) \leq \frac{1}{1 - 2\delta} \int_{\delta}^{1 - \delta}  D(u) \, \dd u \right] \geq 1 - \frac{\alpha}{2}.
\label{eq:int_D1_proof_done}
\end{equation}

Next, we derive the  \gls{ucl} in \eqref{eq:int_DN-appendix}. Consider using Monte Carlo for computing the Sliced Wasserstein distance:
\[
\sot(\vec{x},\vec{x}') = \frac{1}{N}\sum_{k=1}^N \ot(\bm{\theta}^\top\vec{x}, \bm{\theta}^\top\vec{x}')
\]
We construct $\underline{q}_{\alpha, n}$ and $\overline{q}_{\alpha, n}$ independently for each $\bm{\theta}$ ($\bm{\theta}\in\Theta$) projected \gls{1d} distributions $\bm{\theta}^\top\vec{x}$ and $\bm{\theta}^\top\vec{x}'$ based on the confidence band below
\[
  \inf \PP_n\big(F_{\bm{\theta}^\top\vec{x}, n}^{-1}(\underline{q}_{\alpha, n}(u)) \leq  F_{\bm{\theta}^\top\vec{x}}^{-1}(u)\leq  F_{\bm{\theta}^\top\vec{x}, n}^{-1}(\overline{q}_{\alpha, n}(u)), \forall u\in(0,1) \big) \geq 1-\frac{\alpha}{2N}.
\]
where $F^{-1}_{\theta, n}$ is the (empirical) quantile function for the $\theta$-projected distribution and $N=|\Theta|$. Then, for any $\bm{\theta}$,
\[
\ot(\bm{\theta}^\top\vec{x}, \bm{\theta}^\top\vec{x}') =  \int_{0}^1|F^{-1}_{\bm{\theta}^\top\vec{x},n}(u) - F^{-1}_{\bm{\theta}^\top\vec{x}',n}(u)|^2 \dd u\leq \underbrace{\frac{1}{1-2\delta} \int_{\delta}^{1-\delta} D_{\bm{\theta},N}(u)\d u}_{\text{The UCL of } \ot(\bm{\theta}^\top\vec{x}, \bm{\theta}^\top\vec{x}')}
\]
holds with probability $1-\alpha/2N$. By applying the Bonferroni correction across the $N$ projection directions $\bm{\theta}_1,\ldots,\bm{\theta}_N$, we adjust the significance level for each individual test to $\frac{\alpha}{2N}$. This ensures that the combined probability of any of the \gls{ucl} failing is at most $\frac{\alpha}{2}$. Therefore, with probability at least $1-\frac{\alpha}{2}$, all individual \gls{ucl} hold simultaneously. Consequently, conditioned on these fixed projection directions, the inequality below holds almost surely,
\[
\inf_{\vec{x},\vec{x}'} \PP\left[\sot(\vec{x},\vec{x}') \leq \underbrace{\frac{1}{1-2\delta} \int_{\mathbb{S}^{d-1}}\int_{\delta}^{1-\delta} D_{\bm{\theta},N}(u)\d u\dd\sigma_N(\bm{\theta})}_{\text{The UCL of } \sot(\vec{x}, \vec{x}')} ~~\middle|~~ \bm{\theta}_1,\ldots \bm{\theta}_N\right] \geq 1-\frac{\alpha}{2}.
\]
Hence the conclusion.
\end{proof}

\section{Proof of Theorem \ref{thm:optimality}: Partial Optimality Condition}
\label{sec:proof_thm_optimality}

\begin{theorem}[\textbf{Theorem~\ref{thm:optimality}} in the main text]
Assume that the optimization problem in \eqref{eq:main_problem_empirical} is feasible, i.e.  \eqref{eq:main_problem_empirical-U_1} and \eqref{eq:main_problem_empirical-alpha_2} can be achieved simultaneously with at least $1-\alpha/2$. Then, there exists a value \( \eta^* \in [0, 1] \) such that the solution \( \vec{x}^* \) obtained by optimizing \( Q(\vec{x}, \bm{\mu}, \bm{\nu} \mid \eta^*) \), defined as
\[
  \vec{x}^{*} \triangleq \argmin_{\vec{x}, \bm{\mu}, \bm{\nu}} Q(\vec{x}, \bm{\mu}, \bm{\nu}, \eta^*),
\]
is also the optimal solution to the original problem \eqref{eq:main_problem_empirical}, that is,
\[
\vec{x}^* = \argmin_{\vec{x}} \varphi \quad \text{s.t. } \eqref{eq:main_problem_empirical-U_1} - \eqref{eq:main_problem_empirical-alpha_2}.
\]
\end{theorem}
\begin{proof}
Remark that the minimization of $Q(\vec{x},\bm{\mu}, \bm{\nu}, \eta)$ is equivalent to the optimization below with respect to $\vec{x}$:
\[
\min_{\vec{x}} Q(\vec{x},\eta) =  (1-\eta)\cdot \inf_{\bm{\mu}} Q_{x}(\vec{x},\bm{\mu}) + \eta\cdot \inf_{\bm{\nu}} Q_{y}(\vec{x}, \bm{\nu}).
\]
Since the problem \eqref{eq:main_problem_empirical} is feasible, there exists at least one pair $(\vec{x}, \objvar) $ that satisfies the constraints \eqref{eq:main_problem_empirical-U_1}-\eqref{eq:main_problem_empirical-alpha_2} for any $ \eta $ in the interval [0, 1]. 
Given that $ Q_{x} $ and $ Q_{y} $ are continuous in their respective variables, the above definition ensures that $ Q(\vec{x},\eta) $ is continuous in $\vec{x}$ for any fixed $ \eta $.

Let us consider the set $S$ of all $ \eta $ such that the corresponding $ \vec{x}_{\eta} = \arg\min_x Q(\vec{x},\eta) $ is feasible for \eqref{eq:main_problem_empirical}. Since the constraints \eqref{eq:main_problem_empirical-U_1}-\eqref{eq:main_problem_empirical-alpha_2} are continuous in $ x $, and $ x_{\eta} $ is continuous in $\eta$, the set $S$ is nonempty and compact. Thus, there exists a minimum and maximum $ \eta $ in $ S $, denoted as $\eta_{\min}$ and $\eta_{\max}$ respectively.

We now show that at least for one $ \eta^* \in [\eta_{\min}, \eta_{\max}] $, the $ \vec{x}_{\eta^*} $ is optimal for \eqref{eq:main_problem_empirical}. Note that $\eta$ is a parameter that interpolates between $Q_{x}(\vec{x},\bm{\mu})$ and $  Q_{y}(\vec{x}, \bm{\nu})$. Since $ Q(\vec{x},\eta) $ is continuous in $\vec{x}$ and $ \eta $, and the feasible region is closed and bounded, by the Extreme Value Theorem, $ Q(\vec{x},\eta) $ achieves its minimum for some $ \vec{x}^{*} $ in the feasible region of \eqref{eq:main_problem_empirical}.

Hence, there exists an $ \eta^* \in [\eta_{\min}, \eta_{\max}] $ such that $ \vec{x}^* = \arg\min_x Q(\vec{x},\eta^*) $ satisfies all the constraints \eqref{eq:main_problem_empirical-U_1}-\eqref{eq:main_problem_empirical-alpha_2} and thus is optimal for \eqref{eq:main_problem_empirical}. This concludes the proof.
\end{proof}

\section{Optimization of $\eta$}
\label{sec:eta_optimization}
We explore the optimization strategy for $\eta$, as outlined in line 1 of both Algorithm \ref{alg:eta_discrete} and Algorithm \ref{alg:eta_bisection}. The parameter $\eta$, as defined in \eqref{eq:Q-eta}, plays a crucial role in determining the descent direction during the optimization process, as seen in line \ref{alg:riemannian-nabla_Q} of Algorithm \ref{alg:riemannian}. The optimization aims to adjust this direction to fulfill the dual chance constraints \eqref{eq:main_problem_C1} and \eqref{eq:main_problem_C2}, effectively maximizing the probability of meeting both constraints. 

We first consider the Interval Narrowing in Algorithm~\ref{alg:eta_bisection}. The general strategy involves the following steps:
\begin{enumerate}
\item When either $U_{x} - \overline{\sot}$ or $U_{y} - \overline{\ot}$ is negative (indicating constraint violation), while the other is non-negative, priority is given to addressing the constraint that is violated (i.e., the one with the negative value). In this case, optimizing the non-violated constraint is less urgent.
\item If both $U_{x} - \overline{\sot}$ and $U_{y} - \overline{\ot}$ are negative or positive, $\eta$ is assigned a value based on these gaps. Let $a = U_{x} - \overline{\sot}$ and $b = U_{y} - \overline{\ot}$. The value of $\eta$ is then determined as follows:
\begin{equation}
\eta = \left\{\begin{array}{ll}
\frac{b}{a+b} & \text{if $a$ and $b$ are both negative} \\
\frac{a}{a+b} & \text{if $a$ and $b$ are both non-negative (but not both zero)} \\
0.5 & \text{if both $a$ and $b$ are exactly zero (unlikely to happen numerically in computation)}
\end{array}\right.
\label{eq:eta_strategy}
\end{equation}
In the case where both $a$ and $b$ are negative, the term with the larger absolute value (hence, more significantly violating its constraint) receives a greater weight in the optimization of \eqref{eq:Q-eta}. Conversely, if both are non-negative (and not both zero), it indicates that both constraints are satisfied, but the one with the smaller absolute value is closer to its limit, and thus receives more attention in the optimization.
\end{enumerate}

Then, we consider the Set Shrinking in Algorithm~\ref{alg:eta_discrete}. 
\begin{algorithm}[t]
\caption{Set Shrinking}
\begin{algorithmic}[1]
\REQUIRE $\overline{\sot}$, $\overline{\ot}$, $U_{x}$,$U_{y}$, $\K=\{\eta_k\}_{k=1}^K$
\ENSURE $\eta$
\STATE $\eta \leftarrow$  Balance the gaps $U_{x}\!-\overline{\sot}$ and $U_{y}\!-\overline{\ot}$ \label{alg:eta_discrete-eta}
\IF{$K>1$}
    \STATE $\K \leftarrow \K \backslash \{\eta\}$
\ENDIF
\STATE Save $\K$ as the input of the next run
\STATE \textbf{return} $\eta$
\end{algorithmic}
\label{alg:eta_discrete}
\end{algorithm}

Algorithm \ref{alg:eta_discrete} provides a way to select $\eta$ from a pre-defined set $\K$ consisting of multiple candidate values of $\eta$. If one value is expected to be selected more than once, then it is duplicated with corresponding copies in $\K$. 
Line \ref{alg:eta_discrete-eta} optimizes $\eta$ towards balancing the satisfactory of \glsplural{ucl} \eqref{eq:overline_Q_nu} and \eqref{eq:overline_Q_mu}, see Appendix \ref{sec:eta_optimization} for how the balance is achieved, in details. The set $\K$ shrinks in every run as the selected $\eta$ gets removed from it, until $K=1$ and the last $\eta$  is used until convergence.
The main different is that $\eta$ must be selected from $\K$. One could perform sorting on $\K$ first to obtain a $\K_{\text{sorted}}$, and then use bisection search to find the element in $\K_{\text{sorted}}$ that is close to $\eta$ in \eqref{eq:eta_strategy}.

\section{Proof of Theorem~\ref{thm:convergence_combined}: Convergence Rate}
\label{sec:convergence_proof}

\begin{theorem}[\textbf{Theorem~\ref{thm:convergence_combined}} in the main text]
Let $\{\vec{v}^t\}_{t=0}^T$ denote the iterates of Algorithm~\ref{alg:riemannian} combined with either Algorithm~\ref{alg:eta_discrete} or Algorithm~\ref{alg:eta_bisection}, using a stepsize $\tau = \frac{1}{\bL}$. Define 
\begin{align}
C  \triangleq  \sqrt{2\bL} + \rho L\cdot\sqrt{\frac{2}{\bL}} \text{ and } B \triangleq \sup_{\vec{v}}\{Q_{x}(\vec{v}),Q_{y}(\vec{v})\}.
\label{eq:C_and_B}
\end{align}
where $B$ is ascertained finite with finite $b$. Then, the following convergence guarantees hold:
\begin{align}
&\min_{t=0,1,\ldots,T} \norm{\tnabla Q(\vec{v}^t,\eta^t)}\leq C \left[ \frac{1}{T+1} \left( Q_{x}(\vec{v}^{0}) + Q_{y}(\vec{v}^{0}) + \Delta \right) \right]^{\frac{1}{2}},
\label{eq:convergence_conclusion}
\end{align}
where
\[
\Delta = 
\begin{cases}
0, & \text{if Algorithm~\ref{alg:eta_discrete} is used}, \\
\frac{r - l}{\kappa} B, & \text{if Algorithm~\ref{alg:eta_bisection} is used}.
\end{cases}
\]
\end{theorem}
\begin{proof}
Consider the iteration process of \discount~where $\eta$ sequentially gets to be $\eta_{(1)},\eta_{(2)},\ldots\eta_{(K)}$, with $\K=\{\eta_{(1)},\eta_{(2)},\ldots\eta_{(K)}\}$. Correspondingly, the function $Q$ defined in \eqref{eq:Q-eta} given each of these $\eta$ values are denoted by $Q_{(1)}, Q_{(2)}, \ldots Q_{(K)}$. Given that each value of $\eta$ may remain constant for several iterations before transitioning to the next, we denote $T_1, T_2, \ldots, T_K$ as the respective number of iterations for which it persists. 

Consider $Q_{(1)}$ under $\eta_{(1)}$. It is shown in \cite[Proof of Theorem 4]{peng2023block} that the sum of the square of the Riemannian gradient of $Q_{(1)}$ on $\vec{v}$ is bounded by $(b-1)\cdot C_{b-1}^2\cdot [Q_{(1)}(\vec{v}^{0})-Q_{(1)}(\vec{v}^{T_1})]$, where $b$ is the number of blocks of which the variables are subject to optimization. We have two blocks in \discount, $\vec{x}$ and $[\bm{\mu}, \bm{\nu}]$, where the former is subject to Riemannian gradient descent (i.e. lines \ref{alg:riemannian-nabla_Q} and \ref{alg:riemannian-x^{t+1}}) and the latter Exact Minimization (i.e. solving the \gls{ot} problem to the exact minimum in lines \ref{alg:riemannian-mu} and \ref{alg:riemannian-nu}). Hence $b=2$, and we can use $C$ without index $b-1$ to represent the bound, and the value of this $C$ equals exactly the one defined in \eqref{eq:C_and_B} \cite[Proof of Theorem 4]{peng2023block}. This gives the first inequality below. 
\begin{align}
\setcounter{ineq}{0}
\sum_{t=0}^{T_1-1} \norm{\renabla Q_{(1)}(\vec{v}^t)}^2 
&\leq C^2 \left[ Q_{(1)}(\vec{v}^{0}) - Q_{(1)}(\vec{v}^{T_1}) \right] \nonumber \\
&\ineq{=} C^2 \Bigg\{ \left[ (1 - \eta_{(1)}) Q_{x}(\vec{v}^0) + \eta_{(1)} Q_{y}(\vec{v}^0) \right] - \left[ (1 - \eta_{(1)}) Q_{x}(\vec{v}^{T_1}) + \eta_{(1)} Q_{y}(\vec{v}^{T_1}) \right] \Bigg\} \nonumber \\
&\ineq{\leq} C^2 \left\{ \left[ Q_{x}(\vec{v}^0) - Q_{x}(\vec{v}^{T_1}) \right] + \left[ Q_{y}(\vec{v}^0) - Q_{y}(\vec{v}^{T_1}) \right] \right\}
\label{eq:Q_(1)}
\end{align}

Note that (i) holds by the definition of $Q$ in \eqref{eq:Q-eta}, and (ii) by $0\leq \eta \leq 1$, as well as the non-negativity of $Q_{x}(\vec{v}^0) - Q_{x}(\vec{v}^{T_1})$ and  $ Q_{y}(\vec{v}^0) - Q_{y}(\vec{v}^{T_1})$. This non-negativity is due to the fact that the Riemannian gradient descent yields a decreasing objective.

Similarly, we can apply the same conclusions to $Q_{(2)},Q_{(3)}\ldots Q_{(K)}$ and obtain
\begin{align}
\sum_{t=T_1}^{T_2-1} \norm{\renabla Q_{(2)}(\vec{v}^t)}^2 
&\leq C^2 \left[ Q_{(2)}(\vec{v}^{T_1}) - Q_{(2)}(\vec{v}^{T_2}) \right] \nonumber \\ 
&\leq C^2 \left\{ \left[ Q_{x}(\vec{v}^{T_1}) - Q_{x}(\vec{v}^{T_2}) \right] + \left[ Q_{y}(\vec{v}^{T_1}) - Q_{y}(\vec{v}^{T_2}) \right] \right\}, \nonumber \\
&\vdots \nonumber \\
\sum_{t=T_{K-1}}^{T_K-1} \norm{\renabla Q_{(K)}(\vec{v}^t)}^2 
&\leq C^2 \left[ Q_{(K)}(\vec{v}^{T_{K-1}}) - Q_{(K)}(\vec{v}^{T_K}) \right] \nonumber \\ 
&\leq C^2 \left\{ \left[ Q_{x}(\vec{v}^{T_{K-1}}) - Q_{x}(\vec{v}^{T_K}) \right] + \left[ Q_{y}(\vec{v}^{T_{K-1}}) - Q_{y}(\vec{v}^{T_K}) \right] \right\}.
\label{eq:Q_(K)}
\end{align}

Denote $T=T_K-1$. For any $t=0,1,\ldots T$, let $\varrho_t$ be the corresponding index of $\eta$ at the iteration $t$. Summing up the inequalities above, we obtain
\begin{align}
\setcounter{ineq}{0}
    \sum_{t=0}^{T} & \norm{\renabla Q_{(\varrho_t)}}^2 \nonumber \\
    & \leq C^2 \bigg\{ 
      \bigg[ Q_{x}(\vec{v}^0) - Q_{x}(\vec{v}^{T_1}) + Q_{x}(\vec{v}^{T_1}) - Q_{x}(\vec{v}^{T_2}) + \cdots + Q_{x}(\vec{v}^{T_{K-1}}) - Q_{x}(\vec{v}^{T_K}) \bigg] \nonumber \\ 
     & \quad + \bigg[ Q_{y}(\vec{v}^0) - Q_{y}(\vec{v}^{T_1}) + Q_{y}(\vec{v}^{T_1}) - Q_{y}(\vec{v}^{T_2}) + \cdots + Q_{y}(\vec{v}^{T_{K-1}}) - Q_{y}(\vec{v}^{T_K}) \bigg] \bigg\} \nonumber \\
    & = C^2 \bigg\{ \bigg[ Q_{x}(\vec{v}^0) - Q_{x}(\vec{v}^{T_K}) \bigg] + \bigg[ Q_{y}(\vec{v}^0) - Q_{y}(\vec{v}^{T_K}) \bigg] \bigg\} \nonumber \\
    & \ineq{\leq} C^2 \bigg[ Q_{x}(\vec{v}^0) + Q_{y}(\vec{v}^0) \bigg] 
    \label{eq:sum_Q_discrete}
\end{align}
The inequality (i) holds due to the non-negativity of $Q_{x}$ and $Q_{y}$, as defined in \eqref{eq:Q_mu} and \eqref{eq:Q_nu} respectively.
Notice that the minimum of $\renabla Q_{\varrho_t}$ is bounded by this summation. Therefore,
\[
\min_{t=0,1,\ldots T} \norm{\renabla Q_{(\varrho_t)}}^2 \leq  \frac{1}{T+1} \sum_{t=1}^{T} \norm{\renabla Q_{(\varrho_t)}}^2 \leq \frac{C^2}{T+1}\bigg[ Q_{x}(\vec{v}^0) + Q_{y}(\vec{v}^0) \bigg]
\]
Taking root square of both side yields the convergence rate \eqref{eq:convergence_conclusion} with $\delta=0$.

We use the same notations as for the proof of the convergence for Algorithm~\ref{alg:eta_bisection}. Suppose $\eta$ sequentially gets to be $\eta_{(1)},\eta_{(2)},\ldots$, corresponding to functions $Q_{(1)}, Q_{(2)}, \ldots$
Let the initial interval in Interval Narrowing be $[l, r]$. At each iteration of the algorithm, the interval is reduced by a fixed proportion \( \kappa \) where \( 0 < \kappa < 1 \). 
After \( t \) iterations, the length of the interval will be \( (1 - \kappa)^t (r - l) \).

Consider two arbitrary iteration steps corresponding to $\eta_k$ and $\eta_{(h)}$, where $h>k$ in the sequence of $\eta$. By the definition of $Q$ in \eqref{eq:Q-eta}, we have
\[
Q_{(k)}(\vec{v}) =  (1-\eta_k)\cdot Q_{x}(\vec{v}) +  \eta_k \cdot Q_{y}(\vec{v})
\text{ and }
Q_{(h)}(\vec{v}) =  (1-\eta_{(h)})\cdot Q_{x}(\vec{v}) +  \eta_{(h)} \cdot Q_{y}(\vec{v})
\]

Next, we show that the gap between $Q_{(k)}(\vec{v})$ and $Q_{(h)}(\vec{v})$ is bounded by the interval length and $B$. Let $T_k$ be the number of elapsed iterations by the moment of $\eta_k$. 
\begin{align}
\setcounter{ineq}{0}
\bigg|Q_{(k)}(\vec{v}) - Q_{(h)}(\vec{v})\bigg| 
& = \bigg|  (1-\eta_k)\cdot Q_{x}(\vec{v}) +  \eta_k \cdot Q_{y}(\vec{v}) - (1-\eta_{(h)})\cdot Q_{x}(\vec{v}) - \eta_{(h)} \cdot Q_{y}(\vec{v}) \bigg| \nonumber \\ 
 & = \bigg|(\eta_{(h)}-\eta_k) Q_{x}(\vec{v}) + (\eta_k-\eta_{(h)}) Q_{y}(\vec{v})\bigg| \nonumber\\
 & \ineq{\leq} |\eta_{(h)}-\eta_k|\cdot \bigg|Q_{x}(\vec{v}) - Q_{y}(\vec{v})\bigg| \ineq{\leq} (1 - \kappa)^{T_k}(r - l) \bigg|Q_{x}(\vec{v}) - Q_{y}(\vec{v})\bigg|  \ineq{\leq} (1 - \kappa)^{T_k}(r - l)B
 \label{eq:bound_by_B}
\end{align}
The step (i) comes from the triangle inequality. The step (ii) holds due to the interval narrowing mechanism. Namely, the interval narrows down to $(1-\kappa)^{T_k}(r-l)$ when we encounter $\eta_k$. Note that $\eta_{(h)}$ is encountered later than $\eta_k$, hence the interval at the moment of $\eta_{(h)}$ is a subset of that of $\eta_k$. Therefore, the gap between $\eta_k$ and $\eta_{(h)}$ is no larger than $(1-\kappa)^{T_k}(r-l)$. The step (iii) holds because of the definition of $B$ ($B\triangleq\sup_{\vec{v}\in\M}\{Q_{x}(\vec{v}),Q_{y}(\vec{v})\}$), as well as the fact that $Q_{x}$ and $Q_{y}$ are non-negative.

Consider the iteration process in \discount. For any $t=0,1,\ldots T$, let $\varrho_t$ be the corresponding index of $\eta$ at the iteration $t$. We obtain the same inequalities as \eqref{eq:Q_(1)}--\eqref{eq:Q_(K)}. 
The difference is that there are infinite number of such inequalities in Interval Narrowing, rather than the fixed number $K$ as for Set Shrinking. 
Consider an arbitrary $K$ in the $\eta$ sequence $\eta_{(1)},\eta_{(2)}\ldots\eta_{(K)}\ldots$ then let $T=T_K-1$ and sum up all these inequalities:
\begin{align*}
\setcounter{ineq}{0}
\sum_{t=0}^{T} \norm{\renabla Q_{(\varrho_t)}(\vec{v}^t)} 
& \leq C^2 \bigg\{ 
    \bigg[ Q_{(1)}(\vec{v}^{0}) - Q_{(1)}(\vec{v}^{T_{1}}) \bigg] 
    + \bigg[ Q_{(2)}(\vec{v}^{T_{1}}) - Q_{(2)}(\vec{v}^{T_{2}}) \bigg] \\
& \qquad + \cdots 
    + \bigg[ Q_{(K)}(\vec{v}^{T_{K-1}}) - Q_{(K)}(\vec{v}^{T_{K}}) \bigg] 
\bigg\} \\
& \leq C^2 \bigg\{ 
    Q_{(1)}(\vec{v}^{0}) 
    + \bigg[ Q_{(2)}(\vec{v}^{T_{1}}) - Q_{(1)}(\vec{v}^{T_{1}}) \bigg] 
    + \bigg[ Q_{(3)}(\vec{v}^{T_{2}}) - Q_{(2)}(\vec{v}^{T_{2}}) \bigg] \\
& \qquad + \cdots 
    + \bigg[ Q_{(K)}(\vec{v}^{T_{K-1}}) - Q_{(K-1)}(\vec{v}^{T_{K-1}}) \bigg] 
    - Q_{(K)}(\vec{v}^{T_{K}}) 
\bigg\} \\
& \leq C^2 \bigg\{ 
    Q_{(1)}(\vec{v}^{0}) 
    + \bigg| Q_{(1)}(\vec{v}^{T_{1}}) - Q_{(2)}(\vec{v}^{T_{1}}) \bigg| 
    + \bigg| Q_{(2)}(\vec{v}^{T_{2}}) - Q_{(3)}(\vec{v}^{T_{2}}) \bigg| \\
& \qquad + \cdots 
    + \bigg| Q_{(K-1)}(\vec{v}^{T_{K-1}}) - Q_{(K)}(\vec{v}^{T_{K-1}}) \bigg| 
    - Q_{(K)}(\vec{v}^{T_{K}}) 
\bigg\} \\
& \ineq{\leq} C^2 \bigg\{ 
    Q_{(1)}(\vec{v}^{0}) 
    + (1 - \kappa)^{T_{1}} (r - l) B 
    + (1 - \kappa)^{T_{2}} (r - l) B \\
& \qquad + \cdots 
    + (1 - \kappa)^{T_{K-1}} (r - l) B 
    - Q_{(K)}(\vec{v}^{T_{K}}) 
\bigg\} \\
& \ineq{\leq} C^2 \bigg\{ 
    Q_{(1)}(\vec{v}^{0}) 
    + (r - l) B \bigg[ 1 + (1 - \kappa) + (1 - \kappa)^2 + \cdots \bigg] 
    - Q_{(K)}(\vec{v}^{T_{K}}) 
\bigg\} \\
& \ineq{\leq} C^2 \bigg[ 
    Q_{(1)}(\vec{v}^{0}) 
    - Q_{(K)}(\vec{v}^{T_{K}}) 
    + \frac{1}{\kappa} (r - l) B 
\bigg] \\
& \ineq{\leq} C^2 \bigg[ 
    Q_{(1)}(\vec{v}^{0}) 
    + \frac{1}{\kappa} (r - l) B 
\bigg] \\
& \ineq{=} C^2 \bigg\{ 
    \big[ (1 - \eta_{(1)}) Q_{x}(\vec{v}^{0}) + \eta_{(1)} Q_{y}(\vec{v}^{0}) \big] 
    + \frac{1}{\kappa} (r - l) B 
\bigg\} \\
& \ineq{\leq} C^2 \bigg[ 
    Q_{x}(\vec{v}^{0}) + Q_{y}(\vec{v}^{0}) 
    + \frac{1}{\kappa} (r - l) B 
\bigg]
\end{align*}

The step (i) holds because of the inequality \eqref{eq:bound_by_B} for arbitrary two values $\eta_k$ and $\eta_{(h)}$ ($h>k$), following which we obtain the summation of the geometric series in (ii) ($0<\kappa<1$). In step (iii) we take the summation of the series. Step (iv) is because the non-negativity of $Q_{(K)}$. Step (v) is by the definition of $Q$ in \eqref{eq:Q-eta}. Finally, step (vi) is because of $0<\eta<1$ as well as the non-negativity of $Q_{x}$ and $Q_{y}$.

Notice that the minimum of $\renabla Q_{\varrho_t}$ is bounded by this summation $\sum_{t=0}^{T} \norm{\renabla Q_{(\varrho_t)}(\vec{v}^t)}$. Therefore,  we have
\[
\min_{t=0,1,\ldots T} \norm{\renabla Q_{(\varrho_t)}}^2\leq \frac{1}{T+1}\sum_{t=0}^{T}\norm{\renabla Q_{(\varrho_t)}(\vec{v}^t)} \leq \frac{C^2}{T+1}\bigg[ Q_{x}(\vec{v}^{0})  + Q_{y}(\vec{v}^{0})+ \frac{1}{\kappa}(r - l)B \bigg]
\]
We remark that $B$ is finite because of its definition in \eqref{eq:C_and_B}. Note that $Q_{x}$ and $Q_{y}$ are continuous on the bounded and compact manifold $\M_1$ with respect to $\vec{x}$. Since $\bm{\mu}, \bm{\nu}\in\Pi$ (namely, each being non-negative and with elements summed up to 1) during the optimization, the two functions $Q_{x}(\vec{v})$ and $Q_{y}(\vec{v})$ are therefore bounded. 

Hence the conclusion.
\end{proof}

\section{Numerical Results Extended}
\label{sec:numerical_extended}

This subsection extends our nemurical results by quantitative experiments, performed on German Credit \citep{misc_statlog_(german_credit_data)_144}, and Cardiovascular Disease \citep{7qm5-dz13-20}.

\paragraph{Clarification of the Metrics in Numerical Results} Recall that \gls{dce} demonstrates a significant advantage over the baseline methods (AReS, Globe, and DiCE) in \tablename~\ref{tab:baseline_heloc}. While these baseline methods were developed to solve different explanation problems, the metrics in \tablename~\ref{tab:baseline_heloc} were chosen to fairly reflect the gaols of all methods, rather than being tailored specifically for \gls{dce}. Below, we explain the metrics in detail and their relevance:
\begin{itemize}
    \item \textit{Coverage} is a widely used metric for \gls{ce} algorithms. It is defined to be the ratio of the number of valid counterfactual data points over the total data points found by a \gls{ce} algorithm.
    \item \textit{AReS Cost} was proposed in AReS and is used in both AReS and Globe, defined to be the magnitude of feature changes given known cost on each feature. Both the two algorithms were specifically designed to optimize this metric and have direct knowledge of it. Neither DiCE nor our proposed \discount algorithm directly incorporates such knowledge.
    \item \textit{Percentiles Difference} and \textit{Distribution Shift}  These metrics measure the shift in quantile distributions between factual and counterfactual instances. Both \discount and DiCE aim to indirectly minimize such shifts:
        \begin{itemize}
            \item \discount does so by minimizing the Optimal Transport (OT) distance between the factual and counterfactual distributions.
            \item DiCE minimizes Euclidean distances between factual and counterfactual instances, subject to a soft diversity constraint.
        \end{itemize}
    \item \textit{OT} and \textit{MMD} Proximity: \discount is explicitly designed to optimize OT distance, defined in \eqref{eq:ot} and \eqref{eq:eot}. MMD is defined as 
\[
\text{MMD}^2(x, x') = \frac{1}{n^2} \sum_{i=1}^{n} \sum_{j=1}^{n} k(x_i, x_j) 
+ \frac{1}{m^2} \sum_{i=1}^{m} \sum_{j=1}^{m} k(x'_i, x'_j)
- \frac{2}{nm} \sum_{i=1}^{n} \sum_{j=1}^{m} k(x_i, x'_j)
\]
where $k$ is the gaussian kernel.
\end{itemize}
While these metrics reflect different aspects of counterfactual explanations, they were not specifically chosen to favor \gls{dce} or the \discount algorithm. Instead, they were selected to ensure fair evaluation across different CE methods by covering a range of performance goals. Importantly, our results in \tablename~\ref{tab:baseline_heloc} show that even though \discount was not specifically designed to optimize metrics like Coverage or AReS Cost, it still performs well on them when OT proximity is optimized. This highlights the versatility and robustness of Discount across diverse evaluation criteria. The results demonstrate that \gls{dce} and \discount not only excel in tasks they were designed for but also perform competitively in tasks aligned with the objectives of the baselines.

\begin{figure}[htbp]
    \centering
        \scalebox{0.6}{    \begin{tikzpicture}
    \begin{axis}[
    title style={font=\large, yshift=-1.5ex},
    xlabel={Quantiles},
    ylabel={Cardiovascular Risk},
    xlabel style={font=\large},
    ylabel style={font=\large},
    xticklabel style={font=\large},
    yticklabel style={font=\large},
    legend style={
        draw=none,
      font=\large,
       legend pos=north west,
        legend image code/.code={
            \draw[mark repeat=2,mark phase=2]
                plot coordinates {
                    (0cm,0cm)
                    (0.3cm,0cm) 
                };
        },
    },
    xmin=0, xmax=1,
    ymin=-0.02, ymax=1.02,
    grid style=dashed,
    legend cell align={left},
    width=\linewidth,
    height=0.6\linewidth,
    scaled x ticks=false,
    xticklabel style={/pgf/number format/fixed}
    ]
    
    \addplot[
        color=black,
        ]
        coordinates {
 (0.0,0.0) (0.01,0.0) (0.02,0.0) (0.03,0.0) (0.04,0.0) (0.05,0.0) (0.06,0.0) (0.07,0.0) (0.08,0.0) (0.09,0.0) (0.1,0.0) (0.11,0.0) (0.12,0.0) (0.13,0.0) (0.14,0.0) (0.15,0.0) (0.16,0.0) (0.17,0.0) (0.18,0.0) (0.19,0.0) (0.2,0.0) (0.21,0.0) (0.22,0.0) (0.23,0.0) (0.24,0.0) (0.25,0.0) (0.26,0.0) (0.27,0.0) (0.28,0.0) (0.29,0.0) (0.3,0.0) (0.31,0.0) (0.32,0.0) (0.33,0.0) (0.34,0.0) (0.35,0.0) (0.36,0.0) (0.37,0.0) (0.38,0.0) (0.39,0.0) (0.4,0.0) (0.41,0.0) (0.42,0.0) (0.43,0.0) (0.44,0.0) (0.45,0.0) (0.46,0.0) (0.47,0.0) (0.48,0.0) (0.49,0.0) (0.5,0.0) (0.51,1.0) (0.52,1.0) (0.53,1.0) (0.54,1.0) (0.55,1.0) (0.56,1.0) (0.57,1.0) (0.58,1.0) (0.59,1.0) (0.6,1.0) (0.61,1.0) (0.62,1.0) (0.63,1.0) (0.64,1.0) (0.65,1.0) (0.66,1.0) (0.67,1.0) (0.68,1.0) (0.69,1.0) (0.7,1.0) (0.71,1.0) (0.72,1.0) (0.73,1.0) (0.74,1.0) (0.75,1.0) (0.76,1.0) (0.77,1.0) (0.78,1.0) (0.79,1.0) (0.8,1.0) (0.81,1.0) (0.82,1.0) (0.83,1.0) (0.84,1.0) (0.85,1.0) (0.86,1.0) (0.87,1.0) (0.88,1.0) (0.89,1.0) (0.9,1.0) (0.91,1.0) (0.92,1.0) (0.93,1.0) (0.94,1.0) (0.95,1.0) (0.96,1.0) (0.97,1.0) (0.98,1.0) (0.99,1.0)

        };
        \addlegendentry{Fact. $y'$ (Labels)}
    
    \addplot[
        color=red,
        ]
        coordinates {
 (0.0,0.0) (0.01,0.0) (0.02,0.0) (0.03,0.0) (0.04,0.0) (0.05,0.0) (0.06,0.0) (0.07,0.0) (0.08,0.0) (0.09,0.0) (0.1,0.0) (0.11,0.0) (0.12,0.0) (0.13,0.0) (0.14,0.0) (0.15,0.0) (0.16,0.0) (0.17,0.0) (0.18,0.0) (0.19,0.0) (0.2,0.0) (0.21,0.0) (0.22,0.0) (0.23,0.0) (0.24,0.0) (0.25,0.0) (0.26,0.0) (0.27,0.0) (0.28,0.0) (0.29,0.0) (0.3,0.0) (0.31,0.0) (0.32,0.0) (0.33,0.0) (0.34,0.0) (0.35,0.0) (0.36,0.0) (0.37,0.0) (0.38,0.0) (0.39,0.0) (0.4,0.0) (0.41,0.0) (0.42,0.0) (0.43,0.0) (0.44,0.0) (0.45,0.0) (0.46,0.0) (0.47,0.0) (0.48,0.0) (0.49,0.0) (0.5,0.0) (0.51,0.0) (0.52,0.0) (0.53,0.0) (0.54,0.0) (0.55,0.0) (0.56,0.0) (0.57,0.0) (0.58,0.0) (0.59,0.0) (0.6,0.0) (0.61,1.0) (0.62,1.0) (0.63,1.0) (0.64,1.0) (0.65,1.0) (0.66,1.0) (0.67,1.0) (0.68,1.0) (0.69,1.0) (0.7,1.0) (0.71,1.0) (0.72,1.0) (0.73,1.0) (0.74,1.0) (0.75,1.0) (0.76,1.0) (0.77,1.0) (0.78,1.0) (0.79,1.0) (0.8,1.0) (0.81,1.0) (0.82,1.0) (0.83,1.0) (0.84,1.0) (0.85,1.0) (0.86,1.0) (0.87,1.0) (0.88,1.0) (0.89,1.0) (0.9,1.0) (0.91,1.0) (0.92,1.0) (0.93,1.0) (0.94,1.0) (0.95,1.0) (0.96,1.0) (0.97,1.0) (0.98,1.0) (0.99,1.0)
        };
        \addlegendentry{Target $y^*$ (Labels)}

    \addplot[
        color=orange,
    ]
    coordinates {
     (0.0,0.048134) (0.01,0.07646) (0.02,0.080828) (0.03,0.084128) (0.04,0.101229) (0.05,0.103247) (0.06,0.106219) (0.07,0.111953) (0.08,0.116654) (0.09,0.117398) (0.1,0.117465) (0.11,0.117783) (0.12,0.117958) (0.13,0.118062) (0.14,0.119295) (0.15,0.121822) (0.16,0.12226) (0.17,0.125054) (0.18,0.125662) (0.19,0.127109) (0.2,0.129525) (0.21,0.141494) (0.22,0.144456) (0.23,0.147517) (0.24,0.14868) (0.25,0.149598) (0.26,0.152413) (0.27,0.152612) (0.28,0.154169) (0.29,0.154931) (0.3,0.157383) (0.31,0.15901) (0.32,0.161521) (0.33,0.162547) (0.34,0.163527) (0.35,0.163714) (0.36,0.163815) (0.37,0.16453) (0.38,0.173303) (0.39,0.173625) (0.4,0.177459) (0.41,0.177924) (0.42,0.18045) (0.43,0.185905) (0.44,0.188964) (0.45,0.191822) (0.46,0.194386) (0.47,0.202976) (0.48,0.213145) (0.49,0.220688) (0.5,0.227564) (0.51,0.228996) (0.52,0.235626) (0.53,0.240168) (0.54,0.241244) (0.55,0.246224) (0.56,0.261551) (0.57,0.289598) (0.58,0.317896) (0.59,0.627002) (0.6,0.69088) (0.61,0.704609) (0.62,0.705802) (0.63,0.710005) (0.64,0.718468) (0.65,0.719291) (0.66,0.725229) (0.67,0.739858) (0.68,0.75347) (0.69,0.758725) (0.7,0.760235) (0.71,0.768568) (0.72,0.773452) (0.73,0.775532) (0.74,0.777412) (0.75,0.790069) (0.76,0.808646) (0.77,0.815419) (0.78,0.816155) (0.79,0.821147) (0.8,0.830516) (0.81,0.843665) (0.82,0.846648) (0.83,0.846789) (0.84,0.850471) (0.85,0.853076) (0.86,0.855702) (0.87,0.856911) (0.88,0.856972) (0.89,0.859193) (0.9,0.860678) (0.91,0.887495) (0.92,0.893584) (0.93,0.906902) (0.94,0.911537) (0.95,0.924957) (0.96,0.926847) (0.97,0.931793) (0.98,0.933517) (0.99,0.999948)
    };
    \addlegendentry{C.Fact. $y$ (DNN)}

    \addplot[
        color=blue,
    ]
    coordinates {
 (0.0,0.184034) (0.01,0.209718) (0.02,0.21292) (0.03,0.232998) (0.04,0.233065) (0.05,0.237304) (0.06,0.237369) (0.07,0.23754) (0.08,0.237665) (0.09,0.237676) (0.1,0.237704) (0.11,0.237837) (0.12,0.237882) (0.13,0.237918) (0.14,0.237947) (0.15,0.237948) (0.16,0.243224) (0.17,0.245389) (0.18,0.246176) (0.19,0.247066) (0.2,0.251085) (0.21,0.251563) (0.22,0.253309) (0.23,0.254699) (0.24,0.258683) (0.25,0.25891) (0.26,0.26203) (0.27,0.26674) (0.28,0.271724) (0.29,0.27196) (0.3,0.275465) (0.31,0.278265) (0.32,0.28223) (0.33,0.286117) (0.34,0.288391) (0.35,0.292666) (0.36,0.294616) (0.37,0.29477) (0.38,0.297628) (0.39,0.299419) (0.4,0.301728) (0.41,0.30189) (0.42,0.305959) (0.43,0.306051) (0.44,0.307141) (0.45,0.309508) (0.46,0.316239) (0.47,0.316392) (0.48,0.316683) (0.49,0.326244) (0.5,0.330323) (0.51,0.332811) (0.52,0.336754) (0.53,0.342486) (0.54,0.345112) (0.55,0.345338) (0.56,0.353811) (0.57,0.355725) (0.58,0.419265) (0.59,0.648424) (0.6,0.649068) (0.61,0.650692) (0.62,0.651018) (0.63,0.654431) (0.64,0.660094) (0.65,0.660128) (0.66,0.66631) (0.67,0.669187) (0.68,0.674263) (0.69,0.678164) (0.7,0.678997) (0.71,0.685911) (0.72,0.689638) (0.73,0.691877) (0.74,0.6931) (0.75,0.701035) (0.76,0.712812) (0.77,0.724595) (0.78,0.731887) (0.79,0.738776) (0.8,0.744028) (0.81,0.744127) (0.82,0.753754) (0.83,0.754986) (0.84,0.766525) (0.85,0.770845) (0.86,0.771038) (0.87,0.77104) (0.88,0.771052) (0.89,0.771177) (0.9,0.771224) (0.91,0.77134) (0.92,0.77147) (0.93,0.771508) (0.94,0.801439) (0.95,0.813303) (0.96,0.875823) (0.97,0.888618) (0.98,0.902107) (0.99,0.903687)
    };
    \addlegendentry{C.Fact. $y$ (SVM)}

    \addplot[
        color=green,
    ]
    coordinates {
 (0.0,0.05007) (0.01,0.071719) (0.02,0.083807) (0.03,0.096894) (0.04,0.100011) (0.05,0.112778) (0.06,0.117262) (0.07,0.117496) (0.08,0.117498) (0.09,0.117505) (0.1,0.11755) (0.11,0.117557) (0.12,0.118172) (0.13,0.119554) (0.14,0.123147) (0.15,0.127119) (0.16,0.136288) (0.17,0.138016) (0.18,0.138448) (0.19,0.13918) (0.2,0.139212) (0.21,0.139796) (0.22,0.146326) (0.23,0.155174) (0.24,0.162262) (0.25,0.162657) (0.26,0.164757) (0.27,0.165423) (0.28,0.172139) (0.29,0.172433) (0.3,0.174703) (0.31,0.180177) (0.32,0.180924) (0.33,0.181253) (0.34,0.182075) (0.35,0.184099) (0.36,0.188346) (0.37,0.194468) (0.38,0.19783) (0.39,0.201754) (0.4,0.202211) (0.41,0.203184) (0.42,0.20367) (0.43,0.204545) (0.44,0.207768) (0.45,0.216469) (0.46,0.228624) (0.47,0.23067) (0.48,0.234557) (0.49,0.237039) (0.5,0.252618) (0.51,0.255269) (0.52,0.258358) (0.53,0.270614) (0.54,0.282552) (0.55,0.291063) (0.56,0.300976) (0.57,0.305788) (0.58,0.364877) (0.59,0.679752) (0.6,0.696714) (0.61,0.767996) (0.62,0.769294) (0.63,0.781509) (0.64,0.796918) (0.65,0.82687) (0.66,0.827691) (0.67,0.831358) (0.68,0.844014) (0.69,0.851577) (0.7,0.851628) (0.71,0.852545) (0.72,0.856325) (0.73,0.856807) (0.74,0.861041) (0.75,0.863733) (0.76,0.865504) (0.77,0.868324) (0.78,0.872806) (0.79,0.876397) (0.8,0.87646) (0.81,0.879068) (0.82,0.880438) (0.83,0.880632) (0.84,0.880699) (0.85,0.884476) (0.86,0.887539) (0.87,0.887757) (0.88,0.887884) (0.89,0.888038) (0.9,0.888131) (0.91,0.890307) (0.92,0.901975) (0.93,0.904423) (0.94,0.932436) (0.95,0.946241) (0.96,0.949443) (0.97,0.958157) (0.98,0.962398) (0.99,1.0)
    };
    \addlegendentry{C.Fact. $y$ (RBFNet)}
    \end{axis}
    \end{tikzpicture}}
        \caption{\small [\textit{Cardiovascular Disease}] The models are trained on all features whereas the \gls{dce} optimization is performed only on age, weight, and height.}
        \label{fig:models_cf_quantile}
\end{figure}
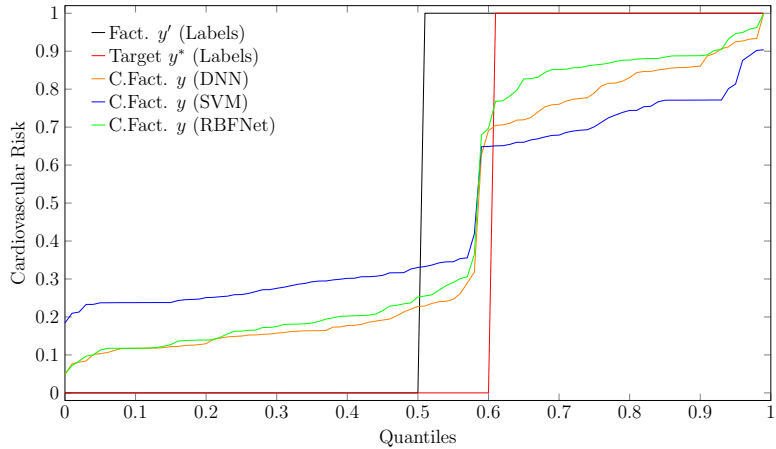

\begin{figure}[h]
    \centering
    \includegraphics[width=\linewidth]{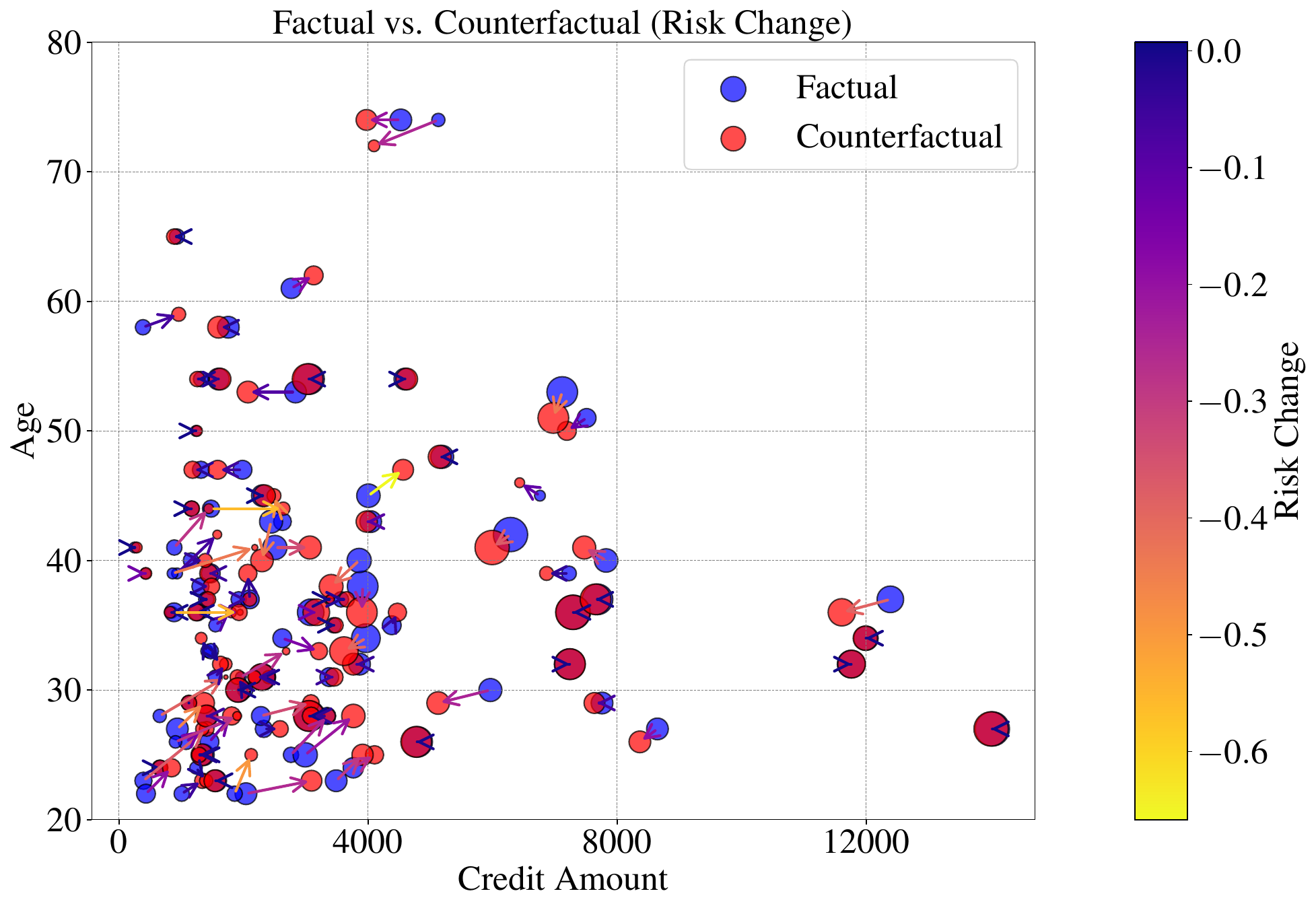}
    \caption{[\textit{German-Credit, DNN}] Risk with respect to credit amount, age, and duration (indicated by the size of each point). The models are trained on all features whereas the DCE optimization is performed only on credit amount, age, and duration.}
    \label{fig:german_credit_scatter}
\end{figure}

\textbf{Counterfactual Reasoning}. Our investigation into the German Credit dataset, depicted in \figurename~\ref{fig:german_credit_validity} and \figurename~\ref{fig:german_credit_scatter}, explores the model's distinct responses to groups with similar feature distributions. Key observations include the model's sensitivity to age and credit amount, where minor variations can result in significant risk alterations, particularly in the 40–50 age group with around 8000 in credit amount. Notably, we identified a counterintuitive behavior where increasing credit amounts paradoxically lowers risk for younger individuals. This suggests a misinterpretation of risk factors by the model, mistaking age-related risk indicators as credit amount related. This is because there are quite many young (age 20--30) people who took a small credit amount for loan in the dataset, and their risk is marked as high because of their age (rather than the fact of taking small credit). Yet the model learns its relation to credit amount, failing to realize this being not causation.
Employing \gls{dce}, we analyze quantile shifts across features, allowing us to delve into the model's decision-making process and discern substantial risk profile discrepancies between groups, even with identical distributions in certain features. This approach yields deeper statistical insights into the model’s distributional behavior and its implications.

In our study with the Cardiovascular Disease dataset, we employed three models: \gls{dnn}, \gls{svm}, and \gls{rbfnet}, training them on all available features. We then selected a sample of 100 data points from the test set to represent our empirical factual distribution. Specifically, we curated a subset of 100 individuals with low cardiovascular risk to form the target output distribution $y^*$. The quantile-based comparison of the factual $y$ and target $y^*$, as shown in \figurename~\ref{fig:models_cf_quantile}, underscores the effectiveness of \gls{dce}. Notably, the risk profiles of these counterfactual distributions, when converted to binary form, closely align with the target distribution. 

We remark such insights for both dataset above are uniquely accessible through a \gls{dce} framework, which goes beyond the capabilities of classical or group-based \glsplural{ce} by focusing on distributional shifts and relationships. Additionally, we discussed how the transportation plan $\bm{\mu}$ and $\bm{\nu}$ can be used for reasoning in Appendix~\ref{sec:geometric}.

\begin{table*}[h]
    \centering
    \caption{\small Performance of counterfactual diversity. For each row, we use bold text to highlight the column with the highest score (hence the best performance). The DPC Score is defined to be the diversity score normalized by OT proximity then multiplied by coverage. The data diversity shows the diversity score computed on the dataset rather than just the counterfactual, for reference. The experiments are averaged over 10 runs.}
    \label{tab:diversity}
    \begin{tabular}{cccccccccc}
\multicolumn{10}{c}{HELOC (Data Diversity 0.239)}                                                                                                                                                           \\ \toprule
\multicolumn{1}{c|}{\multirow{2}{*}{Metric}} & \multicolumn{3}{c|}{DNN}                            & \multicolumn{3}{c|}{RBF}                                              & \multicolumn{3}{c}{SVM}        \\
\multicolumn{1}{c|}{}                        & Globe & DiCE  & \multicolumn{1}{c|}{Discount}            & Globe          & DiCE           & \multicolumn{1}{c|}{Discount}            & Globe & DiCE  & Discount            \\ \hline
\multicolumn{1}{c|}{Diversity Score}         & 0.244 & 0.003 & \multicolumn{1}{c|}{\textbf{0.278}} & 0.254          & \textbf{0.295} & \multicolumn{1}{c|}{0.225}          & 0.239 & 0.001 & \textbf{0.246} \\
\multicolumn{1}{c|}{DPC Score}               & 0.028 & 0.007 & \multicolumn{1}{c|}{\textbf{1.351}} & 0.003          & 0.156          & \multicolumn{1}{c|}{\textbf{0.384}} & 0.017 & 0.001 & \textbf{0.720} \\ \bottomrule
\multicolumn{10}{l}{}                                                                                                                                                                                       \\
\multicolumn{10}{c}{COMPAS (Data Diversity 3.266)}                                                                                                                                                          \\ \toprule
\multicolumn{1}{c|}{\multirow{2}{*}{Metric}} & \multicolumn{3}{c|}{DNN}                            & \multicolumn{3}{c|}{RBF}                                              & \multicolumn{3}{c}{SVM}        \\
\multicolumn{1}{c|}{}                        & Globe & DiCE  & \multicolumn{1}{c|}{Discount}            & Globe          & DiCE           & \multicolumn{1}{c|}{Discount}            & Globe & DiCE  & Discount            \\ \hline
\multicolumn{1}{c|}{Diversity Score}         & 3.266 & 1.738 & \multicolumn{1}{c|}{\textbf{3.861}} & 3.299          & 1.964          & \multicolumn{1}{c|}{\textbf{3.339}} & 3.361 & 2.011 & \textbf{3.448} \\
\multicolumn{1}{c|}{DPC Score}               & 0.408 & 4.022 & \multicolumn{1}{c|}{\textbf{28.50}} & \textbf{6.847} & 0.000          & \multicolumn{1}{c|}{6.668}          & 0.109 & 8.526 & \textbf{23.78} \\ \bottomrule
\end{tabular}
\end{table*}

\textbf{Counterfactual Diversity}. Additionally, a high diversity score guarantees that the explainer provides a wide range of suggested changes offering users multiple options.  In \tablename~\ref{tab:diversity}, the diversity Score is computed as the average pairwise distance within the empirical counterfactual distributions, following \citep{mothilal2020explaining}. The diversity-proximity-coverage (DPC) score is defined to be the diversity score divided by the OT proximity then multiplied by coverage. The DPC score is a comprehensive metric to justify the diversity, as it amortizes the payoff of obtaining diversity over the proximity. Practically, we want the counterfactual to score highly in all of coverage, proximity, and diversity.
\discount~achieves the best performance in almost all datasets and all models. One exception is \gls{globe} with RBF, which trades off its coverage significantly for gaining a higher diversity score.

\begin{figure*}[t]
    \centering
    \includegraphics[width=\textwidth]{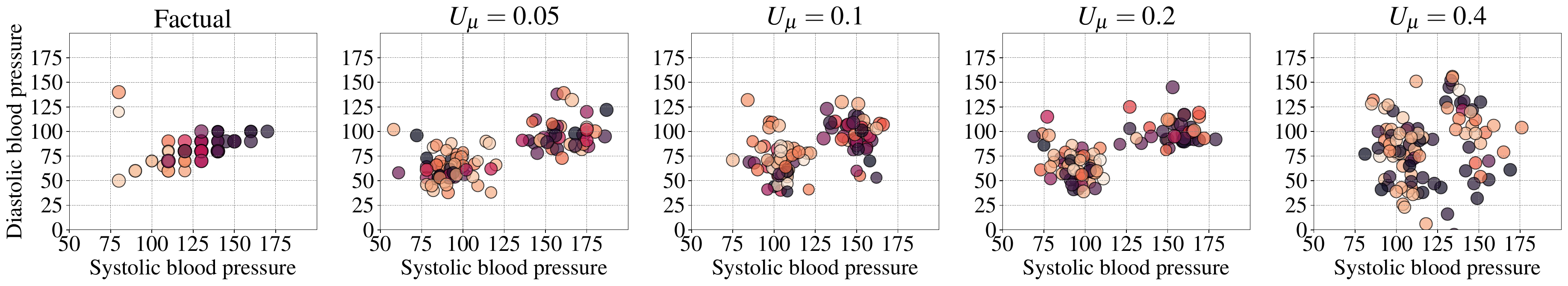}
    \caption{\small [\textit{Cardiovascular Disease}, RBFNet] The model is trained on all features, whereas the \gls{dce} optimization is performed only on the two features Diastolic/Systolic blood pressure. The four counterfactuals are obtained by optimizing from the same factual (show left). The darker color indicates the higher risk (see the color bar in \figurename~\ref{fig:german_credit_scatter}).  }
    \label{fig:diversity}
\end{figure*}

Besides, \figurename~\ref{fig:diversity} shows the capability of \discount~for generating diverse counterfactual distributions with different setups of $U_{x}$ given fixed $U_{y}$. These  counterfactuals differ in their features distributions, yet all yield very close prediction distributions. A small $U_{x}$ constraints the flexibility of counterfactual optimization but ascertains the generated distribution is close to observations. One could use $U_{x}$ to control the trade-off between validity and proximity.

\section{Geometric Explanations and Open Problems Discussions}
\label{sec:geometric}

The $2$-Wasserstein distance effectively determines how ``far apart'' two distributions are in a geometric sense. The confidence interval defined by the Wasserstein ball then represents a geometric region on this manifold, encompassing counterfactual distributions that are within a certain 'distance' (as measured by the $2$-Wasserstein metric) from the central factual distribution.  

\begin{figure}[h]
\centering
\begin{tikzpicture}
    \begin{axis}[
        trig format plots=rad,
        view={60}{80}, 
        hide axis, 
        width=0.4\linewidth,
        at={(0,0)}
    ]
        \addplot3 [surf, opacity=0.3, domain=0:6, domain y=0:6, samples=30, 
        samples y=50] 
        {(1+sin(x/10))*sin(x/2)*cos(y)};

        \addplot3 [
            only marks,
            mark=ball,
            ball color=red, 
            mark size=0.6mm, 
            nodes near coords={$x'$},
            name path=node1 
        ] coordinates {(3, 3, 0.3)};
        
        \addplot3 [
            only marks,
            mark=ball,
            ball color=blue, 
            mark size=0.6mm, 
            nodes near coords={$x$},
            name path=node2 
        ] coordinates {(4, 3.5, 0.1)};

        \addplot3 [
            no marks,
            smooth, 
            thick, 
            color=red, 
        ] coordinates {
            (3, 3, 0.3)
            (3.5, 3.25, { (1+sin(deg(3.5)/10))*sin(deg(3.5)/2)*cos(deg(3.25)) })
            (3.75, 3.375, { (1+sin(deg(3.75)/10))*sin(deg(3.75)/2)*cos(deg(3.375)) })
            (4, 3.5, { (1+sin(deg(4)/10))*sin(deg(4)/2)*cos(deg(3.5)) })
        };

    \end{axis}
\end{tikzpicture}
\hspace{2cm}
\begin{tikzpicture}
    \begin{axis}[
        view={120}{40}, 
        axis lines=middle, 
        zmin=-0.5, zmax=0.5, 
        ymin=-0.5, ymax=0.5, 
        xmin=-0.5, xmax=0.5, 
        ticks=none, 
        axis equal, 
        axis line style={draw=none}, 
        clip=false,
        scale uniformly strategy=units only,
    ]

    \addplot3 [
        only marks,
        mark=ball,
        ball color=red, 
        mark size=1.25, 
        nodes near coords={$x'$},
        nodes near coords style={above},
    ] coordinates {(0, 0, 0)}; 

    \draw [dashed, thick, blue] (axis cs:0,0,0) -- (axis cs:0.5,0,0) node[midway, below right] {$U_{x}$};

    \addplot3 [
        surf,
        opacity=0.1, 
        fill=blue!50!white, 
        domain=0:360, 
        y domain=0:180, 
        samples=42, 
        variable=\u, 
        variable y=\v, 
        point meta=u, 
    ] (
        {0.5 * sin(v) * cos(u)}, 
        {0.5 * sin(v) * sin(u)}, 
        {0.5 * cos(v)} 
    );

    \end{axis}
\end{tikzpicture}
\caption{\small Geometric illustration of geodesic distance between two distributions $x$ and $x'$ and the Wasserstein ball centered at $x'$.}
\label{fig:ball}
\end{figure}
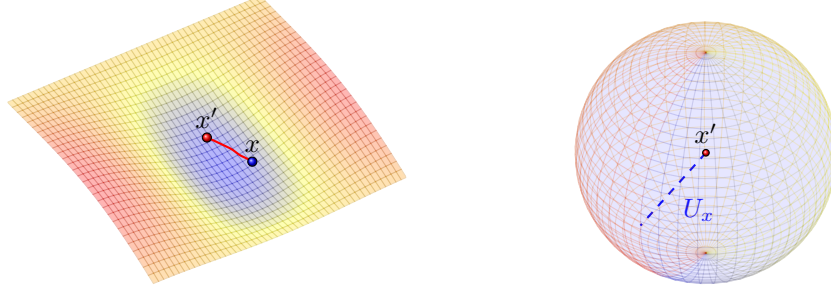

In the context of a Riemannian manifold, geodesics are the shortest paths between two points (in this case, between two probability distributions). The $2$-Wasserstein distance is a geodesic distance on the manifold \cite[Corollary 7.22]{cedric2008optimal}. This framework allows for a geometric interpretation of how distributions relate to each other within this space. For any given factual distribution $\vec{x}'$, the region of counterfactual distribution $\vec{x}$ defined by the probability constraint can geometrically be visualized as a kind of radius $U_{x}$ sphere (i.e. Wasserstein Ball) centered at $\vec{x}'$ where a significant majority (more than the probability $\objvar$) strictly adheres to the boundary constraint of the sphere. See \figurename~\ref{fig:ball} for an illustration.

\textbf{Transportation Plan}. The transportation plan can be conceptualized as a geometric mapping that delineates the optimal transfer of probability mass across the manifold from the factual input distribution $\vec{x}'$ to the counterfactual input distribution $x$. Embedded within the structure of the manifold, the values of $\bm{\mu}$ and $\bm{\nu}$ represent the relative extent of mass reallocation necessary between specific points or regions in the probability space. Large values in $\vec{\Pi}$ indicate areas that require substantial mass movement, reflecting significant distributional changes, while smaller values suggest minor adjustments. Zero values denote regions where the existing distribution aligns with the target and requires no modification. This transportation plan, therefore, becomes an integral tool in understanding and visualizing the nature of distributional shifts required in counterfactual scenarios. 

\begin{figure}
    \centering
    \includegraphics[width=0.7\textwidth]{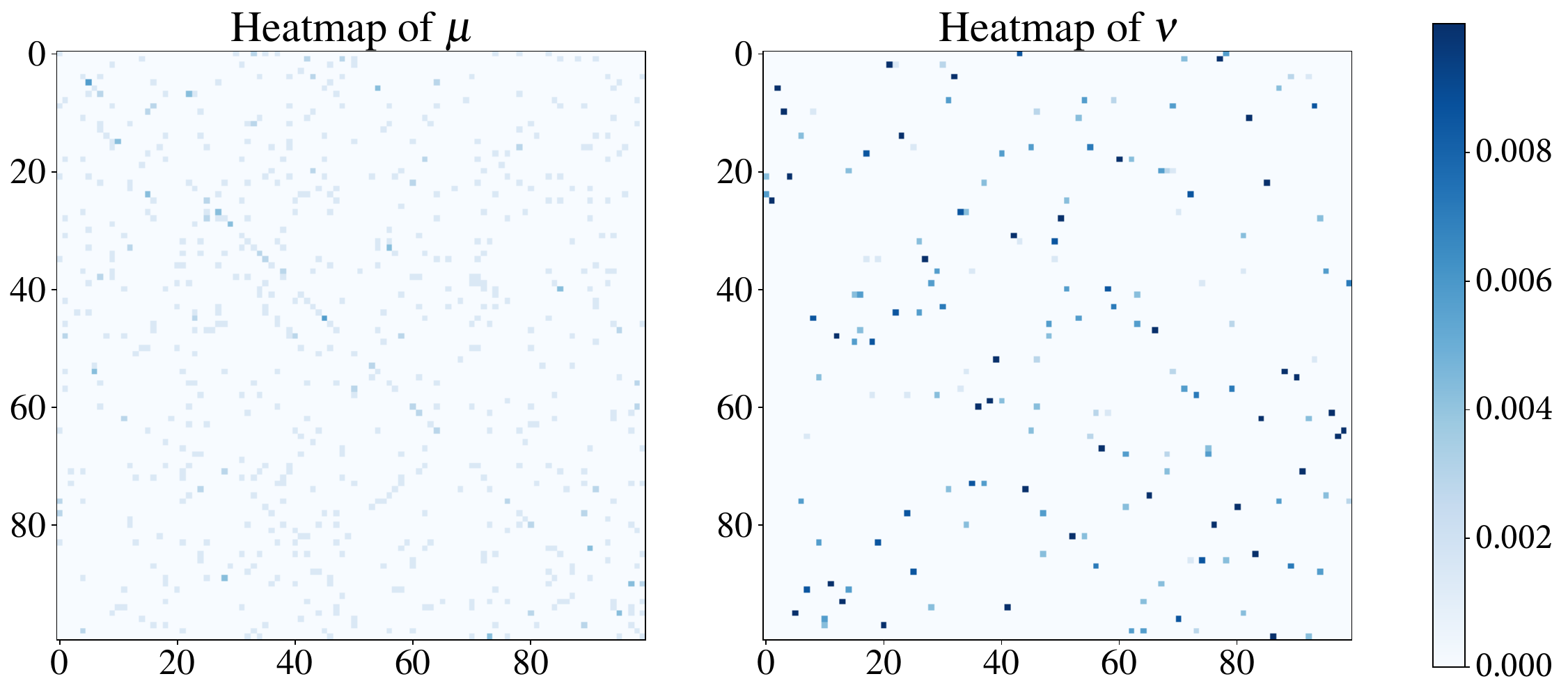}
\caption{\small [\textit{Cardiovascular Disease}, DNN] Visualization of the transportation plan for $\bm{\mu}$ and $\bm{\nu}$, as computed in lines \ref{alg:riemannian-mu} and \ref{alg:riemannian-nu} of Algorithm~\ref{alg:riemannian}, respectively. This heatmap depicts the averaged values of $\bm{\mu}$ over $\bm{\theta}\in\Theta$, normalized to form a distribution with elements summing to one. Factual and counterfactual points are aligned (in their original order) along the x-axis and y-axis, respectively.}
    \label{fig:transportation_plan}
\end{figure}

\figurename~\ref{fig:transportation_plan} demonstrates the use of the transport plans $\bm{\mu}$ and $\bm{\nu}$ in \gls{dce}. A key observation from the heatmap of $\bm{\mu}$ is the pronounced diagonal, suggesting that the predominant mass transfer occurs between each factual point and its direct counterpart in the counterfactual data set. This implies that individuals in the factual group are more likely to correspond to their original counterparts in the counterfactual group, rather than to others. This pattern is typical when the distribution distance between $\vec{x}$ and $\vec{x}'$ is restricted to be very close. In the extreme case of identical distributions, \gls{ot} would occur strictly along the diagonal (and hence the heatmap $\bm{\nu}$ also looks strictly diagonal).

In contrast, the heat map of $\bm{\nu}$ shows a less distinct diagonal pattern, indicating a notable shift in the risk ranking of individuals from the factual group to the counterfactual group. This suggests that the individual with the highest cardiovascular risk in the factual group might not maintain a comparably high risk in the counterfactual group. Although this is not inherently problematic, it highlights that the model may be highly sensitive to certain features, with minor changes in $\vec{x}$ that potentially lead to significant alterations in $\vec{y}$. However, if the diagonal is prominently visible in $\bm{\mu}$ but not in $\bm{\nu}$, further investigation is warranted to determine whether the model is acting as intended. This discrepancy may suggest that, while the model maintains individual correspondences between the factual and counterfactual groups (as indicated by $\bm{\mu}$), the implications or outputs (reflected in $\bm{\nu}$) do not preserve these correspondences. Such a scenario could reveal potential issues or sensitivities in the model's learning process that merit closer examination.

\paragraph{Combining DCE with previous work.} Once a distributional counterfactual $\vec{x}$ has been constructed via the DCE framework, the methods proposed in \citep{kulinski2023towards} and \citep{koebler2023towards} become natural post-hoc complements for enhanced explainability. Specifically, \citep{kulinski2023towards} interprets the shift between two \emph{observed} distributions by constructing an OT-based decomposition of where and how the mass is moved. Although the counterfactual distribution $\vec{x}$ in DCE is \emph{not} empirically observed, one can sample from it (or treat it as a reweighted empirical set), thereby enabling \citep{kulinski2023towards} to generate interpretable “transportation maps” showing which subpopulations or feature ranges changed most substantially. Meanwhile, \citep{koebler2023towards} examines how real-world distribution drifts affect model performance and reliability. Applied analogously to $\vec{x}'$ (factual) and $\vec{x}$ (DCE counterfactual), the approach in \citep{koebler2023towards} can reveal how performance, fairness metrics, or predictive reliability might degrade or improve under the hypothesized distributional shift. Thus, \citep{kulinski2023towards} and \citep{koebler2023towards} effectively extend DCE’s basic capability—beyond simply constructing a new distribution—to provide finer-grained analysis of \emph{why} particular transport plans emerge and \emph{what} their consequence is for the model’s predictive behavior.

\textbf{\Gls{eot} and Information Geometry} There is also a link between \gls{ot} and information geometry, established by introducing an entropy-regularization term for $\ot$ \citep{NIPS2013_af21d0c9}, i.e. 
\begin{equation}
\ot_{\varepsilon}(\gamma_1, \gamma_2) =  \inf_{\pi\in\Pi(\gamma_1, \gamma_2)} \bigg\{ \int_{\RR^{d\times d}} \norm{a_1-a_2}^2 \d{\pi(a_1,a_2)}  + \varepsilon H(\pi|\gamma_1 \otimes \gamma_2)\bigg\}
\end{equation}
where the regularization term $H(\pi|\gamma_1 \otimes \gamma_2)$ denotes the relative entropy with respect to the product measure $\gamma_1\otimes \gamma_2$,  defined as
\begin{equation}
   H(\pi|\gamma_1 \otimes \gamma_2) = \int_{\RR^d} \log\left(\frac{\d{\pi}}{\d{\gamma_1}\d{\gamma_2}}\right) \d\pi.
\end{equation}
Using $\ot_{\varepsilon}$ as replacement for its counterparts in the objective and the constraint of the formulation~\eqref{eq:main_problem} reduces the computational effort to obtain the transportation plan $\pi$. Furthermore, this regularization term is equivalent to using \gls{kl} divergence as a regularizer and naturally defines certain geometrical structures from the information geometry point of view \citep{karakida2017information, khan2022optimal}. The role of the product of marginal distributions $\gamma_1\otimes \gamma_2$ in the divergence term \gls{kl} is to establish a baseline of independence between the distributions being compared. By measuring the divergence of the transport plan $\pi$ from this baseline, the entropy term encourages solutions that respect the underlying structure of the individual distributions, rather than overly concentrating the mass transfer in a few specific regions. By increasing the value of $\varepsilon$, one tends to seek a counterfactual $\vec{x}$ that is drawn "more independently" from the distribution of the original input $\vec{x}'$, leading to solutions that consider a broader range of possibilities, resulting in more diverse and potentially more insightful counterfactuals that better capture the overall structure of the data.

The integration of the entropic term makes $\ot$ a quasi-distance (due to loss of symmetry). The entropic Wasserstein quasi-distance may be regarded as such a metric in scenarios where the expense associated with the transfer of information across distinct nodes is contingent upon the spatial coordinates of these nodes \citep{oizumi2014phenomenology}, and the entropic term controls the amount of information integration \citep{amari2018information}, which is measured by the amount of interactions of information among different nodes.

\textbf{Open Problems of \gls{dce} with \gls{eot}}. The solution of the chance-constrained optimization problem \eqref{eq:main_problem} is essentially based on a statistically trustworthy estimation of the upper bounds of $\ot$ and $\sot$. Remark that Theorem~\ref{thm:interval} hinges on the fact that the
\gls{1d} Wasserstein distance may be expressed as quantile functions, so it does not generalize to entropic regularized $\ot$ or $\sot$ (because they are no longer rigorously tied to the quantile expressions). \gls{clt} is proved for the \gls{eot} cost that is centered at the population cost, and this yields an asymptotically valid confidence interval for \gls{eot} \cite[Theorem 3.6]{del2023improved}. Furthermore, bootstrap is shown to be valid for \gls{eot} \cite[Theorem 7]{goldfeld2022statistical}, making asymptotic inference for the interval efficient and straightforward. To our knowledge, it remains open whether the sliced version of \gls{eot} accepts an (even asymptotic) inference for the confidence interval. Hence, if one uses \gls{eot} in \eqref{eq:main_problem_C1}, its theoretical foundation calls for further investigation.

\section{Reproducibility}
\label{sec:reproducibility}

To ensure the reproducibility of our experiments for the proposed \gls{dce}, we provide detailed information on our computational environment and data preprocessing steps.

The experiments were carried out on a high performance computing (HPC) cluster, utilizing four nodes (one for each dataset) in parallel. Each node is equipped with two Intel Xeon Processor 2660v3 (10 core, 2.60GHz) CPUs and 128 GB of memory. The duration of the experiment is approximately 5 hours per node, depending on the size of the dataset. Although reproducing these experiments on a standard laptop is feasible, it generally requires more computational time compared to using an HPC cluster.

For the four datasets used in our experiments, numerical features are standardized to ensure that each feature has a mean of zero and a standard deviation of one. Categorical features are encoded using either label encoding or one-hot encoding. We did not observe significant differences in the performance of the \gls{dce} framework between these two encoding methods. The data sets are divided into training and testing sets with a ratio of $0.8:0.2$.

The experiments were implemented using Python, with key libraries including NumPy, Pandas, Scikit-learn, and PyTorch. Detailed versions and dependencies are given in the source code to facilitate exact replication of the experiments.

The complete codebase, along with detailed instructions for setting up the environment and running the experiments, is available in a public repository (see the footnote in Section~\ref{sec:numerical}). This includes scripts for data preprocessing, model training, evaluation, and the generation of \gls{ce}.

\vfill

\end{document}